%% file: neurips_2025.tex
\documentclass{article}

    \PassOptionsToPackage{numbers, compress}{natbib}
\input{sections/fig1_helpers}

    \usepackage[preprint]{neurips_2025}

\usepackage[utf8]{inputenc} % allow utf-8 input
\usepackage[T1]{fontenc}    % use 8-bit T1 fonts
\usepackage{hyperref}       % hyperlinks
\usepackage{url}            % simple URL typesetting
\usepackage{booktabs}       % professional-quality tables
\usepackage{amsfonts}       % blackboard math symbols
\usepackage{nicefrac}       % compact symbols for 1/2, etc.
\usepackage{microtype}      % microtypography
\usepackage{xcolor}         % colors
\usepackage{amsmath}
\usepackage{amssymb}
\usepackage{mathtools}
\usepackage{amsthm}
\usepackage{cleveref}

\newcommand{\R}{\mathbb{R}}  % real numbers
\DeclareMathOperator{\diag}{diag}

\usepackage{algorithm}
\usepackage[noend]{algpseudocode}
\usepackage{xcolor}
\definecolor{diff}{rgb}{0.0,0.2,0.8}  % blue

\usepackage{tikz}
\usetikzlibrary{arrows.meta, positioning, 3d}

\theoremstyle{plain}      % For theorems, lemmas, etc. (default style: italicized text)
\newtheorem{theorem}{Theorem}[section]
\newtheorem{lemma}[theorem]{Lemma}        % Numbered in sequence with theorems
\newtheorem{corollary}[theorem]{Corollary}

\theoremstyle{definition}  % Roman font for definitions, examples, etc.
\newtheorem{definition}[theorem]{Definition}

\theoremstyle{remark}      % For remarks, notes, claims, etc.

\usepackage{enumitem}

\usepackage{graphicx}
\usepackage{subcaption}
\usepackage{wrapfig}

\usepackage{enumitem}
\usepackage[subtle]{savetrees}
\usepackage{twemojis} 

\title{\textsc{BWLer} \texttwemoji{bowling}  : Barycentric Weight Layer Elucidates a Precision-Conditioning Tradeoff for PINNs}

\newcommand{\methodname}{\textsc{BWLer}}

\author{
\textbf{Jerry Liu\textsuperscript{1}}\thanks{Corresponding author: \texttt{jwl50@stanford.edu}}\quad
\textbf{Yasa Baig\textsuperscript{2}}\quad
\textbf{Denise Hui Jean Lee\textsuperscript{3}}\quad
\textbf{Rajat Vadiraj Dwaraknath\textsuperscript{1}}\\[0.3ex]
\textbf{Atri Rudra\textsuperscript{4}}\quad
\textbf{Chris R\'{e}\textsuperscript{3}} \\[0.75ex]
\textsuperscript{1}Institute for Computational \& Mathematical Engineering, Stanford University \\
\textsuperscript{2}Department of Bioengineering, Stanford University \\
\textsuperscript{3}Department of Computer Science, Stanford University \\
\textsuperscript{4}Department of Computer Science \& Engineering, University at Buffalo
}

\begin{document}

\maketitle

\addtocounter{footnote}{-1}

\begin{abstract}
  \input{sections/abstract}
\end{abstract}

\input{sections/introduction}

\input{sections/background}

\input{sections/interpolation}

\input{sections/method}

\input{sections/baselines}

\input{sections/discussion}

\input{sections/acknowledgements}
% \newpage

\bibliography{references}
\bibliographystyle{icml2025}

\newpage
% \input{sections/appendix/checklist}

%%%%%%%%%%%%%%%%%%%%%%%%%%%%%%%%%%%%%%%%%%%%%%%%%%%%%%%%%%%%

\newpage
\appendix

\input{sections/appendix/related_work}

\newpage
\input{sections/appendix/method}

\newpage
\input{sections/appendix/experiments}

\newpage
\input{sections/appendix/theory}

%%%%%%%%%%%%%%%%%%%%%%%%%%%%%%%%%%%%%%%%%%%%%%%%%%%%%%%%%%%%

\end{document}

%% file: sections/fig1_helpers.tex
\usepackage{tikz}
\usepackage{pgfplots}
\pgfplotsset{compat=1.18}
\usetikzlibrary{positioning,arrows.meta,calc}
\usepgfplotslibrary{patchplots}   % <— enables 'patch type=polygon'

\newcommand{\PlaneSize}{4}
\newcommand{\PlaneZ}{0}
\newcommand{\GridN}{3}

\definecolor{PlaneCol}{gray}{0.85}
\definecolor{SurfCol}{RGB}{100,100,255}
\definecolor{StalkCol}{RGB}{245,166,35} % not used here but kept
\definecolor{RandCol}{RGB}{0,140,0}     % green

\newcommand{\SurfExpr}[2]{2 + 0.3*sin(deg(#1))*cos(2*deg(#2))}

\tikzset{
plane grid/.style  ={black!40, line width=0.25pt},
up arrow/.style    ={RandCol, -{Stealth[length=6.5pt]}, line width=1.6pt},
surf mesh line/.style={SurfCol!70, line width=0.25pt},  % surface lattice
stalk style/.style={StalkCol, line width=1.6pt,
    mark=*, mark size=1.4pt,
    mark options={fill=StalkCol, draw=StalkCol}},
rand  style/.style ={RandCol , line width=2.5pt,
    mark=*, mark size=1.8pt, ->, >=Stealth,
    mark options={fill=RandCol , draw=RandCol }},
bary  arrow/.style={StalkCol, dotted, line width=1.8pt, ->, >=Stealth, mark=none,
  },
}

\newcommand{\DrawBarycentricArrows}[2]{%
  \pgfmathsetmacro{\step}{\PlaneSize/(\GridN-1)}%
  \foreach \i in {0,...,\numexpr\GridN-1}{%
      \foreach \j in {0,...,\numexpr\GridN-1}{%
          \pgfmathsetmacro\xpt{-\PlaneSize/2 + \i*\step}%
          \pgfmathsetmacro\ypt{-\PlaneSize/2 + \j*\step}%
          \addplot3+[bary arrow, forget plot, z buffer=none]
          coordinates {(\xpt,\ypt,\PlaneZ) (#1,#2,\PlaneZ)};
        }%
    }%
}
\newcommand{\DrawRandomSamplesBWLER}{%
  \pgfmathsetseed{2025}%
  \def\xTarget{}%
  \def\yTarget{}%
  \foreach \k [evaluate=\k as \flag using int(\k==1)] in {1,2}{%
      \pgfmathsetmacro{\xr}{rnd*\PlaneSize - \PlaneSize/2}%
      \pgfmathsetmacro{\yr}{rnd*\PlaneSize - \PlaneSize/2}%
      \pgfmathsetmacro{\zr}{\SurfExpr{\xr}{\yr}}%
      \ifnum\flag=2
        \DrawBarycentricArrows{\xr}{\yr}%
        \addplot3+[rand style, forget plot]
        coordinates {(\xr,\yr,\PlaneZ) (\xr,\yr,\zr)};%
      \fi
    }%
}

\pgfplotsset{
  colormap={solidSurf}{color=(SurfCol) color=(SurfCol)}
}

%% file: sections/abstract.tex
Physics-informed neural networks (PINNs) offer a flexible way to solve partial differential equations (PDEs) with machine learning, yet they still fall well short of the machine-precision accuracy many scientific tasks demand.
This motivates an investigation into whether the precision ceiling comes from the ill-conditioning of the PDEs themselves or from the typical multi-layer perceptron (MLP) architecture.
We introduce the \textbf{Barycentric Weight Layer (\methodname)}, which models the PDE solution through barycentric polynomial interpolation.
A \methodname~can be added on top of an existing MLP (a \methodname-hat) or replace it completely (explicit \methodname), cleanly separating how we \emph{represent} the solution from how we take its \emph{derivatives} for the physics loss.
Using \methodname, we identify fundamental precision limitations within the MLP: on a simple 1-D interpolation task, even MLPs with $O(10^5)$ parameters stall around $10^{-8}$ relative error -- about eight orders above \texttt{float64} machine precision -- before any PDE terms are added.
In PDE learning, adding a \methodname~lifts this ceiling and exposes a tradeoff between achievable accuracy and the conditioning of the PDE loss.
For linear PDEs we fully characterize this tradeoff with an explicit error decomposition and navigate it during training with spectral derivatives and preconditioning.
Across five benchmark PDEs, adding a \methodname~on top of an MLP improves $\ell_2$ relative error by up to $30\times$ for convection, $10\times$ for reaction, and $1800\times$ for wave equations while remaining compatible with first-order optimizers.
Replacing the MLP entirely lets an explicit \methodname~reach near-machine-precision on convection, reaction, and wave problems (up to \emph{10 billion times better} than prior results) and match the performance of standard PINNs on stiff Burgers’ and irregular-geometry Poisson problems.
Together, these findings point to a practical path for combining the flexibility of PINNs with the precision of classical spectral solvers.

%% file: sections/introduction.tex
\section{Introduction}

Partial differential equations (PDEs) are the standard tool for modeling complex phenomena across science and engineering~\citep{evans2010partial}. Traditionally, PDEs have been solved using numerical methods (e.g. finite element or spectral methods~\citep{hughes2003finite, boyd1989chebyshev}) but there has been recent interest in leveraging modern machine learning (ML) techniques to solve these classical problems~\citep{brunton2022data, karniadakis2021physics}. Producing better ML-based methodologies for PDEs could enable faster simulations while maintaining the high fidelity of traditional numerical methods, with applications from weather forecasting to design optimization~\citep{mcgovern2017using, brunton2021data}.

Physics-informed neural networks (PINNs)~\citep{Raissi_2017} parametrize the solution of a PDE with a multi-layer perceptron (MLP) and enforce PDE constraints with least-squares losses during training.
The main benefit of this \textit{physics-informed} framework is flexibility: it requires no meshing, handles irregular geometries gracefully, and provides a unified methodology for diverse PDE types~\citep{raissi2019pinns, karniadakis2021physics, mishra2023estimates}.
However, PINNs have struggled to achieve high-precision solutions~\citep{Michaud_2023, McGreivy_2024, liu2025towards},
crucial for scientific applications
such as turbulence modeling or maintaining stable temporal rollouts~\citep{frisch1995turbulence}.
PDEs are particularly challenging because of their fundamentally ill-conditioned differential operators;
despite recent progress investigating the precision saturation of PINNs on PDE problems~\citep{wang2021understanding, krishnapriyan2021characterizing, wang2023multistageneuralnetworksfunction, rathore2024challengestrainingpinnsloss},
it remains unclear to what extent the issues stem from problem-inherent ill-conditioning versus the models' parameterizations.

\begin{figure}[t!]
\centering
\includegraphics[width=0.92\textwidth]{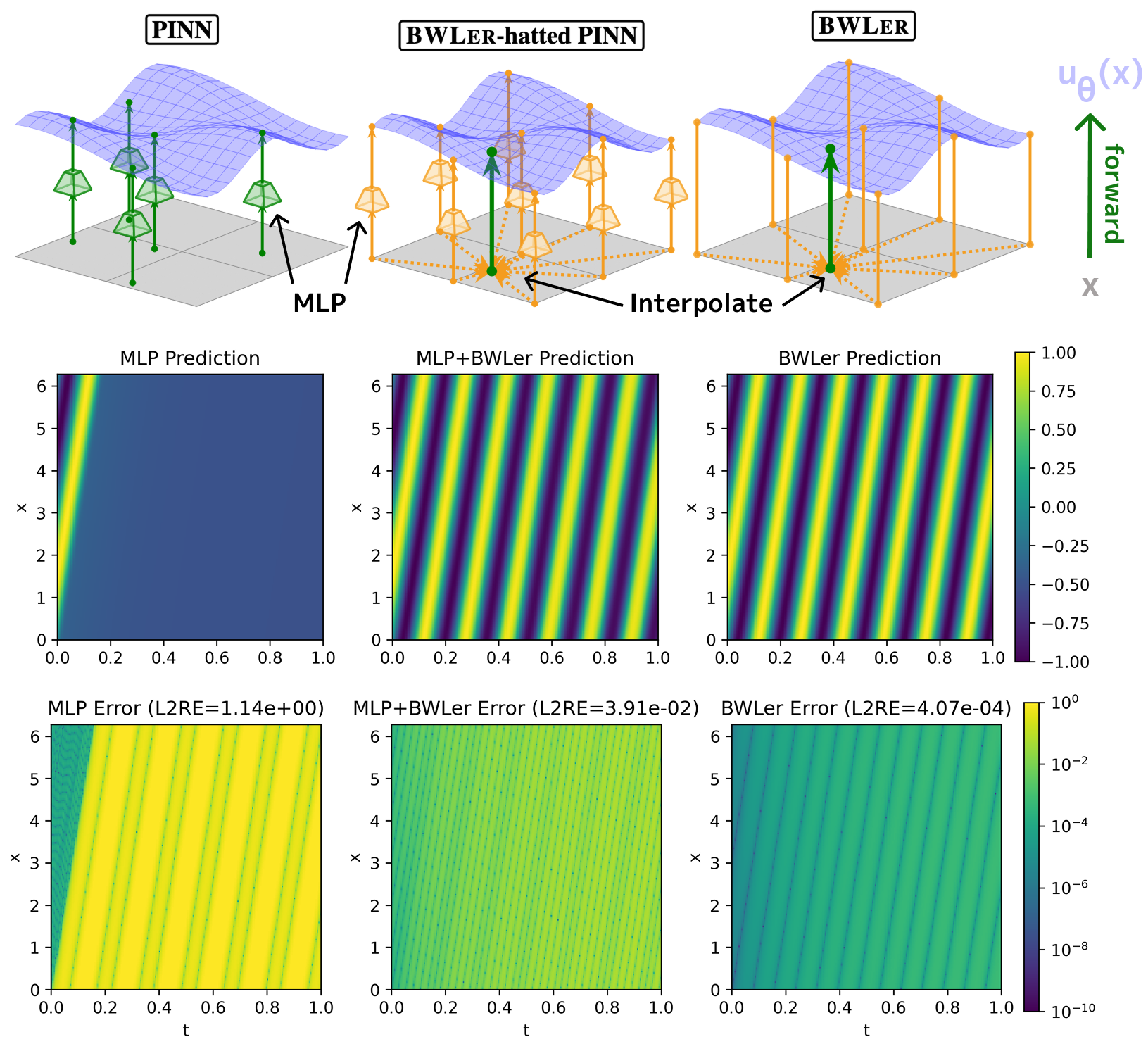}
\label{fig:pde_convection_method_comparison}

\caption{\textbf{Top: model architecture comparison.} Standard PINN evaluates an MLP throughout the domain (left). \methodname~interpolates globally based on values at discrete grid nodes; \methodname-hatted MLP obtains values using an MLP (middle), explicit \methodname~parameterizes values directly (right).
\textbf{Bottom: results for convection equation}~\citep{rathore2024challengestrainingpinnsloss}. Standard PINN stagnates at a suboptimal local minimum (left); \methodname-hatted MLP finds a qualitatively correct solution (middle); explicit \methodname~converges to higher precision (right).
}
\label{fig:main}
\end{figure}

In this work, we aim to disentangle and analyze the sources of precision limitations in PINNs. Specifically, we ask: \textit{(i) are there inherent precision bottlenecks in the MLP architectures used by PINNs, and (ii) how does the difficulty of the underlying PDE affect the precision that can be achieved?} Our study has the following three parts:
\begin{itemize}[leftmargin=*]

\item \textbf{We identify fundamental MLP precision limitations in a simple setting.}
Through systematic experiments on 1-D interpolation, we show that MLP precision plateaus around $10^{-8}$ $\ell_2$ relative error (L2RE). This is roughly eight orders of magnitude above \texttt{float64}'s machine epsilon ($2^{-52} \approx 2.22 \times 10^{-16}$), even without PDE constraints.
We demonstrate that this limitation persists as we scale network width by $16\times$ and depth by $4\times$, with precision improving by just $1$--$2$ orders of magnitude even with $1000\times$ more parameters (\Cref{fig:1d_interpolation_scaling}).
In contrast, classical polynomial representations with just $10$--$100$ parameters can provably achieve machine precision (\Cref{thm:spectral_convergence}).
Our results point to precision bottlenecks stemming from the neural network parameterization itself even beyond optimization challenges specific to PDEs.

\item \textbf{We propose a barycentric interpolation framework for PDE learning.}
Motivated by the precision of polynomials, we introduce \methodname\footnotemark\footnotetext{Code available at \href{https://github.com/HazyResearch/bwler}{\texttt{https://github.com/HazyResearch/bwler}}.}, a simple baseline that can be used as a drop-in replacement for standard PINN architectures.
\methodname~parameterizes the solution function
as a \emph{barycentric polynomial interpolant}~\citep{trefethen2013approximation, berrut2004barycentric}, where the model is defined by the function values it takes on a pre-specified discrete grid in the domain (\Cref{fig:main}, top; \Cref{alg:pinn_compare}).
Our method builds upon decades of work using polynomial interpolants to solve numerical PDEs~\citep{boyd1989chebyshev, fornberg1998practical, trefethen2000spectral, canuto2006spectral, battles2004extension}; \methodname~effectively embeds a pseudo-spectral solver into the physics-informed framework while leveraging auto-differentiation and ML optimizers.
Excitingly, \methodname~lets us decouple our choice of model parameterization (e.g. explicit grid, neural network) from our PDE derivatives calculations (e.g. finite differences, spectral derivatives). Using \methodname, we next ablate the MLP to study the effect of model parameterization vs. the PDE ill-conditioning on precision.

\item \textbf{We characterize a precision-conditioning tradeoff with \methodname.} We investigate two variants of \methodname~(\Cref{alg:pinn_compare}). \textbf{\methodname-hatted MLPs}, which apply \methodname~atop existing MLPs, outperform standard MLPs on convection, reaction, and wave equation benchmark problems by $30\times$, $10\times$, and $1800\times$ respectively.
We find \methodname~improves the PDE loss conditioning, decreasing mean eigenvalue by $5$--$10\times$ (\Cref{fig:pde_reaction_method_comparison_spectrum},~\Cref{fig:pde_wave_method_comparison_spectrum}).
We then turn to \textbf{explicit \methodname}, where the model is directly parameterized by its function values on the pre-specified grid.
We fully characterize the training error of explicit \methodname~on linear PDEs, identifying a fundamental tradeoff between \methodname's maximum achievable precision and the conditioning of the optimization problem (\Cref{thm:bwler_pde_informal}).
Motivated by our error decomposition, we vary preconditioning and derivative computations to navigate the tradeoff space during training.
For the first time, we achieve near \emph{machine precision} with PINNs on convection, reaction, and wave equation benchmarks (up to \emph{10 billion times better L2RE} than prior PINN methods) and match state-of-the-art performance on Burgers' and irregular-geometry Poisson problems (\Cref{tab:pdes_sota}).

\end{itemize}

%% file: sections/background.tex
\section{Background} \label{sec:background}

We provide background information on physics-informed neural networks and barycentric Lagrange interpolation. We defer a lengthier discussion of related work to~\Cref{app:related_work}.

\subsection{Physics-Informed Neural Networks} \label{subsec:background_pinns}
Physics-Informed Neural Networks (PINNs)~\citep{raissi2019pinns} propose a flexible and general framework to solve PDEs using neural networks, capable of treating a variety of boundary conditions and geometries. Consider a PDE of the form:
\begin{equation}\label{eqn:pde}
\begin{cases}
    \mathcal{F}(u,x) = 0, & x \in \Omega_{\text{PDE}} \\
    u(x) = u_0(x), & x \in \Omega_{\text{IBC}}
\end{cases}
\end{equation}
where $\mathcal{F}$ is a differential operator, $\Omega_{\text{PDE}}$ is the domain, and $\Omega_{\text{IBC}} \subset \Omega_{\text{PDE}}$ denotes the initial condition region.
The PINN framework represents the solution to~\Cref{eqn:pde} as a parametric model $u_\theta$ and formulates a composite loss function combining a physics term and a boundary term:
\begin{subequations} \label{eqn:physicsNNet}
\begin{align}
    \mathcal{L}(u) &= \mathcal{L}_{\text{PDE}}(u) + \lambda_{\text{IBC}} \mathcal{L}_{\text{IBC}}(u), \\
    \mathcal{L}_{\text{PDE}}(u) &= \mathbb{E}_{x \in \Omega_{\text{PDE}}}\left[(\mathcal{F}(u,x))^2\right], \\
    \mathcal{L}_{\text{IBC}}(u) &= \mathbb{E}_{x \in \Omega_{\text{IBC}}} \left[(u(x) - u_0(x))^2\right]
\end{align}
\end{subequations}

This framework requires the following operations from its model class: (i) \textbf{Evaluation}, computing $u_\theta(x)$ for any $x \in \Omega_{\text{PDE}}$, and (ii) \textbf{Differentiation}, computing partial derivatives $\partial^{(k)} u_\theta/{\partial x_i^{(k)}}(x)$ for any $x \in \Omega_{\text{PDE}}$.
In order to leverage auto-differentiation during optimization, both operations must be differentiable with respect to model parameters $\theta$. Although any models satisfying these properties can be used for physics-informed learning (\textit{e.g.} Gaussian processes~\citep{Raissi_2017}),
recent work focuses on neural networks~\citep{raissi2019pinns}.

\subsection{Barycentric interpolants and spectral methods} \label{subsec:background_interpolation}

Barycentric Lagrange interpolation is a classical technique for polynomial-based function approximation, specified entirely by the function’s values at a set of interpolation nodes~\cite{berrut2004barycentric}.

\begin{definition}
    Given $N+1$ distinct nodes $\{x_j\}$ and values $f(x_j)$, the \emph{barycentric Lagrange interpolant} is:\footnotemark
    \begin{equation}
    \label{eqn:barycentric_lagrange}
        p_N(x) = \frac{\sum_{j=0}^N \frac{w_j}{x - x_j} f(x_j)}{\sum_{j=0}^N \frac{w_j}{x - x_j}},
    \end{equation}
    where $\{w_j\}$ are \emph{barycentric weights}: $w_j = 1/\left({\prod_{k \neq j} (x_j - x_k)}\right)$.
    \footnotetext{Although~\Cref{eqn:barycentric_lagrange} is a rational function with poles at the interpolation nodes, the barycentric form is numerically stable even for large $N$, and avoids the catastrophic cancellation associated with the standard Lagrange formula~\citep{berrut2004barycentric, battles2004extension}.}
\end{definition}

Derivatives can be efficiently computed using differentiation matrices or FFT-based methods, depending on the node distribution~\citep{trefethen2000spectral}. See~\Cref{sec:method} and~\Cref{app:method} for more details.

For well-chosen nodes (\textit{e.g.} Chebyshev-distributed~\citep{trefethen2013approximation}) and smooth functions, the resulting interpolants exhibit spectral convergence -- \emph{exponentially decaying error} for the function and its derivatives.
 
\begin{theorem}[Chebyshev interpolants exhibit spectral convergence~\citep{trefethen2013approximation, boyd1989chebyshev}]
\label{thm:spectral_convergence}
Let $f : [-1, 1] \to \mathbb{R}$ extend to an analytic function on a Bernstein ellipse $E_\rho$ with foci at $\pm 1$ and sum of semiaxes $\rho > 1$. Let $p_n$ be the degree-$n$ Chebyshev interpolant of $f$. Then:
\[
\|f - p_n\|_\infty \leq \frac{4M \rho^{-n}}{\rho - 1}, \qquad
\|f^{(k)} - p_n^{(k)}\|_\infty \leq \frac{C_k M \rho^{-n}}{(\rho - 1)^{k+1}},
\]
for some constant $C_k$ depending on $k$ and $\rho$, where $M = \max_{z \in E_\rho} |f(z)|$.
\end{theorem}

Barycentric interpolation forms the foundation of classical pseudo-spectral methods for solving numerical PDEs~\citep{boyd1989chebyshev, trefethen2000spectral, canuto2006spectral}, where function values on a fixed grid are used to compute derivatives spectrally. This approach underlies well-established numerical solvers (e.g., Chebfun~\citep{battles2004extension}), and provides a principled framework for high-precision computation. Our method, \methodname, builds on this line of work, adapting it for use with gradient-based optimization and machine learning.

%% file: sections/interpolation.tex
\section{Neural networks struggle with precise 1-D interpolation}
\label{sec:interpolation}

\begin{figure}
    \centering
    \includegraphics[width=0.99\linewidth]{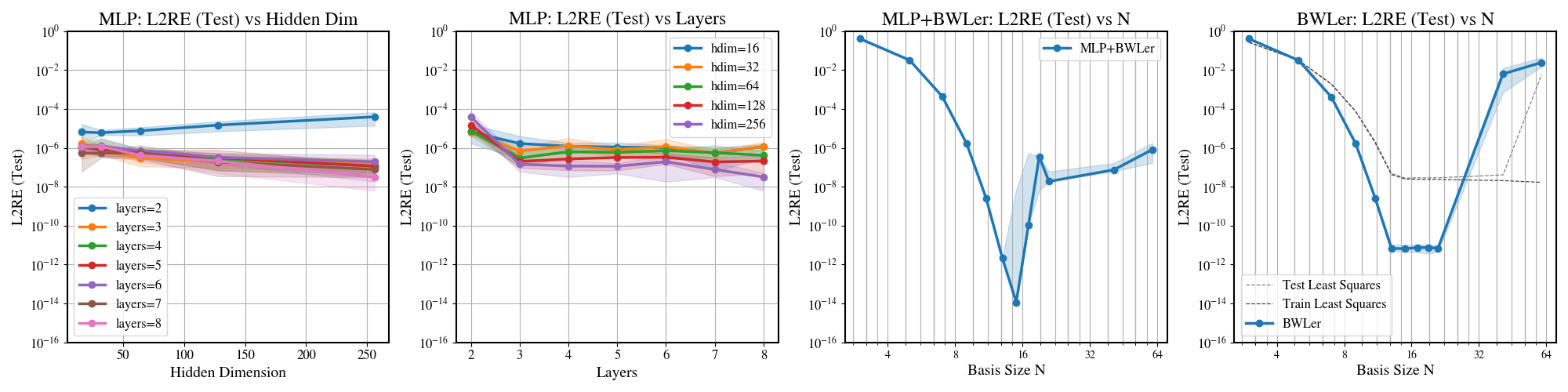}
    \caption{Left: MLPs struggle to interpolate 1-D functions beyond $10^{-8}$ MSE (pictured: $f(x) = \sin(4x)$), even as we scale model width and depth. Right: \methodname~achieves spectral accuracy ($10^{-12}$ MSE); \methodname-hat~improves MLP's MSE by more than $100,000\times$. Least squares Chebyshev interpolation (fit on train, evaluated on test) is also reported (right).}
    \label{fig:1d_interpolation_scaling}
\end{figure}

To disentangle the effects of PDE conditioning from model parameterization, we begin with a simplified setting: one-dimensional smooth function interpolation. This lets us isolate the approximation behavior of different model classes in a well-conditioned regime. Surprisingly, we find that MLPs consistently plateau in precision and scale poorly with larger models (\Cref{subsec:interpolation_mlp}). In contrast, polynomial interpolation admits a complete theoretical analysis and provably converges to (near)-machine precision (\Cref{subsec:interpolation_spectral}).

\subsection{Experimental setup}
\label{subsec:interpolation_setup}

We study the task of one-dimensional function interpolation on the domain $[-1, 1]$, isolating approximation behavior from PDE constraints. We evaluate on sinusoids of the form $\sin(kx)$ for frequencies $k \in \{1,2,4,8,16,32\}$.
See~\Cref{subapp:expt_interpolation} for more details.

For each target function, we generate a training set of $N_{\text{train}}=100$ points sampled uniformly at random from the domain, and evaluate performance on a dense test grid of $N_{\text{test}} = 1000$ points. We report relative $\ell_2$ error (L2RE) on the test grid (\Cref{app:method}).

\subsection{MLPs exhibit precision bottlenecks}
\label{subsec:interpolation_mlp}

We use a fully-connected MLP with $\tanh$ activations, a standard architecture in prior work on PINNs \cite{karniadakis2021physics, hao2023pinnaclecomprehensivebenchmarkphysicsinformed, wang2023expertsguidetrainingphysicsinformed}. Given a set of training points $\{(x_i, f(x_i))\}_{i=1}^{N_{\text{train}}}$, the model is trained to minimize the mean squared error (MSE).
We use the Adam optimizer~\citep{kingma2014adam} with a learning rate of $10^{-3}$ and a cosine decay schedule. To study the effect of model capacity, we sweep network widths within $\{2^4, \dots, 2^8\}$ and depths from $2$--$8$ layers ($16\times$ and $4\times$ ranges, respectively). We also sweep across different levels of function smoothness.

\Cref{fig:1d_interpolation_scaling} (left) shows representative results for $f(x)=\sin(4x)$; comprehensive results are provided in~\Cref{subapp:expt_interpolation}. We find that MLPs consistently stagnate well above machine epsilon -- for the function shown, relative error plateaus around $10^{-8}$ -- $8$ orders of magnitude worse than \texttt{float64}'s machine precision of $2.22\times10^{-16}$. Moreover, precision scales poorly with model size: even when increasing the number of parameters by over $1000\times$ ($400,000$ parameters for the largest MLP we consider), we observe only a $10$--$100\times$ improvement in accuracy. These results suggest that the MLP architecture itself imposes a fundamental bottleneck on achievable precision.

\subsection{Polynomials achieve exponential convergence}
\label{subsec:interpolation_spectral}

In our experiments, we use an $N$-element Chebyshev polynomial basis and solve for the optimal coefficients via least squares on the training set. This reduces to solving a linear system $A \mathbf{c} = \mathbf{f}$, where $A$ is the matrix of Chebyshev basis functions evaluated at the training nodes, and $\mathbf{f}$ contains the target function values at those nodes. More details about the polynomial baseline are in~\Cref{subsubapp:expt_interpolation_chebls}.

Figure~\ref{fig:1d_interpolation_scaling} (right, dotted) shows the empirical error decay of polynomial interpolation on the same target used in the MLP experiment. As predicted by theory (\Cref{thm:spectral_convergence}), the error decays exponentially in $N$. In particular, for the optimal basis size, the polynomial baseline achieves relative errors near machine epsilon -- up to $10,000\times$ better L2RE than the MLP with just $20$--$50$ basis functions. These results motivate the \methodname~architecture we introduce in the next section.

%% file: sections/method.tex
\section{\methodname: a simple baseline using barycentric interpolants}
\label{sec:method}

\methodname~proposes \emph{barycentric interpolants} as a drop-in replacement for MLPs in the physics-informed framework of PINNs. For clarity, we present our method in the 1-D setting over the domain $\Omega = [-1, 1]$, though the approach directly generalizes to periodic domains (using trigonometric instead of Chebyshev polynomials) and higher dimensions (via tensor products). See~\Cref{app:method} for details.

\subsection{Model parameterization} \label{subsec:method_parameterization}
Let $\{x_j\}_{j=0}^N$ denote the Chebyshev nodes of the second kind (Chebyshev-Gauss-Lobatto points~\citep{trefethen2000spectral}) defined by $x_j = \cos\left(j \pi / N\right), j = 0, \dots, N$.
Our model is defined as the unique polynomial $f_\theta$ that interpolates the points
$$(x_0, f_\theta(x_0)), \, \dots, \, (x_N, f_\theta(x_N)),$$
where we specify our model via the values $\{f_\theta(x_j)\}$ it takes on the discrete set of Chebyshev nodes\footnotemark
\footnotetext{\methodname~effectively parameterizes the Lagrange interpolant in \emph{value space}, i.e., directly in terms of the function values at interpolation nodes, rather than in \emph{coefficient space} as in classical polynomial bases. This avoids the instability and ill-conditioning associated with solving for global polynomial coefficients~\citep{berrut2004barycentric}.}.

We consider two possible parameterizations of these node values:
\begin{itemize}[leftmargin=*]
    \item \textbf{Explicit}. Each value is treated as its own, independently trainable parameter, meaning the full set of trainable parameters in the model is $\theta = [\theta_0, \dots, \theta_N]^\top$, where $\theta_j:=f(x_j)$.
    \item \textbf{Implicit}. Like standard PINNs, these define an MLP that specifies function values at discrete node locations. Unlike standard PINNs, the MLP is only evaluated at these node locations, and barycentric interpolation is used to define function values over the full domain. We also term this a \textbf{\methodname-hatted MLP}.
\end{itemize}
\begin{center}
\begin{minipage}[t]{0.4\linewidth}
\begin{algorithm}[H]
\caption*{Standard PINN}
\begin{algorithmic}[1]
\Function{Evaluate}{$x$, $\theta$}
    \State \Return $MLP_\theta(x)$ \Comment{forward pass on $x$}
\EndFunction
\Function{Differentiate}{$x$, $\theta$, $k$}
    \State \Return $\frac{\partial^k u_\theta(x)}{\partial x^k}$  \Comment{autodiff}
\EndFunction
\end{algorithmic}
\end{algorithm}
\end{minipage}
\hfill
\begin{minipage}[t]{0.58\linewidth}
\begin{algorithm}[H]
\caption*{\methodname-hatted PINN}
\begin{algorithmic}[1]
\Function{Evaluate}{$x$, $\theta$}
    \State $f_j \gets \mathrm{MLP}_\theta(\textcolor{diff}{x_j})$ \Comment{forward pass on grid}
    \State \Return \textcolor{diff}{$\mathrm{BaryInterp}(x,\{f_j\})$ \, (\Cref{alg:bary_interp})}
\EndFunction
\Function{Differentiate}{$x$, $\theta$, $k$}
    \State $f_j \gets \mathrm{MLP}_\theta(\textcolor{diff}{x_j})$ \Comment{forward pass on grid}
    \State \textcolor{diff}{$d^{(k)}_j \gets \mathrm{SpectralDeriv}(\{f_j\},k)$ \, (\Cref{alg:cheb_fft})}
    \State \Return \textcolor{diff}{$\mathrm{BaryInterp}(x,\{d^{(k)}_j\})$ \, (\Cref{alg:bary_interp})}
\EndFunction
\end{algorithmic}
\end{algorithm}
\end{minipage}
\captionof{algorithm}{Evaluation and differentiation operations for a standard physics-informed neural network (left) versus a \methodname-hatted neural network (right) -- \textcolor{diff}{differences in blue}.}
\label{alg:pinn_compare}
\end{center}
As required for the physics-informed framework (see~\Cref{subsec:background_pinns}), our model has efficient and auto-differentiable implementations of both \emph{evaluation} and \emph{differentiation} operations:

\paragraph{Evaluation.}
Given the node values $\{f_\theta(x_j)\}_{j=0}^N$, we compute the interpolant $f_\theta(x)$ at any point $x \in \Omega$ using the barycentric formula,~\Cref{eqn:barycentric_lagrange},
where the barycentric weights for Chebyshev–Gauss–Lobatto nodes are \( w_j = ((-1)^j) \, / \, (1 + \delta_{j0} + \delta_{jN}) \).

\paragraph{Differentiation.}
Derivatives are computed efficiently via the Discrete Cosine Transform (DCT). Given node values $\{f_\theta(x_j)\}_{j=0}^N$, the differentiation operation involves transforming to frequency space via DCT, applying differentiation in frequency space (a diagonal operator), and transforming back to physical space via inverse DCT~\citep{boyd1989chebyshev}.
This yields the derivative values $\mathbf{f}' = [f'(x_0), \, \ldots, \, f'(x_N)]^\top$ at the Chebyshev nodes in $O(N \log N)$ operations. To compute the derivative at any point $x \in \Omega$, we apply the barycentric interpolation formula (\Cref{eqn:barycentric_lagrange}), plugging in the derivative $\mathbf{f}'$ as the node values. Higher-order derivatives can be obtained by repeating this process. Detailed descriptions of the evaluation and differentiation operations are provided in~\Cref{app:method}, including pseudocode implementations (Algorithms~\ref{alg:bary_interp},~\ref{alg:cheb_fft}).

\input{sections/interpolation_bwler}

%% file: sections/interpolation_bwler.tex
\subsection{\methodname~achieves exponential convergence on interpolation}
In \Cref{fig:1d_interpolation_scaling}~and \Cref{subapp:expt_interpolation}, we empirically evaluate \methodname~in the 1-D interpolation setting, comparing both explicit \methodname~models and \methodname-hatted MLPs (where \methodname~acts as a final layer atop a standard MLP).
We find that explicit \methodname, trained with Adam, closely follows the exponential convergence behavior of the polynomial baseline from~\Cref{subsec:interpolation_spectral}. Moreover, with proper choice of $N$, \methodname-hatted MLPs substantially outperform standard MLPs on smooth targets, improving L2RE by up to $100,000\times$ (\Cref{subsubapp:expt_interpolation_mlp}).

In this setting, we can fully characterize the error convergence of \methodname; we prove that explicit \methodname~achieves exponentially decaying test error and convergence under gradient descent:

\begin{theorem}[Exponential convergence of \methodname~on interpolation, informal]
\label{thm:bwler_interpolation_informal}
We approximate an analytic function \(f\) by fitting an \((N+1)\)-parameter \methodname~for \(t\) steps of gradient descent on \(M\) sample points in \([-1,1]\). Then its sup-norm error decomposes into
\begin{equation} \label{eqn:decomp_interpolation}
\|f - f_N^{(t)}\|_\infty 
\;\le\;
\underbrace{O(\rho^{-N})}_{\text{expressivity gap}}
\;+\;
\underbrace{\widetilde{O}\left(e^{-\,t/\kappa^2}\right)}_{\text{optimization gap}},
\end{equation}
where
\(\rho>1\) only depends on the function smoothness,
% \(C\) is a fixed GD‐setup constant,
and \(\kappa\) is the interpolation matrix’s condition number.
\end{theorem}

In~\Cref{eqn:decomp_interpolation}, the expressivity gap is unavoidable and comes from the standard Cheybshev approximation bound (\Cref{thm:spectral_convergence}) and the optimization gap comes from gradient descent's convergence on least squares~\citep{boyd2004convex}. Intuitively, after we choose $N$ large enough that \methodname~can express the target function up to error $\epsilon$, gradient descent will then converge exponentially to it in $O(\log(1/\epsilon))$ steps.
See~\Cref{thm:bwler_interpolation_formal} for the precise theorem statement and~\Cref{subapp:theory_interpolation} for the proof.

%% file: sections/baselines.tex
%%%%%%%%%%%%
% 5.1 Implicit
%%%%%%%%%%%%

\section{Physics-informed \methodname~and the precision-conditioning tradeoff}
\label{sec:results}

We evaluate \methodname, first implicit (\methodname-hatted MLP), then explicit, on benchmark PDE problems~\citep{rathore2024challengestrainingpinnsloss, hao2023pinnaclecomprehensivebenchmarkphysicsinformed}, including linear (convection, wave), nonlinear (reaction), stiff (Burgers'), and irregular-domain problems (Poisson).
In doing so, we aim to disentangle the effects of model architecture and PDE conditioning on precision and optimization behavior.
Our key result is~\Cref{thm:bwler_pde_informal}, where we fully characterize the tradeoffs between maximum achievable precision and training convergence for explicit \methodname~in the linear PDE setting.

\subsection{\methodname-hatted MLPs have smoother loss landscapes} \label{subsec:results_mlpinterp}

\paragraph{Experimental setup.}
We start by evaluating \methodname-hatted MLPs on the three benchmark PDEs from~\citet{rathore2024challengestrainingpinnsloss}: convection, reaction, and wave equations. Each model is trained using the Adam optimizer with identical network architecture and training settings. We compare three variants: a standard MLP, an (implicit) \methodname-hatted MLP, and an explicit \methodname~model.

To probe convergence behavior beyond early training dynamics, we train each model for $10^6$ iterations -- significantly longer than prior work -- to examine both the final precision after saturation and the consistency of convergence trends. Full experimental details are provided in \Cref{subapp:expt_pdes}.

\paragraph{Results.}
\Cref{tab:pdes_mlp_vs_mlpinterp_vs_interp} reports final $\ell_2$ relative errors (L2RE) across all methods. \methodname-hatted MLPs consistently outperform standard MLPs, improving L2RE by around $30\times$ on the convection equation, $10\times$ on reaction, and $1800\times$ on wave.
We replicate findings from prior work~\citep{krishnapriyan2021characterizing, rathore2024challengestrainingpinnsloss} that pure MLPs often converge to suboptimal local minima when trained with Adam alone. For instance, on the convection equation (\Cref{fig:pde_convection_method_comparison}), the baseline MLP only recovers a single oscillation of the ground truth periodic solution; similarly, on the wave equation (\Cref{fig:pde_wave_method_comparison}), the MLP recovers the high-level structure of the solution but not the fine-grained details.
In contrast, the \methodname-hatted model finds a solution qualitatively matching the ground truth, and precision improves consistently with Adam alone (\Cref{fig:pde_wave_method_comparison}). 

\paragraph{Improved loss landscape conditioning.}
Towards understanding why \methodname~improves optimization, we estimate the spectral density of the PINN loss's Hessian after convergence, following~\citet{rathore2024challengestrainingpinnsloss}. We find that \methodname~makes the loss landscape less ill-conditioned on the wave and reaction equations, reducing the maximum eigenvalue by $10\times$ and mean eigenvalue by $5$--$10\times$ (Figures~\ref{fig:pde_reaction_method_comparison_spectrum}, ~\ref{fig:pde_wave_method_comparison_spectrum}). See~\Cref{subapp:expt_pdes} for discussion and ablations.

\paragraph{Precision limitations of \methodname-hatted MLPs.}
Although \methodname-hatted MLPs outperform standard PINNs, their precision nonetheless plateaus more than $10$ orders of magnitude L2RE worse than machine precision (\Cref{tab:pdes_mlp_vs_mlpinterp_vs_interp}) -- mirroring the limitations seen in the interpolation setting (\Cref{subsec:interpolation_mlp}).
We attribute this stagnation to the underlying MLP parameterization.
In~\Cref{subsec:results_interp}, we show that switching to an explicit representation and training with a preconditioned second-order method allows \methodname~to overcome this barrier, achieving high precision solutions on the convection and wave equations.

\begin{table}[h]
\centering
\renewcommand{\arraystretch}{1.2}
\begin{tabular}{|c||c|c|c|}
    \hline
    \textbf{L2RE} ↓ & MLP & \methodname-hatted MLP & \methodname \\
    \hline
    \hline
    Convection & $1.14 \times 10^{0}$ 
               & $3.91 \times 10^{-2}$ {\scriptsize ($29.2\times$)} 
               & $\mathbf{4.07 \times 10^{-4}}$ {\scriptsize ($\mathbf{2800\times}$)} \\
    \hline
    Reaction   & $4.02 \times 10^{-3}$ 
               & $\mathbf{3.91 \times 10^{-4}}$ {\scriptsize ($\mathbf{10.3\times}$)} 
               & $7.10 \times 10^{-2}$ {\scriptsize ($0.057\times$)} \\
    \hline
    Wave       & $5.22 \times 10^{-1}$ 
               & $\mathbf{2.88 \times 10^{-4}}$ {\scriptsize ($\mathbf{1800\times}$)} 
               & $9.99 \times 10^{-1}$ {\scriptsize ($0.52\times$)} \\
    \hline
\end{tabular}
\vspace{0.3cm}
\caption{L2 relative errors (L2RE) on benchmark PDEs: convection, reaction, and wave equations from~\cite{rathore2024challengestrainingpinnsloss}. Multiplicative improvements (in parentheses) are relative to the MLP baseline. All models are trained with Adam for $10^6$ iterations.}
\label{tab:pdes_mlp_vs_mlpinterp_vs_interp}
\end{table}

%%%%%%%%%%%%
% 5.2 Explicit
%%%%%%%%%%%%

\subsection{Explicit \methodname~solves PDEs to high precision} \label{subsec:results_interp}
Motivated by the precision limitations in the \methodname-hatted MLP setting, we next study explicit \methodname~models.
This formulation eliminates the extra precision bottlenecks introduced by the neural network parameterization -- all ill-conditioning in the loss arises purely from the PDE and its discretization -- allowing us to probe the precision limits of the physics-informed framework.

\begin{wrapfigure}{r}{0.7\textwidth} \label{fig:fd_precision_conditioning}
  \centering
  \vspace{-10pt} % adjust vertical space if needed
  \includegraphics[width=0.7\textwidth]{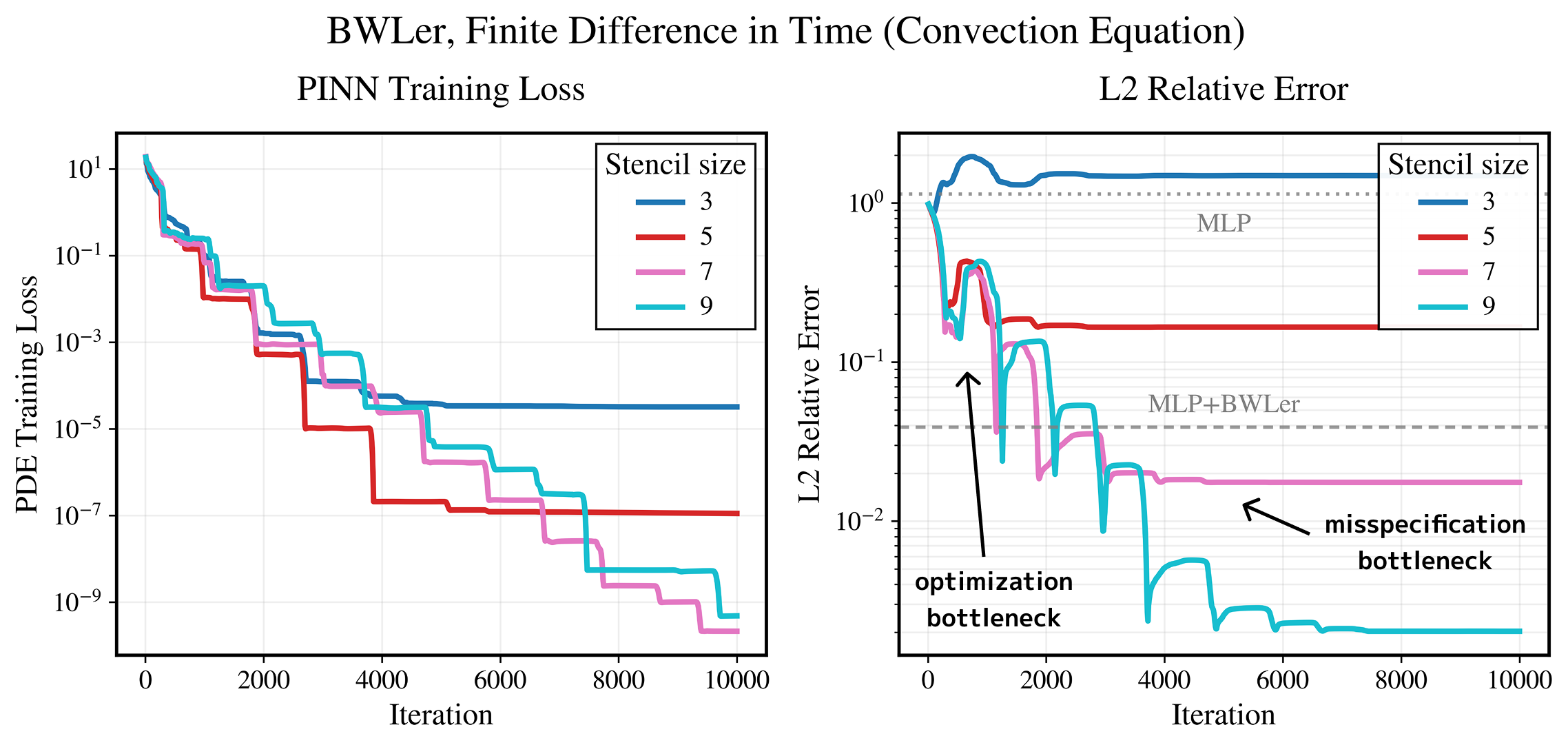}
  \caption{Precision-conditioning tradeoff. We train explicit \methodname~models on the convection equation, using finite-difference derivatives in time, and vary the stencil size. Smaller stencils improve the problem's conditioning, improving initial convergence rate, but increase misspecification error, producing a precision saturation threshold.}
  \vspace{-10pt}
\end{wrapfigure}

\paragraph{Optimization challenges.}
We begin by training explicit \methodname~models with Adam alone on the three benchmark PDEs from~\citet{rathore2024challengestrainingpinnsloss} (\Cref{tab:pdes_mlp_vs_mlpinterp_vs_interp}, rightmost column).
While the models are expressive enough to precisely represent the true solution, they converge slowly under standard first-order optimizers.
On the reaction equation, explicit \methodname~underperforms even standard PINNs by a factor of $20\times$ in L2RE, and makes almost no progress during training on the wave equation. Note that although explicit \methodname~with Adam outperforms standard PINNs by $2800\times$ and \methodname-hatted MLPs by $100\times$ on the convection equation, this is mostly due to the extremely high number of training steps ($10^6$ iterations with Adam). See~\Cref{subapp:expt_pdes} for more detailed results.

\paragraph{Theory: convergence-conditioning tradeoff for 1-D linear differential operators.}
For explicit \methodname, the PDE setting admits an error decomposition mirroring the interpolation setting of~\Cref{thm:bwler_interpolation_informal}. We present the 1-D linear setting, but note that the decomposition extends directly to higher-dimensional linear problems.

\begin{theorem}[Precision-conditioning tradeoff for physics-informed \methodname, informal]
\label{thm:bwler_pde_informal}
We consider solving the $d$-th order PDE problem 
\(
L\,u = 0,
\)
where \(u\) satisfies the usual analyticity assumptions, by approximating \(L\) with a \(k\)-th order finite‐difference scheme. Fitting an \((N+1)\)-parameter \methodname~via \(t\) steps of gradient descent on this discretized operator yields
\begin{equation}
\label{eqn:decomp_pde}
\|u - u_N^{(t)}\|_\infty 
\;\le\;
\underbrace{O(\rho^{-N})}_{\text{expressivity gap}}
\;+\;
\underbrace{\widetilde{O}\!\bigl(N^{-(k+1-d)}\bigr)}_{\text{bias/misspecification gap}}
\;+\;
\underbrace{\widetilde{O}\left(e^{-\,t/\kappa(N)^2}\right)}_{\text{optimization gap}},
\end{equation}
where \(\rho>1\) only depends on the function smoothness,
% \(C\) is a fixed GD‐setup constant,
and \(\kappa(N)\) is the condition number of the discretized operator.
\end{theorem}

In~\Cref{eqn:decomp_pde}, the expressivity gap is the standard Cheybshev approximation bound (\Cref{thm:spectral_convergence}), the misspecification gap comes from the order of the finite-difference approximation~\citep{fornberg1988generation}, and the optimization gap is the standard gradient descent convergence rate on least squares~\citep{boyd2004convex}. Refer to~\Cref{thm:bwler_pde_formal} and~\Cref{app:theory} for a formal theorem statement and proof.

Two precision-conditioning tradeoffs directly emerge from~\Cref{thm:bwler_pde_informal}:
\begin{itemize}[leftmargin=*]
    \item \textbf{Expressivity vs. optimization.}
    The conditioning of order-$d$ derivatives with spectral differentiation scales as $O(N^{2d})$~\citep{trefethen2000spectral}.
    Decreasing the expressivity gap relies on increasing $N$, but this necessarily worsens the problem's conditioning and convergence rate.
    \item \textbf{Misspecification vs. optimization.}
    One way to improve the problem's conditioning is to try alternate derivative formulations, e.g. finite-difference schemes instead of spectral differentiation.
    Although FD matrices are better-conditioned (for 3-point stencils, $O(N^{d})$ instead of $O(N^{2d})$~\citep{fornberg1988generation}), the misspecification gap increases in turn.
\end{itemize}

We note that a similar decomposition can be stated for nonlinear problems with standard PINNs, but a precise analysis of the precision-conditioning tradeoff is challenging~\citep{karniadakis2021physics, mishra2023estimates}.

\paragraph{Training techniques towards efficient, high precision training.}
To navigate the precision-conditioning tradeoff, we combine the following three techniques:
\begin{itemize}[leftmargin=*]

    \item \textbf{Nystr\"{o}m-Newton-CG (NNCG)}~\citep{rathore2024challengestrainingpinnsloss}. NNCG is a second-order method that approximates the Newton step using a low-rank Nystr\"{o}m approximation to the Hessian. We tune the preconditioner rank and the number of CG iterations per Newton step to control the convergence rate.
    
    \item \textbf{Derivative quality tuning}. \methodname~allows us to freely swap out different derivative computation methods. We experiment with spectral derivatives and finite differences where we vary the stencil size (\textit{e.g.} 3-point stencil yields a 1st-order approximation, while a global stencil recovers the spectral derivative). See~\Cref{app:method} for implementation details.
    
    \item \textbf{Multi-stage training.} Since \methodname~is parameterized directly via its values on a discrete grid, we can warm-start training using any pretrained PINN (including another \methodname~or a standard MLP).
    
\end{itemize}

\paragraph{High precision solutions to benchmark PDEs.}

\Cref{tab:pdes_sota} reports final $\ell_2$ relative errors (L2RE) across five benchmark PDEs. On the convection, reaction, and wave equations, explicit \methodname~achieves (near-)machine-precision solutions: $8$--$10$ orders of magnitude improvements relative to the L2RE of PINNs reported in the literature.
On Burgers' and an irregular-geometry Poisson problem, explicit \methodname~matches the precision of prior state-of-the-art PINNs.
Appendix~\ref{subapp:expt_pdes} provides implementation details, training diagnostics, and problem-specific strategies used to achieve these results.

We note that our results are \emph{not} time- or parameter-matched: see~\Cref{subapp:expt_pdes} for details. In particular, \methodname~inherits the weaknesses of classical spectral methods, including handling PDEs with sharp solutions or problems over irregular domains; on the Burgers' and the Poisson problems, we need to train for much longer (up to $15\times$) to match standard MLPs. We view our results as a proof-of-concept: they show that machine-precision solutions are in fact possible within the PINN framework, but high precision requires careful codesign of models and optimizers, with the precision-conditioning tradeoff in mind.
We also note that existing training techniques (\textit{e.g.} NTK loss reweighting, temporal causality~\citep{wang2023expertsguidetrainingphysicsinformed}) are complementary to our approach and likely could be leveraged to speed up training.

\begin{table}[h!]
\centering
\renewcommand{\arraystretch}{1.2}
\begin{tabular}{|c||c|c|}
    \hline
    \textbf{L2RE} ↓ & MLP (from literature) & Explicit \methodname \\
    \hline
    \hline
    Convection ($c=40$) & $1.94 \times 10^{-3}$~\citep{rathore2024challengestrainingpinnsloss}
              & $\mathbf{2.04 \times 10^{-13}}$ {\scriptsize ($9.51 \times 10^{9}\times$)} \\
    \hline
    Convection ($c=80$) & $6.88 \times 10^{-4}$~\citep{wang2023expertsguidetrainingphysicsinformed}
              & $\mathbf{1.10 \times 10^{-12}}$ {\scriptsize ($6.25 \times 10^{8}\times$)} \\
    \hline
    Wave      & $1.27 \times 10^{-2}$~\citep{rathore2024challengestrainingpinnsloss}
              & $\mathbf{1.26 \times 10^{-11}}$ {\scriptsize ($1.00 \times 10^{9}\times$)} \\
    \hline
    Reaction  & $9.92 \times 10^{-3}$~\citep{rathore2024challengestrainingpinnsloss}
              & $\mathbf{6.94 \times 10^{-11}}$ {\scriptsize ($1.43 \times 10^{8}\times$)} \\
    \hline
    \hline
    Burgers (1D-C)  & $1.33 \times 10^{-2}$~\citep{hao2023pinnaclecomprehensivebenchmarkphysicsinformed} 
              & $\mathbf{4.63 \times 10^{-3}}$ {\scriptsize ($2.87\times$)} \\
    \hline
    \hline
    Poisson (2D-C)   & $1.23 \times 10^{-2}$~\citep{hao2023pinnaclecomprehensivebenchmarkphysicsinformed} 
              & $\mathbf{1.08 \times 10^{-2}}$ {\scriptsize ($1.14\times$)} \\
    \hline
\end{tabular}
\vspace{0.3cm}
\caption{L2 relative errors (L2RE) on benchmark PDEs problems. SOTA column reports (to our knowledge) the best PINN results from the literature. Note: results are \emph{not} time- or parameter-matched.}
\label{tab:pdes_sota}
\end{table}

%% file: sections/discussion.tex
\section{Conclusion} \label{sec:discussion}

\paragraph{Discussion.} Our results demonstrate that incorporating barycentric interpolants into the PINN framework dramatically improves attainable precision while maintaining the flexibility to handle diverse PDEs and complex geometries. On 1-D interpolation tasks, explicit \methodname~models recover spectral convergence, reaching relative errors near $10^{-12}$. \methodname-hatted MLPs, our drop-in variant, similarly boost precision by up to $10,000\times$ over standard MLPs (\Cref{fig:1d_interpolation_scaling}). On PDE benchmarks, \methodname-hatted MLPs boost the precision of standard PINNs by up to $1800\times$ (\Cref{tab:pdes_mlp_vs_mlpinterp_vs_interp}). Using a second-order optimizer, we reach near-machine precision for convection, reaction, and wave equations (between $10^{-13}$--$10^{-11}$), $8$--$10$ orders of magnitude better than prior state-of-the-art. To our knowledge, this is the first instance of a PINN reaching machine-precision solutions even on 2-D problems.

\paragraph{Limitations.}Despite these gains in precision, several limitations remain. First, the runtime cost of training is substantial. Because of their ill-conditioned loss landscapes, explicit \methodname s require significantly longer to train than both traditional numerical solvers and prior PINN architectures. Although our results establish a new precision ceiling for PINNs, they do not yet outperform classic numerical methods in terms of precision per unit time. Second, \methodname~thrives on PDEs with smooth solutions but performance deteriorates on stiff PDEs with sharp features or on irregular domains, settings where spectral methods traditionally struggle. Finally, our use of explicit grids may pose scalability challenges in higher-dimensional problems, where mesh-free methods often hold an advantage. 

%% file: sections/acknowledgements.tex
\section*{Acknowledgements}

The authors thank Owen Dugan, Sabri Eyuboglu, Roberto Garcia, William Gilpin, Kade Heckel, Junmiao Hu, Jeffrey Lai, Pratik Rathore, Benjamin Spector, Ben Viggiano, and Michael Zhang for their helpful feedback and discussion.

The authors gratefully acknowledge the support of NIH under No. U54EB020405 (Mobilize), NSF under Nos. CCF2247015 (Hardware-Aware), CCF1763315 (Beyond Sparsity), CCF1563078 (Volume to Velocity), and 1937301 (RTML); US DEVCOM ARL under Nos. W911NF-23-2-0184 (Long-context) and W911NF-21-2-0251 (Interactive Human-AI Teaming); ONR under Nos. N000142312633 (Deep Signal Processing); Stanford HAI under No. 247183; NXP, Xilinx, LETI-CEA, Intel, IBM, Microsoft, NEC, Toshiba, TSMC, ARM, Hitachi, BASF, Accenture, Ericsson, Qualcomm, Analog Devices, Google Cloud, Salesforce, Total, the HAI-GCP Cloud Credits for Research program,  the Stanford Data Science Initiative (SDSI), and members of the Stanford DAWN project: Meta, Google, and VMWare. The U.S. Government is authorized to reproduce and distribute reprints for Governmental purposes notwithstanding any copyright notation thereon. Any opinions, findings, and conclusions or recommendations expressed in this material are those of the authors and do not necessarily reflect the views, policies, or endorsements, either expressed or implied, of NIH, ONR, or the U.S. Government.
JL is supported by the Department of Energy Computational Science Graduate Fellowship under Award Number DE-SC0023112.
AR's research is supported by NSF grant CCF\#2247014.

%% file: sections/appendix/related_work.tex
\section{Related work} \label{app:related_work}

\subsection{High-precision machine learning for PDEs} \label{subsec:background_precision_ml}

The difficulty of achieving high precision in machine learning for scientific applications is well-documented: despite progress from the scientific ML community in recent years, traditional numerical methods still outperform existing PDE learning approaches in precision, even on simple benchmark problems~\citep{McGreivy_2024}.
We are not aware of any physics-informed neural network approaches that obtain machine-precision solutions, even on standard 2-D linear PDE benchmarks.

Recent work in the PINN literature has explored architectural modifications, addressed optimization challenges, and proposed specialized training strategies~\citep{krishnapriyan2021characterizing, wang2023expertsguidetrainingphysicsinformed, rathore2024challengestrainingpinnsloss}.
The inherent precision limitations of existing ML architectures is comparatively underexplored.
\cite{Michaud_2023, wang2023multistageneuralnetworksfunction} focus on the regression setting using MLPs and propose alternate training recipes towards higher precision.
\cite{liu2025towards} identifies precision bottlenecks resulting from the Transformer architecture and standard LR schedulers in the setting of least squares.

Unlike prior work that focuses primarily on studying either PDE optimization or model architecture, in this paper we attempt to study both jointly:
\begin{itemize}
    \item We demonstrate that the MLP parameterization of standard PINNs limits precision in the simple setting of 1-D function approximation, even without the additional challenges of PDE constraints (\Cref{sec:interpolation}).
    
    \item We propose \methodname, which decouples model parameterization from derivative computation (\Cref{sec:method}). This allows us to separately study the precision limitations induced by the model versus the PDE conditioning.
    
    \item Using \methodname, we detail an explicit tradeoff between precision and conditioning in the linear PDE setting (\Cref{thm:bwler_pde_informal}). Along the way, we provide empirical evidence that our barycentric interpolants represent a simple yet surprisingly effective baseline parameterization for high-precision PDE learning; they achieve the high precision of traditional spectral methods on benchmark PDEs with smooth solutions, while maintaining compatibility with physics-informed frameworks (Tables~\ref{tab:pdes_mlp_vs_mlpinterp_vs_interp},~\ref{tab:pdes_sota}).
\end{itemize}

\subsection{Hybrid approaches: combining PINNs with numerical methods}
\label{subapp:hybrid}

Recent work has explored integrating classical numerical techniques with machine learning-based PDE solvers to improve robustness, accuracy, and convergence. We highlight two relevant approaches:
\begin{itemize}
    \item \textbf{Time-marching with PINNs.} Recent works attempt to embed numerical solvers directly into the training loop of PINNs, most commonly in handling time-dependent PDEs. For example,~\citep{krishnapriyan2021characterizing, wang2023expertsguidetrainingphysicsinformed} propose to divide the time domain into multiple subdomains and perform curriculum learning within a PINN framework to boost performance.
    Another set of approaches~\citep{chen2023implicit, berman2023randomized, chen2024teng} directly incorporate time-stepping via classical integrators (e.g., Runge–Kutta) within a neural network framework to stabilize temporal dynamics, especially for stiff or chaotic systems.
    
    \item \textbf{ODIL.} The ODIL framework~\citep{karnakov2022optimizing, karnakov2024solving} formulates PDE learning as the minimization of discretized residuals over mesh-based domains, preserving the structure and sparsity of finite volume and finite difference discretizations while enabling gradient-based optimization with neural networks.
    
    Like ODIL, our method also reintroduces an explicit grid under the hood of a physics-informed learning framework, but \methodname~differs in two ways. \textbf{(1)} Just as ODIL exactly embeds finite volume and finite difference methods into an auto-differentiable ML setup, \methodname~respectively embeds pseudo-spectral methods into physics-informed learning. This means that unlike the lower-precision PDE discretizations of finite differences, \methodname~leverages the spectral convergence of polynomial approximation on smooth functions. \textbf{(2)} Additionally, \methodname~can be flexibly treated both as a self-standing architecture or as an additional layer that goes atop existing PINN architectures.
\end{itemize}

\subsection{Spectral methods and barycentric interpolation} \label{subapp:spectral}
\label{subsec:method_spectral_connection}

Classical spectral methods, including Chebyshev and Fourier-based solvers, have long been used for high-accuracy PDE solutions on regular domains~\citep{boyd1989chebyshev, trefethen2000spectral}. These methods excel when the solution is smooth and the domain is simple, offering exponential convergence rates in both function and derivative approximation. Spectral element methods~\citep{canuto2006spectral} extend these ideas to complex geometries by combining high-order accuracy with domain decomposition, and remain state-of-the-art in areas like fluid dynamics where both precision and geometric flexibility are critical~\citep{frisch1995turbulence, wilcox2006turbulence}.

Recent frameworks like Chebfun~\citep{battles2004extension} have revived interest in spectral approaches by enabling function-level computation with near-machine precision. Barycentric Lagrange interpolation~\citep{berrut2004barycentric} provides a numerically stable alternative to classical polynomial bases and serves as the foundation for many pseudo-spectral techniques. Our work is, to our knowledge, the first to integrate barycentric interpolation directly into a physics-informed learning framework.

Our approach closely parallels classical pseudo-spectral methods while introducing key flexibilities from the machine learning paradigm. Like traditional pseudo-spectral solvers, we represent the solution via its values at Chebyshev nodes, and we compute high-precision derivatives spectrally. However, \methodname~additionally inherits the generality of the physics-informed paradigm:
\begin{itemize}
    \item \textbf{Modern optimization and auto-differentiation.} \methodname~seamlessly integrates with auto-differentiation frameworks on GPU-accelerated hardware. Instead of using classical iterative solvers, we can leverage ML optimizers such as Adam~\citep{kingma2014adam}, L-BFGS~\citep{liu1989limited}, and NNCG~\citep{rathore2024challengestrainingpinnsloss}.
    
    \item \textbf{Flexible derivative computation.} While classical solvers typically rely on differentiation matrices, \methodname~allows switching between spectral (FFT-based) and finite difference derivatives to match the problem structure. See~\Cref{app:method} for details.
    
    \item \textbf{Support for irregular geometries.} \methodname's barycentric formulation accommodates non-rectangular domains and complex boundary conditions using the least-squares framework of physics-informed learning. This avoids the manual domain transformations that traditional spectral methods need~\citep{boyd1989chebyshev, canuto2006spectral}.
\end{itemize}

%% file: sections/appendix/method.tex
\section{Method} \label{app:method}

In this section, we provide more details about the implementation of the \methodname~architecture and training.

\subsection{\methodname~architecture}
\label{app:bwler-arch}

We provide full algorithmic details of the \methodname~architecture. We first focus on the 1-D case before generalizing to higher dimensions. The core components are:

\begin{itemize}
    \item \textbf{Chebyshev setting (1-D, non-periodic).} We describe barycentric interpolation and spectral differentiation using the Chebyshev-Gauss-Lobatto grid. This forms the foundation for interpolation and differentiation in non-periodic domains.
    
    \item \textbf{Fourier setting (1-D, periodic).} For periodic boundary conditions, we instead use Fourier nodes and basis functions. We describe both interpolation and differentiation with trigonometric polynomial interpolants.
    
    \item \textbf{Finite difference matrices.} As an alternative to spectral differentiation, we optionally use finite difference (FD) methods with Fornberg’s algorithm to generate sparse banded derivative matrices.
    
    \item \textbf{Domain transformation.} All 1-D methods assume canonical domains ($[-1,1]$ for Chebyshev; $[0,2\pi]$ for Fourier), but are extended to arbitrary physical domains via affine coordinate maps.
    
    \item \textbf{Higher-dimensional extension.} We extend all components to multiple dimensions using tensor-product constructions, which factorize evaluation and differentiation along each axis.
\end{itemize}

This appendix provides explicit pseudocode for each of the above settings and highlights the computational properties relevant to their use in PINN frameworks.

\subsubsection{Chebyshev setting, non-periodic}

\paragraph{Evaluation.}
We begin with interpolation on the canonical Chebyshev-Gauss-Lobatto (CGL) grid. We consider interpolating a 1-D function $f : [-1, 1] \to \mathbb{R}$. Let
\[
x_j = \cos\!\Bigl(\tfrac{j\pi}{N}\Bigr),
\quad
w_j = (-1)^j
\begin{cases}
\frac12, & j\in\{0,N\},\\
1, & \text{otherwise},
\end{cases}
\quad j=0,\dots,N.
\]
Then for any query \(x\in[-1,1]\), we recall the barycentric formula (\Cref{eqn:barycentric_lagrange}) gives
\[
f_\theta(x)
=\frac{\displaystyle\sum_{j=0}^N \frac{w_j\,f_j}{x - x_j}}
      {\displaystyle\sum_{j=0}^N \frac{w_j}{x - x_j}}.
\]
\Cref{alg:bary_interp} represents a pseudocode description of the barycentric formula, which is how \methodname~implements the \emph{evaluation} operation required for the physics-informed framework.

\begin{algorithm}[H]
\caption{\textsc{BaryInterp}: Chebyshev barycentric interpolation}
\label{alg:bary_interp}
\begin{algorithmic}[1]
\State \textbf{Input:} \(x\in[-1,1]\); node values \(\{f_j\}_{j=0}^N\)
\State \textbf{Output:} interpolated value \(f_\theta(x)\)
\medskip
\Statex \(\triangleright\) Compute CGL nodes and weights (if not precomputed)
\For{\(j=0\) \textbf{to} \(N\)}
  \State \(x_j \gets \cos\!\bigl(\tfrac{j\pi}{N}\bigr)\)
  \State \(w_j \gets (-1)^j \times \bigl(\tfrac12 \text{ if } j\in\{0,N\}\text{ else }1\bigr)\)
\EndFor
\medskip
\Statex \(\triangleright\) Handle exact node case
\For{\(j=0\) \textbf{to} \(N\)}
  \If{\(x = x_j\)} 
    \State \Return \(f_j\)
  \EndIf
\EndFor
\medskip
\Statex \(\triangleright\) Accumulate barycentric sums
\[
  N_{\mathrm{sum}} \gets \sum_{j=0}^N \frac{w_j\,f_j}{x - x_j},
  \quad
  D_{\mathrm{sum}} \gets \sum_{j=0}^N \frac{w_j}{x - x_j}
\]
\Statex \textbf{Return} \(\displaystyle f_\theta(x)=\frac{N_{\mathrm{sum}}}{D_{\mathrm{sum}}}\)
\end{algorithmic}
\end{algorithm}

\paragraph{Differentiation.}
We compute first-order derivatives at the Chebyshev nodes in \(\mathcal O(N\log N)\) via the FFT. \Cref{alg:cheb_fft} represents a pseudocode implementation of \methodname's \emph{differentiation} operation.

\begin{algorithm}[H]
\caption{\textsc{ChebFFTDerivative}: spectral derivative at CGL nodes}
\label{alg:cheb_fft}
\begin{algorithmic}[1]
\State \textbf{Input:} node values \(\mathbf{u} = (u_0,\dots,u_N)\)
\State \textbf{Output:} derivative \(\mathbf{d} = \bigl(f'(x_0),\dots,f'(x_N)\bigr)\)
\medskip
\Statex \(\triangleright\) Mirror data (even extension)
\[
  V \;\gets\;\bigl[u_0,\,u_1,\,\dots,\,u_N,\,u_{N-1},\,\dots,\,u_1\bigr]
\]
\Statex \(\triangleright\) Forward FFT
\[
  \hat V \;\gets\;\mathrm{FFT}(V)
\]
\medskip
\Statex \(\triangleright\) Differentiate in frequency space
\For{\(k=0\) \textbf{to} \(2N-1\)}
  \State \(k_{\mathrm{eff}} \gets \begin{cases}
    k, & k\le N,\\
    k-2N, & k>N,
  \end{cases}\)
  \State \(\hat W_k \gets i\,k_{\mathrm{eff}}\,\hat V_k\)
\EndFor
\medskip
\Statex \(\triangleright\) Inverse FFT
\[
  W \;\gets\;\mathrm{IFFT}(\hat W)
\]
\medskip
\Statex \(\triangleright\) Chain-rule correction
\For{\(j=1\) \textbf{to} \(N-1\)}
  \State \(d_j \gets -\,W_j \;/\;\sqrt{1 - x_j^2}\)
\EndFor
\State \(d_0 \gets \sum_{n=0}^N n^2\,\hat u_n,\)
\quad
\(d_N \gets \sum_{n=0}^N (-1)^{n+1}n^2\,\hat u_n\)
\medskip
\Statex \textbf{Return} \(\mathbf{d}\)
\end{algorithmic}
\end{algorithm}

\subsubsection{Fourier setting, periodic} 
For periodic domains, we use \(N\) equispaced nodes 
\[
x_j = \frac{2\pi\,j}{N},\quad j=0,\dots,N-1,
\]
instead of Chebyshev-Gauss-Lobatto nodes. We consider interpolating a 1-D function $f : [0, 2\pi] \to \mathbb{R}$. To do so, we represent \(f\) by its discrete Fourier series. 

\paragraph{Evaluation.} For a trigonometric interpolant on equispaced nodes, \emph{evaluation} can be performed using the FFT. We provide a pseudocode implementation in~\Cref{alg:fourier_interp}.

\begin{algorithm}[H]
\caption{\textsc{FourierInterp}: trigonometric interpolation}
\label{alg:fourier_interp}
\begin{algorithmic}[1]
\State \textbf{Input:} query point \(x\in[0,2\pi]\); node values \(\{f_j\}_{j=0}^{N-1}\)
\State \textbf{Output:} interpolated value \(f_\theta(x)\)
\medskip
\Statex \(\triangleright\) Precompute grid
\For{\(j=0\) \textbf{to} \(N-1\)}
  \State \(x_j \gets \tfrac{2\pi\,j}{N}\)
\EndFor
\medskip
\Statex \(\triangleright\) Compute Fourier coefficients
\[
  \hat f \;\gets\;\mathrm{FFT}\bigl(\{f_j\}\bigr)\,/\,N
\]
\medskip
\Statex \(\triangleright\) Evaluate interpolant
\[
  f_\theta(x)
  \;=\;\sum_{k=0}^{N-1} \hat f_k \, e^{\,i\,k\,x}
\]
\Statex \textbf{Return} \(f_\theta(x)\)
\end{algorithmic}
\end{algorithm}

\paragraph{Differentiation.}

To perform differentiation on equispaced nodes, we define an explicit differentiation matrix, following~\citep{trefethen2000spectral}.
Specifically, the Fourier differentiation matrix on \(N\) equispaced points \(x_j = 2\pi j/N\) is defined as:
\begin{equation}
\label{eqn:fourier_diff_matrix}
(D_N)_{ij} \;=\;
\begin{cases}
0, & i = j,\\[6pt]
\displaystyle\frac{(-1)^{\,i-j}}{2}\,\cot\!\Bigl(\tfrac{\pi(i - j)}{N}\Bigr), & i \ne j.
\end{cases}
\end{equation}
We provide a pseudocode implementation of \emph{differentiation} for trigonometric interpolants in~\Cref{alg:fourier_deriv_matrix}.

\begin{algorithm}[H]
\caption{\textsc{FourierDerivativeMatrix}: periodic derivative via explicit matrix}
\label{alg:fourier_deriv_matrix}
\begin{algorithmic}[1]
\State \textbf{Input:} node values \(\mathbf{u}=(u_0,\dots,u_{N-1})\)
\Statex \quad\(\triangleright\) on equispaced grid \(x_j=2\pi j/N\)
\State \textbf{Output:} derivatives \(\mathbf{d}=(f'(x_0),\dots,f'(x_{N-1}))\)
\medskip
\Statex \(\triangleright\) Assemble Fourier differentiation matrix (\Cref{eqn:fourier_diff_matrix})
\For{\(i=0\) \textbf{to} \(N-1\)}
  \For{\(j=0\) \textbf{to} \(N-1\)}
    \If{\(i = j\)}
      \State \((D)_{ij} \gets 0\)
    \Else
      \State \((D)_{ij} \gets \tfrac{(-1)^{\,i-j}}{2}\,\cot\!\bigl(\tfrac{\pi(i-j)}{N}\bigr)\)
    \EndIf
  \EndFor
\EndFor
\medskip
\Statex \(\triangleright\) Apply matrix to values
\[
  \mathbf{d} \;\gets\; D \,\mathbf{u}
\]
\medskip
\Statex \textbf{Return} \(\mathbf{d}\)
\end{algorithmic}
\end{algorithm}

\subsubsection{Finite–difference differentiation matrices}
For arbitrary node distributions \(\{x_j\}_{j=0}^N\), we employ Fornberg’s algorithm~\citep{fornberg1988generation} to construct a sparse, banded matrix \(D^{(m,k)}\in\mathbb R^{(N+1)\times(N+1)}\) that approximates the \(m\)-th derivative using a stencil of half-bandwidth \(k\).  The entries satisfy
\[
(D^{(m,k)}\mathbf{u})_i \;=\;\sum_{j=\max(0,i-k)}^{\min(N,i+k)}
w_{ij}^{(m)}\,u_j,
\]
and yield \(O(h^{2k-m})\) accuracy on non-uniform grids. In the periodic setting, we recover the standard finite difference stencils on equispaced nodes.

The algorithm for performing differentiation using finite difference instead of spectral derivatives is outlined in pseudocode in~\Cref{alg:fd_derivative}.

\begin{algorithm}[H]
\caption{\textsc{FDDerivativeMatrix}: Fornberg finite‐difference derivative}
\label{alg:fd_derivative}
\begin{algorithmic}[1]
\State \textbf{Input:} node locations \(\{x_j\}_{j=0}^N\), values \(\mathbf{u}\in\mathbb R^{N+1}\), derivative order \(m\), half‐bandwidth \(k\)
\State \textbf{Output:} \(\mathbf{d}=(f^{(m)}(x_0),\dots,f^{(m)}(x_N))\)
\medskip
\Statex \(\triangleright\) Build differentiation matrix via Fornberg’s method
\State \(D^{(m,k)} \gets \mathrm{FornbergMatrix}(\{x_j\},\,m,\,k)\)~\cite{fornberg1988generation}
\medskip
\Statex \(\triangleright\) Apply to node values
\State \(\mathbf{d} \gets D^{(m,k)}\,\mathbf{u}\)
\medskip
\Statex \textbf{Return} \(\mathbf{d}\)
\end{algorithmic}
\end{algorithm}

\subsubsection{Domain transformation to arbitrary intervals}
All of the above 1-D formulas assume canonical domains (\([-1,1]\) for Chebyshev, \([0,2\pi]\) for Fourier).  To handle a physical interval \([a,b]\), we apply an affine map \(x\mapsto \tilde x\):
\[
\tilde x = 
\begin{cases}
\dfrac{2\,(x - a)}{b - a} - 1, & \text{Chebyshev},\\[8pt]
\dfrac{2\pi\,(x - a)}{b - a}, & \text{Fourier}.
\end{cases}
\]
All node locations, weights, and differentiation matrices are computed in \(\tilde x\)-space, and final function values or derivatives are re-mapped to the physical coordinate \(x\).  This preserves both the interpolation accuracy and spectral convergence properties on arbitrary intervals. We note that we must account for the rescaling factor when mapping function values to and from the physical and canonical domains.

\subsubsection{Extension to higher dimensions}
Let \(\mathbf x = (x_1,\dots,x_d)\in\Omega\subset\mathbb R^d\).  We construct a tensor-product interpolant:
\[
f(\mathbf x)
=\sum_{j_1=0}^{N_1}\!\cdots\sum_{j_d=0}^{N_d}
f_{j_1,\dots,j_d}
\;\prod_{\ell=1}^d \phi^{(\ell)}_{j_\ell}(x_\ell),
\]
where each \(\phi^{(\ell)}\) is the 1-D Chebyshev or Fourier barycentric basis on the \(\ell\)-th axis.  Evaluation and differentiation factorize along each dimension:
\[
\partial^{k_1}_{x_1}\cdots\partial^{k_d}_{x_d} f(\mathbf x)
=\sum_{j_1,\dots,j_d}
f_{j_1,\dots,j_d}\,
\prod_{\ell=1}^d
\bigl(\phi^{(\ell)}_{j_\ell}\bigr)^{(k_\ell)}(x_\ell).
\]
Thus, in practice, \methodname~applies the 1-D interpolation or derivative operators sequentially (or, for differentiation matrices, via Kronecker-product routines) to achieve efficient interpolation and differentiation in higher dimensions.

%%%%%%%%%%
%
%%%%%%%%%%
\subsection{Training} \label{subsec:method_algorithm}
Our training algorithm consists of two key components: the optimizer for updating model parameters and the scheme for selecting collocation points where PDE constraints are enforced.

\paragraph{Optimizer.}
We experiment with two different optimizers:
\begin{itemize}
    \item \textbf{Adam}~\citep{kingma2014adam}: The standard first-order optimizer in deep learning. By default, we use an initial learning rate of $10^{-3}$ with cosine decay learning rate schedule with a minimum learning rate of $10^{-6}$.
    \item \textbf{Nystr\"{o}m-Newton CG}~\citep{rathore2024challengestrainingpinnsloss}: A specialized second-order method designed for PINNs that approximates the Hessian using Nystr\"{o}m sampling. We use the default hyperparameters from~\citet{rathore2024challengestrainingpinnsloss} except for the rank of the preconditioner and the number of CG steps per Newton update, which we tune per problem. See~\Cref{subapp:expt_pdes} for problem-specific hyperparameters.
\end{itemize}

\paragraph{Collocation scheme.}
For selecting collocation points where the PDE residual is enforced, we explore two strategies:
\begin{itemize}
    \item \textbf{Random sampling.} Following standard PINN practice, we sample collocation points at each iteration. We compare two distributions:
    \begin{itemize}
        \item Uniform sampling on $[-1, 1]$: $x \sim \text{Unif}([-1, 1])$
        \item Chebyshev-weighted sampling: $x = \cos(\theta)$ where $\theta \sim \text{Unif}([0, \pi])$, which has density $\propto 1/\sqrt{1-x^2}$
    \end{itemize}
    We default to the latter as it matches the node distribution in the model parameterization.
    
    \item \textbf{Fixed nodal collocation.} Unlike traditional PINNs which require dense sampling to ensure the PDE holds everywhere, our polynomial representation allows us to enforce the PDE only at the Chebyshev nodes $\{x_j\}_{j=0}^N$.
\end{itemize}
We find that nodal collocation suffices for the benchmark PDE problems we consider in this work. See~\Cref{subapp:expt_pdes} for more details about hyperparameters for specific experiments.

\paragraph{L2 Relative Error Formula.}
For assessing the quality of interpolants and PDE solutions of all models used in this paper we leverage the standard $\ell_2$ relative error (L2RE):
\begin{equation} \label{eqn:interpolation_metrics}
    \text{L2RE}(f_\theta, f) = \frac{\left\| f_\theta - f \right\|_2}{\left\| f \right\|_2}
= \sqrt{ \frac{\sum_{i=1}^{N_{\text{test}}} \left( f_\theta(x_i) - f(x_i) \right)^2 }{ \sum_{i=1}^{N_{\text{test}}} f(x_i)^2 } }.
\end{equation}

%% file: sections/appendix/experiments.tex
\section{Additional experimental details}

\subsection{1-D Interpolation} \label{subapp:expt_interpolation}
Here, we describe the experimental setup for our 1-D interpolation experiments (\Cref{sec:interpolation}).

\subsubsection{Task description}
To study interpolation across functions of varying smoothness, we consider sinusoids $f(x) = \sin(kx)$, $x \in [-1, 1]$, with varying frequency $k$. These serve as a controlled test case for examining how model precision scales with oscillatory complexity. We vary $k \in \{1, 2, 4, 8, 16, 32\}$. For each target function, we generate a training set of $N_{train}=100$ points sampled uniformly at random from the domain, and evaluate on a dense equispaced test grid of $N_{test} = 1000$ points. We run our interpolation experiments over five seeds, and report the median three results.

\subsubsection{Model architecture and optimizer details}
\label{subsubapp:expt_interpolation_mlp}
We compare standard MLPs, \methodname-hatted MLPs, and explicit \methodname s on the 1-D interpolation task. We train all models to minimize MSE using the Adam optimizer and use a cosine decay learning rate scheduler with a minimum learning rate of $10^{-6}$.

\paragraph{Standard MLPs}
We use fully-connected MLPs with $\texttt{tanh}$ activations. We sweep network widths within $\{2^4, \dots, 2^8\}$ and depths from $2-8$ layers. We choose our initial learning rate by sweeping LR for the smallest MLP, and adjust the LR for larger MLPs by decreasing the initial learning rate by $\sqrt{ab}$ whenever we scale up the width by a factor of $a \times$ and the depth by a factor of $b \times$. Our base LR for the smallest MLP is $0.05$.

\paragraph{\methodname-hatted MLPs}
We apply our \methodname-hats atop standard fully-connected MLPs with $\texttt{tanh}$ activations, with width 256 and 3 hidden layers. We evaluate how precision scales with $N$, the number of nodes in the \methodname-hat, as we vary $N \in \{2^0, \dots, 2^6\}$. We use an initial LR of $0.05$ for our \methodname-hatted MLPs.

\paragraph{Explicit \methodname s}
As with \methodname-hatted MLPs, we sweep $N \in \{2^0, \dots, 2^6\}$ and evaluate the precision scaling. We use an initial LR of $0.01$ for all our explicit \methodname s.

\subsubsection{Chebyshev least squares} \label{subsubapp:expt_interpolation_chebls}

As a classical baseline for function interpolation, we fit Chebyshev polynomials via least squares regression. Given a target function \( f \), we construct a design matrix \( A \in \mathbb{R}^{N_{\text{train}} \times (d+1)} \), where each row contains the values of the first \( d+1 \) Chebyshev polynomials \( T_0(x), \ldots, T_d(x) \) evaluated at a training point \( x_i \). We then solve the linear system \( A c \approx f \) in the least-squares sense, where \( c \in \mathbb{R}^{d+1} \) are the polynomial coefficients. Note that this baseline performs polynomial interpolation in \emph{coefficient space}, whereas explicit \methodname~performs polynomial interpolation in \emph{value space}~\citep{trefethen2013approximation}.

\begin{figure}[h!]
    \centering
    \includegraphics[width=0.99\linewidth]{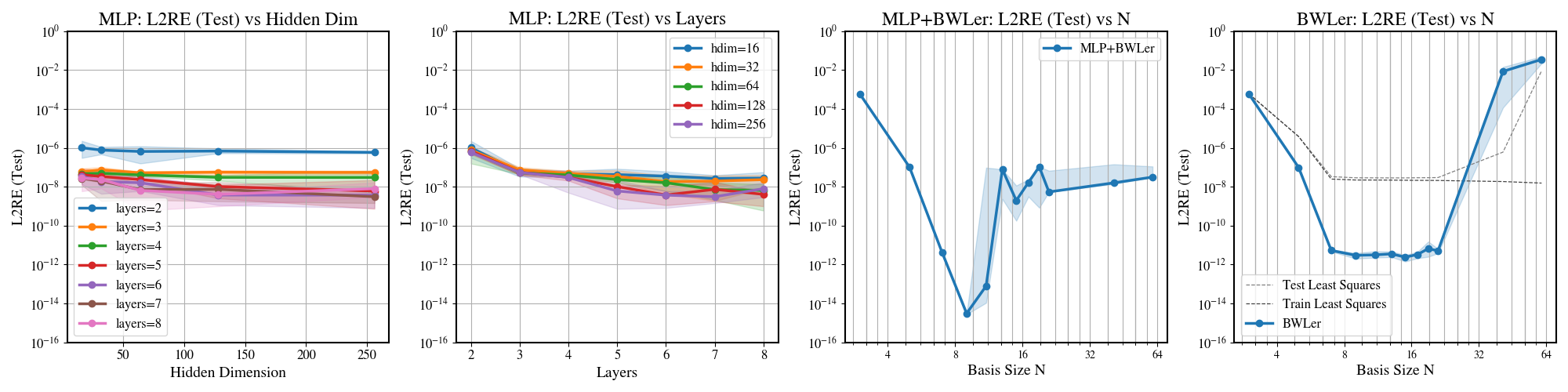}
    \includegraphics[width=0.99\linewidth]{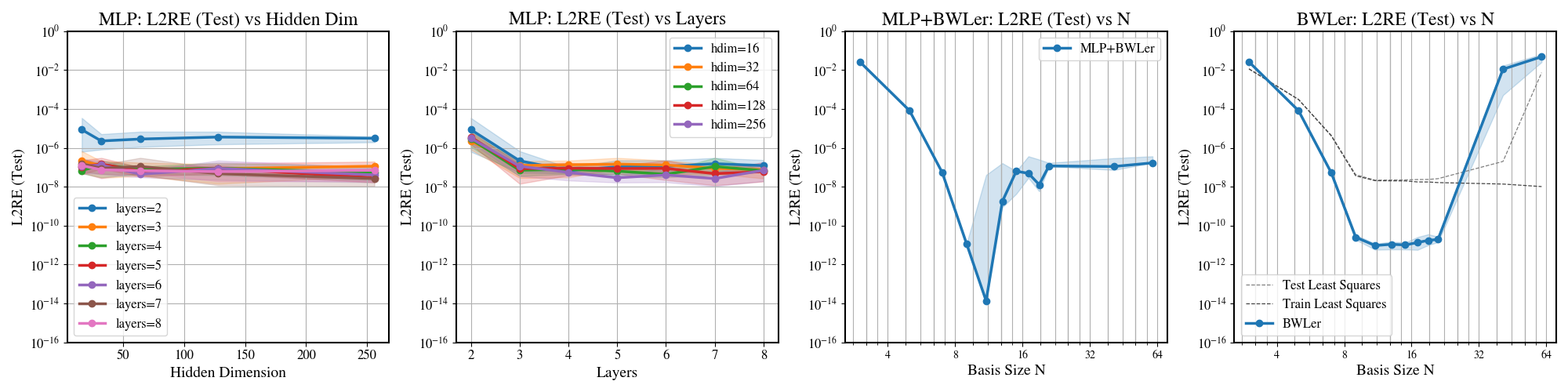}
    \includegraphics[width=0.99\linewidth]{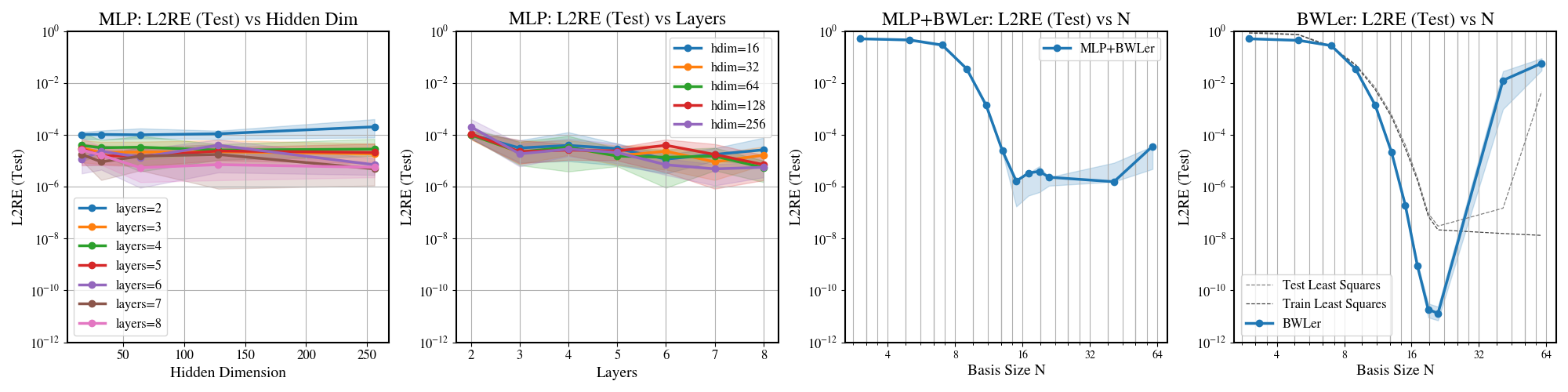}
    \includegraphics[width=0.99\linewidth]{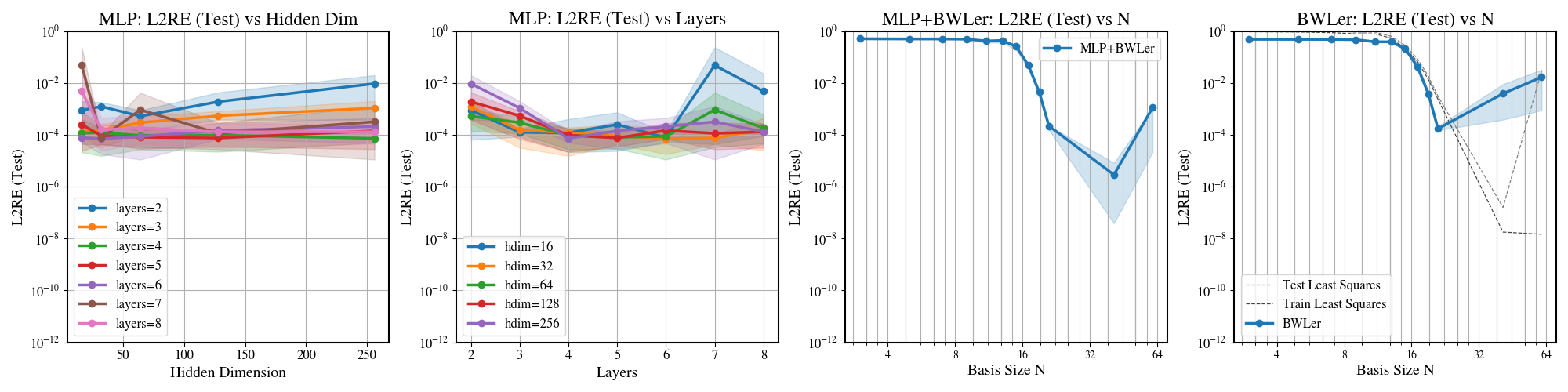}
    \includegraphics[width=0.99\linewidth]{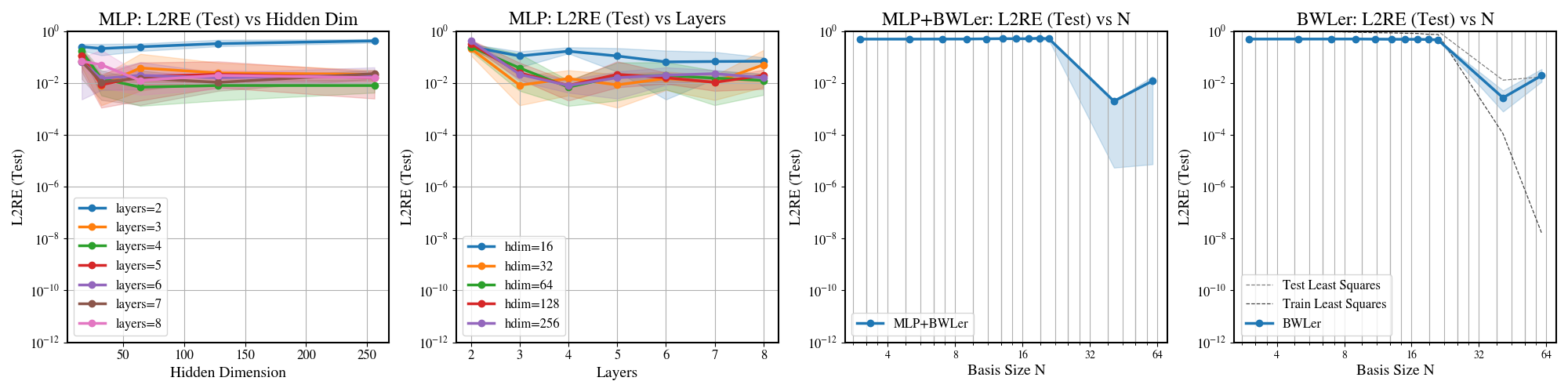}
    \caption{Comparison of standard MLPs, \methodname-hatted MLPs, and explicit \methodname s on 1-D interpolation with the target functions $f(x) = \sin(kx)$. From top to bottom: $k = 1,2,4,16,32$. Chebyshev least squares baseline plotted in dotted line on rightmost plots.}
    \label{fig:interpolation_appendix}
\end{figure}

We implement this using NumPy’s \texttt{numpy.polynomial.chebyshev.chebfit} function to fit the coefficients on the training data, and \texttt{chebval} for evaluation on the test grid. This provides an efficient and numerically stable method for approximating smooth functions, and serves as a reference for assessing model convergence in Section~\ref{sec:interpolation}. Interestingly, we find that as the least squares problem becomes more ill-conditioned (i.e. as the degree of the polynomial $N$ approaches the dataset size $M$), our explicit \methodname~sometimes outperforms the least squares baseline on the test data (\Cref{fig:interpolation_appendix}). We attribute this to the early-stopping regularization effect of gradient descent on ill-conditioned least squares~\citep{boyd2004convex}.

\newpage
\subsection{PDEs}
\label{subapp:expt_pdes}

\subsubsection{Benchmark problems}
\label{subsubapp:expt_pdes_tasks}
We perform our experiments on five benchmark PDE problems from prior work:

\paragraph{Convection Equation.}
The one-dimensional convection equation is a first-order hyperbolic PDE commonly used to model phenomena in fluids, physics, and biology. We use the problem formulation from~\citet{rathore2024challengestrainingpinnsloss} and~\citet{wang2023expertsguidetrainingphysicsinformed}:
\begin{align*}
\frac{\partial u}{\partial t} + c \frac{\partial u}{\partial x} = 0, \quad &x \in (0,2\pi), \, t \in (0,1), \\
u(x,0) = \sin(x), \quad &x \in [0,2\pi], \\
u(0,t) = u(2\pi,t), \quad &t \in [0,1].
\end{align*}
The analytical solution is $u(x,t) = \sin(x - c t)$, where we set $c = 40, 80$ in our experiments.

\paragraph{Reaction Equation.}
The one-dimensional reaction equation is a non-linear ODE that models chemical reactions. We use the problem formulation from~\citet{rathore2024challengestrainingpinnsloss}:
\begin{align*}
\frac{\partial u}{\partial t} - \rho u(1 - u) = 0, \quad &x \in (0,2\pi), \, t \in (0,1), \\
u(x,0) = \exp\left(-\frac{(x-\pi)^2}{2(\pi/4)^2}\right), \quad &x \in [0,2\pi], \\
u(0,t) = u(2\pi,t), \quad &t \in [0,1].
\end{align*}
The analytical solution is $u(x,t) = \frac{h(x)e^{\rho t}}{h(x)e^{\rho t}+1-h(x)}$, where $h(x) = \exp\left(-\frac{(x-\pi)^2}{2(\pi/4)^2}\right)$ and $\rho = 5$ in our experiments.

\paragraph{Wave Equation.}
The one-dimensional wave equation is a second-order hyperbolic PDE that models wave propagation. We use the problem formulation from~\citet{rathore2024challengestrainingpinnsloss}:
\begin{align*}
\frac{\partial^2 u}{\partial t^2} - 4\frac{\partial^2 u}{\partial x^2} = 0, \quad &x \in (0,1), t \in (0,1), \\
u(x,0) = \sin(\pi x) + \frac{1}{2}\sin(\beta\pi x), \quad &x \in [0,1], \\
\frac{\partial u(x,0)}{\partial t} = 0, \quad &x \in [0,1], \\
u(0,t) = u(1,t) = 0, \quad &t \in [0,1].
\end{align*}
The analytical solution is $u(x,t) = \sin(\pi x)\cos(2\pi t) + \frac{1}{2}\sin(\beta\pi x)\cos(2\beta\pi t)$, where $\beta = 5$ in our experiments.

\paragraph{Burgers' Equation.}
The one-dimensional viscous Burgers' equation is a nonlinear PDE often used as a prototype for modeling shock waves. We follow the problem formulation from~\citet{hao2023pinnaclecomprehensivebenchmarkphysicsinformed}:
\begin{align*}
\frac{\partial u}{\partial t} + u \frac{\partial u}{\partial x} = \nu \frac{\partial^2 u}{\partial x^2}, \quad &x \in (-1, 1), \; t \in (0, 1), \\
u(x, 0) = -\sin(\pi x), \quad &x \in [-1, 1], \\
u(-1, t) = u(1, t) = 0, \quad &t \in [0, 1].
\end{align*}
We use $\nu = \frac{0.01}{\pi}$ in our experiments.

\paragraph{Poisson Equation.}
We consider the Poisson equation
\begin{align*}
-\Delta u &= 0,
\end{align*}
 on an irregular domain with four circular holes, following the setup in~\citet{hao2023pinnaclecomprehensivebenchmarkphysicsinformed}. The domain is defined as a square with four circular cutouts:
\[
\Omega = \Omega_{\text{rec}} \setminus \bigcup_i R_i, \quad \text{where } \Omega_{\text{rec}} = [-0.5, 0.5]^2,
\]
and the four circles are:
\begin{align*}
R_1 &= \{(x, y) : (x - 0.3)^2 + (y - 0.3)^2 \leq 0.1^2 \}, \\
R_2 &= \{(x, y) : (x + 0.3)^2 + (y - 0.3)^2 \leq 0.1^2 \}, \\
R_3 &= \{(x, y) : (x - 0.3)^2 + (y + 0.3)^2 \leq 0.1^2 \}, \\
R_4 &= \{(x, y) : (x + 0.3)^2 + (y + 0.3)^2 \leq 0.1^2 \}.
\end{align*}

The boundary conditions are:
\begin{align*}
u &= 0, \quad x \in \partial R_i, \\
u &= 1, \quad x \in \partial \Omega_{\text{rec}}.
\end{align*}

\subsubsection{Results with \methodname-hatted MLPs}
\label{subsubapp:expt_pdes_bwlerhatted}

\paragraph{Experiment setup.}
\begin{itemize}
    \item \textit{Benchmark PDE problems.}
    We compare standard MLPs vs. \methodname-hatted MLPs vs. explicit \methodname s on the convection, reaction, and wave equation benchmarks from~\citet{rathore2024challengestrainingpinnsloss}. Details are described in~\Cref{subsubapp:expt_pdes_tasks}.

    \item \textit{Model settings.}
    All the MLPs we use for the standard and \methodname-hatted MLP experiments use 3 layers and a hidden dimension of 256. The \methodname-hatted MLPs and explicit \methodname~models use the problem-specific \methodname~hyperparameters described in~\Cref{subsubapp:expt_pdes_bwler}.

    \item \textit{Optimization settings.}
     We train all models using Adam~\citep{kingma2014adam} for $10^6$ iterations. We use an initial learning rate of $10^{-3}$ and a cosine annealing learning rate schedule with a minimum learning rate of $10^{-6}$. We use the standard momentum hyperparameters $(\beta_1, \beta_2) = (0.9, 0.999)$.
\end{itemize}

\begin{figure}[h!]
    \centering
    \includegraphics[width=0.99\textwidth]{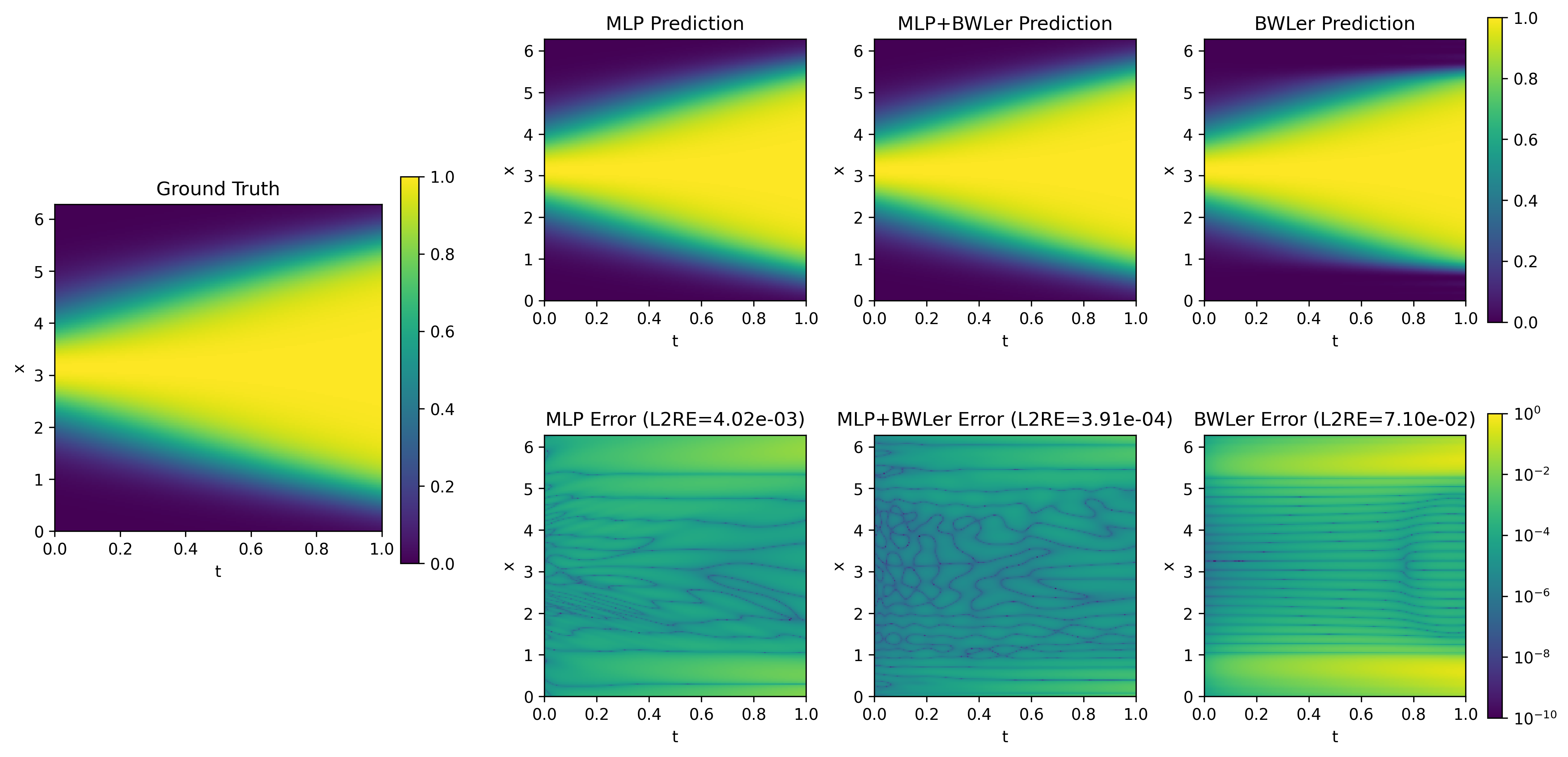}
    \caption{Standard MLP vs. \methodname-hatted MLP vs. explicit \methodname, evaluated on the reaction equation. For all models, we train for $10^6$ iterations with Adam.}
    \label{fig:pde_reaction_method_comparison}
\end{figure}

\begin{figure}[h!]
    \centering
    \includegraphics[width=0.99\textwidth]{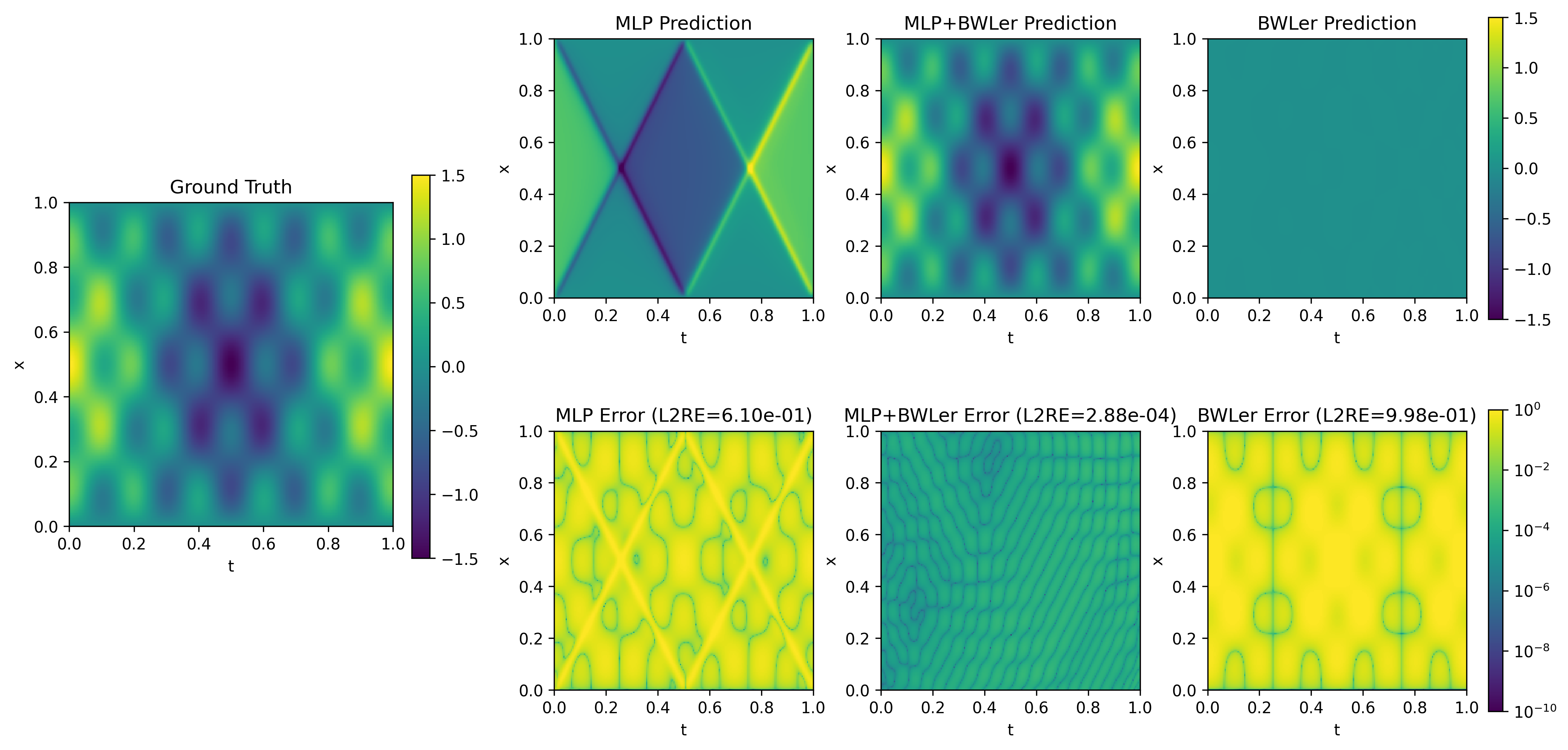}
    \caption{Standard MLP vs. \methodname-hatted MLP vs. explicit \methodname, evaluated on the wave equation. For all models, we train for $10^6$ iterations with Adam.}
    \label{fig:pde_wave_method_comparison}
\end{figure}

\newpage
\paragraph{\methodname~inherits the shortcomings of spectral methods.}
Although \methodname~can be flexibly applied to problems with complex boundary conditions and irregular domains, like standard PINNs, we do not expect \methodname-hatting to provide a consistent boost in performance across all PDE problems.
Since \methodname's performance guarantees depend on the smoothness of the target function, like standard polynomial approximation methods, it exhibits similar shortcomings to spectral solvers.

To highlight this, we compare the performance of a standard MLP vs. a \methodname-hatted MLP and explicit \methodname~on Burgers' equation, commonly used as a toy problem for shock capturing.
This is an adversarial test problem for spectral methods, as the solution is nearly discontinuous; standard results about polynomial approximation imply approximation error should converge as $O(1/N)$~\citep{trefethen2013approximation}, where $N$ is the number of nodes used in \methodname.
We note that our explicit \methodname~is equivalent to treating the 1+1D Burgers' equation spectrally in both space \emph{and time}. This is unorthodox and suboptimal; a more standard approach is treating space spectrally and performing time marching, e.g. via Exponential Time Differencing~\citep{cox2002exponential}.

We provide the results in~\Cref{tab:pdes_mlp_vs_mlpinterp_vs_interp_app}, alongside the results for the convection, reaction, and wave equations from~\Cref{tab:pdes_mlp_vs_mlpinterp_vs_interp} for comparison. We train with $10^6$ iterations of Adam for all methods.
We indeed find that \methodname's global treatment of the solution \emph{boosts} performance for smooth solutions, like the convection, reaction, and wave equations, but \emph{worsens} performance for the nearly-discontinuous solution of Burgers' equation.

\begin{table}[h]
\centering
\renewcommand{\arraystretch}{1.2}
\begin{tabular}{|c||c|c|c|}
    \hline
    \textbf{L2RE} ↓ & MLP & \methodname-hatted MLP & \methodname \\
    \hline
    \hline
    Convection & $1.14 \times 10^{0}$ 
               & $3.91 \times 10^{-2}$ {\scriptsize ($29.2\times$)} 
               & $\mathbf{4.07 \times 10^{-4}}$ {\scriptsize ($\mathbf{2800\times}$)} \\
    \hline
    Reaction   & $4.02 \times 10^{-3}$ 
               & $\mathbf{3.91 \times 10^{-4}}$ {\scriptsize ($\mathbf{10.3\times}$)} 
               & $7.10 \times 10^{-2}$ {\scriptsize ($0.057\times$)} \\
    \hline
    Wave       & $5.22 \times 10^{-1}$ 
               & $\mathbf{2.88 \times 10^{-4}}$ {\scriptsize ($\mathbf{1800\times}$)} 
               & $9.99 \times 10^{-1}$ {\scriptsize ($0.52\times$)} \\
    \hline
    \hline
    Burgers' & $\mathbf{4.99 \times 10^{-3}}$ & $2.43 \times 10^{-1}$ {\scriptsize ($0.021\times$)} & $9.49 \times 10^{-1}$ {\scriptsize ($0.005\times$)} \\
    \hline
    % Allen-Cahn & $4.95 \times 10^{-1}$ & $6.77 \times 10^{-2}$ & $7.98 \times 10^{-1}$ \\
    % \hline
\end{tabular}
\vspace{0.3cm}
\caption{L2 relative errors (L2RE) on benchmark PDEs: convection, reaction, and wave equations from~\citet{rathore2024challengestrainingpinnsloss}, and Burgers' equation from~\citet{hao2023pinnaclecomprehensivebenchmarkphysicsinformed}. Multiplicative improvements (in parentheses) are relative to the MLP baseline, where a factor less than 1 means a \emph{worse} performance than the standard MLP. All models are trained with Adam for $10^6$ iterations.}
\label{tab:pdes_mlp_vs_mlpinterp_vs_interp_app}
\end{table}

\newpage
\paragraph{Ablation: effect of \methodname-hatted MLP evaluation and differentiation.}
Note that when applying \methodname-hatting to an MLP, we can independently choose to use either \methodname's or the standard MLP's \emph{evaluate} and \emph{differentiate} operations.
By default, our \methodname-hatted MLPs use \methodname~for both evaluation and differentiation when solving PDEs. Here, we ablate the effect of two \methodname-hatted MLP variants:
\begin{itemize}
    \item \textit{\methodname-hatted MLP, forward only.} Uses \methodname's interpolation for evaluation but auto-differentiation of the MLP parameterization for the PDE derivatives.
    \item {\methodname-hatted MLP, derivative only.} Uses the standard MLP forward pass for evaluation but spectral derivatives from \methodname~for the PDE derivatives.
\end{itemize}
We compare to the standard \methodname-hatted MLP, which uses \methodname~for both evaluation and differentiation.

We evaluate on the convection equation from~\citet{rathore2024challengestrainingpinnsloss}, where the standard MLP only recovers a single oscillation of the true PDE solution, but the standard \methodname-hatted MLP recovers a qualitatively correct global solution (\Cref{tab:pdes_mlp_vs_mlpinterp_vs_interp}).
Interestingly, we find both variants of \methodname-hatting, forward only and derivative only, \emph{fail} to recover the global solution that the standard \methodname-hatted MLP does. This result supports our hypothesis that \methodname-hatting improves the ill-conditioning of the loss landscape by enforcing global consistency. The locality of the MLP, even when used only for evaluation alone or for differentiation alone, appears to disturb this effect, causing the ablation variants to converge to suboptimal local minima just like the standard MLP.

\begin{table}[h]
\centering
\renewcommand{\arraystretch}{1.2}
\begin{tabular}{|c||c|c|c|c|}
    \hline
    \textbf{L2RE} ↓ 
    & MLP 
    & \shortstack{ \methodname-hatted\\MLP (full) } 
    & \shortstack{ \methodname-hatted\\MLP (forward only) } 
    & \shortstack{ \methodname-hatted\\MLP (deriv only) } \\
    \hline
    \hline
    Convection & $1.14 \times 10^{0}$ 
               & $\mathbf{3.91 \times 10^{-2}}$ {\scriptsize ($\mathbf{29.2\times}$)} 
               & $9.59 \times 10^{-1}$ {\scriptsize ($1.19\times$)} 
               & $9.59 \times 10^{-1}$ {\scriptsize ($1.19\times$)} \\
    \hline
\end{tabular}
\vspace{0.3cm}
\caption{L2 relative errors (L2RE) of standard MLP and \methodname-hatted MLP variants on convection PDE from~\citet{rathore2024challengestrainingpinnsloss}. Multiplicative improvements (in parentheses) are relative to the MLP baseline. All models are trained with Adam for $10^6$ iterations.}
\label{tab:pdes_mlpinterp_ablation}
\end{table}

\newpage
\paragraph{Hessian spectral density plots.}
Here, we include plots of the Hessian spectral density as described in~\Cref{subsec:results_mlpinterp}. We compare standard MLPs, \methodname-hatted MLPs, and explicit \methodname s on the convection, reaction, and wave equations, trained for $10^6$ iterations with Adam (\Cref{tab:pdes_mlp_vs_mlpinterp_vs_interp}). After training, we take the final trained models and approximate the Hessian spectral density for each using PyHessian~\citep{yao2020pyhessian}; it implements the stochastic Lanczos algorithm and uses Hessian-vector products.

\begin{figure}[h!]
    \centering
    \includegraphics[width=0.9\textwidth]{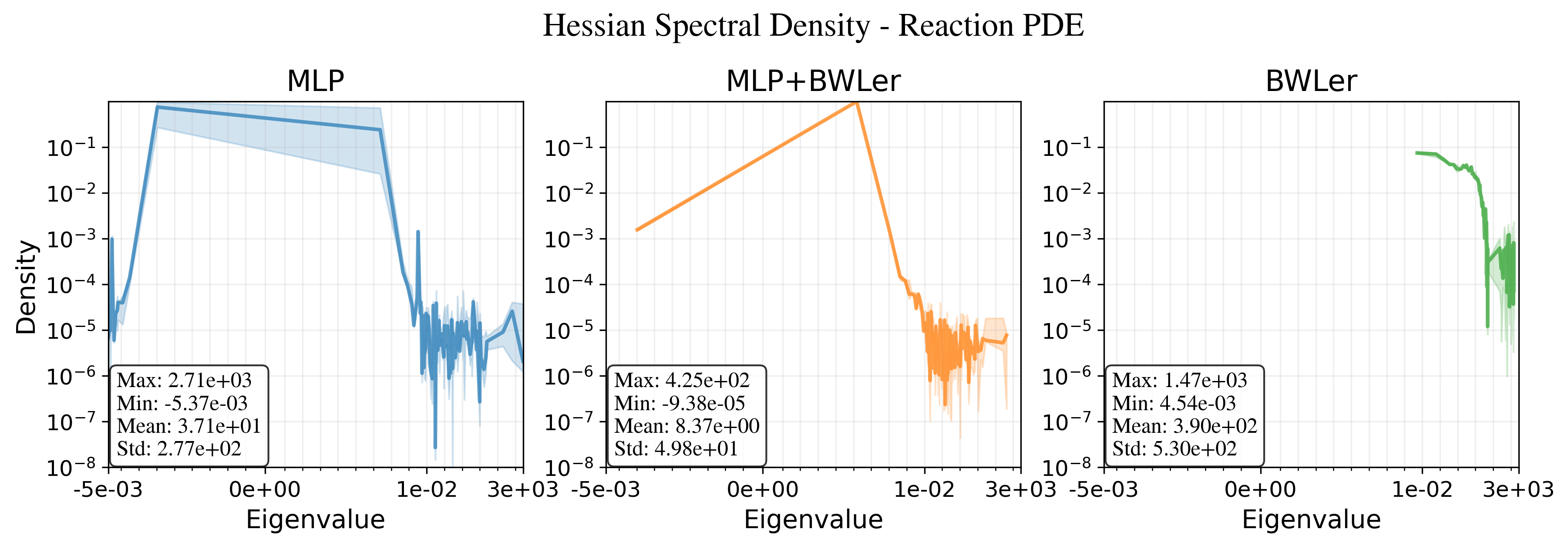}
    \caption{Hessian spectral density for the reaction equation.}
    \label{fig:pde_reaction_method_comparison_spectrum}
\end{figure}

\begin{figure}[h!]
    \centering
    \includegraphics[width=0.9\textwidth]{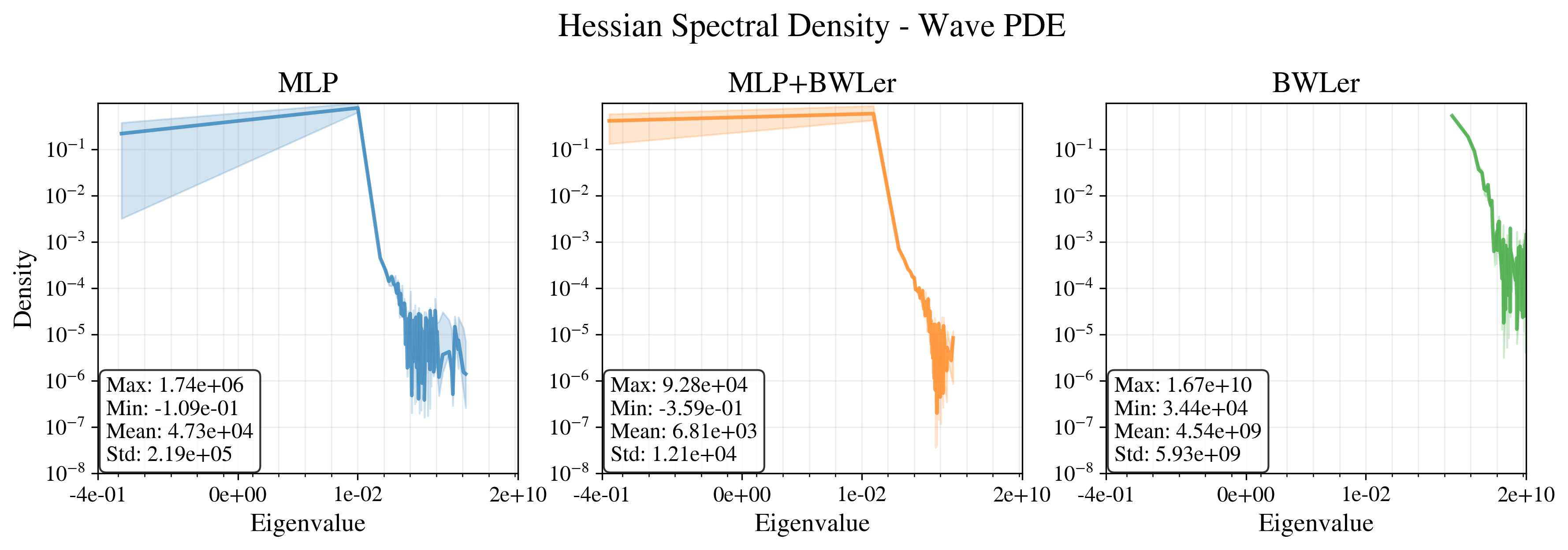}
    \caption{Hessian spectral density for the wave equation.}
    \label{fig:pde_wave_method_comparison_spectrum}
\end{figure}

We find that \methodname-hatting reduces the maximum eigenvalue by $10\times$ and the mean eigenvalue by $5$--$10\times$ on the reaction and wave equations (\Cref{fig:pde_reaction_method_comparison_spectrum}, \Cref{fig:pde_wave_method_comparison_spectrum}). This supports our hypothesis that \methodname's \emph{evaluate} and \emph{differentiate} operations, which depend globally on function values across the full domain, induce a less ill-conditioned loss landscape.

\begin{figure}[h!]
    \centering
    \includegraphics[width=0.9\textwidth]{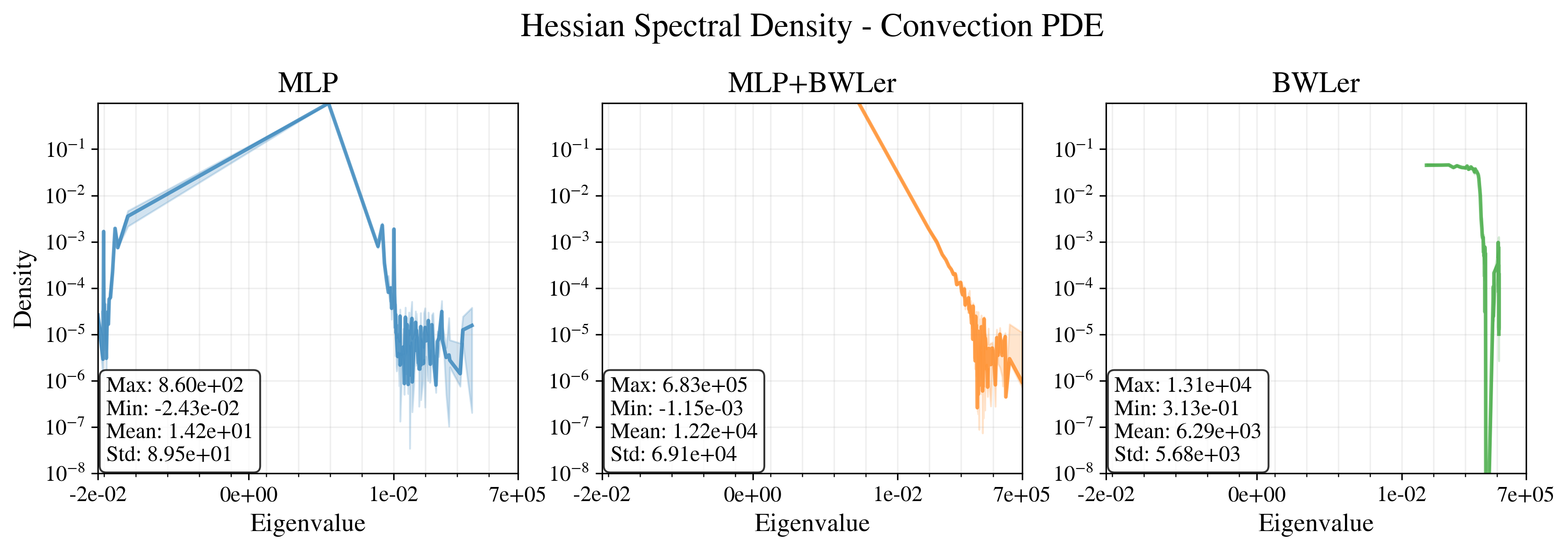}
    \caption{Hessian spectral density for the convection equation.}
    \label{fig:pde_convection_method_comparison_spectrum}
\end{figure}

\begin{figure}[h!]
    \centering
    \includegraphics[width=0.65\linewidth]{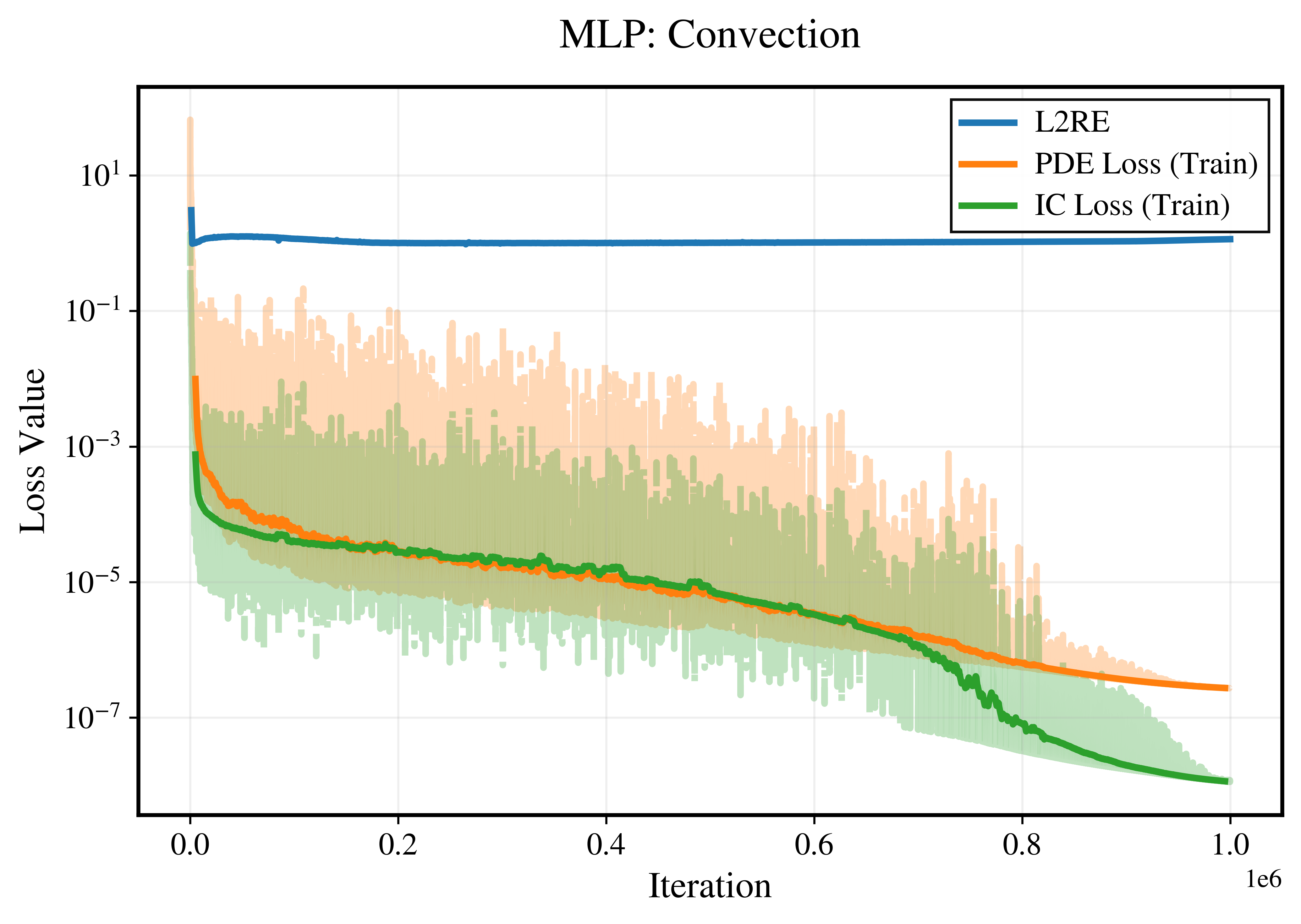}
    \includegraphics[width=0.65\linewidth]{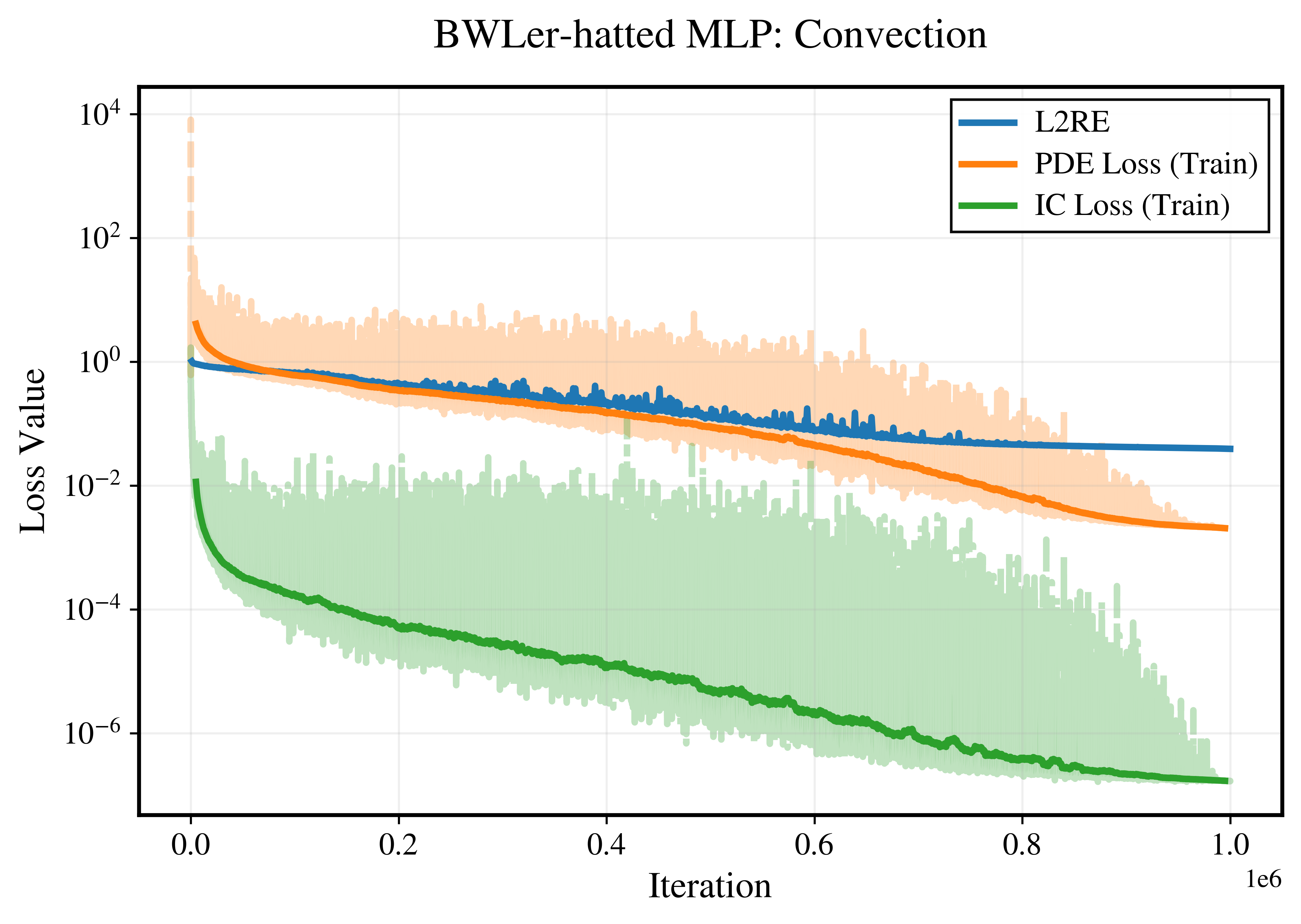}
    \includegraphics[width=0.65\linewidth]{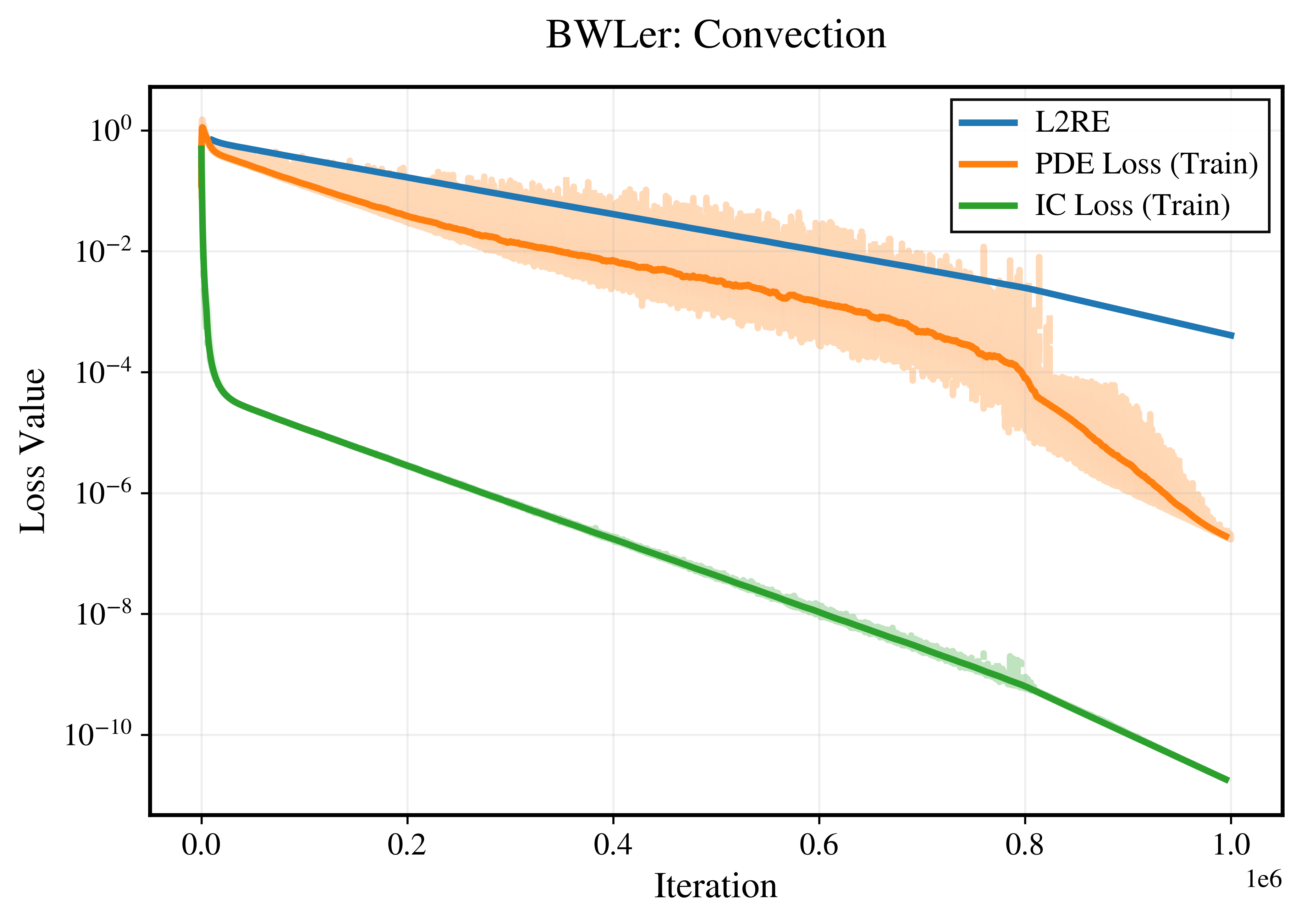}
    \caption{Loss curves for standard MLP, \methodname-hatted MLP, and explicit \methodname~trained with Adam on convection equation with $c=40$ (\Cref{tab:pdes_mlp_vs_mlpinterp_vs_interp}).}
    \label{fig:loss_adam_convection}
\end{figure}

Interestingly, we find that \methodname-hatting worsens the conditioning on the convection equation compared to the standard MLP (\Cref{fig:pde_convection_method_comparison_spectrum}) -- but this is because the standard MLP converges to a suboptimal local minima which is surprisingly effective at minimizing the PINN loss. See Figures~\ref{fig:loss_adam_convection} and~\ref{fig:main} for visualizations.

\newpage
\paragraph{Loss curves.}
We provide loss curves for the experiments comparing standard MLPs, \methodname-hatted MLPs, and explicit \methodname s trained with Adam (\Cref{tab:pdes_mlp_vs_mlpinterp_vs_interp}).

\textbf{Convection Equation.}

\newpage
\textbf{Reaction Equation.}
\begin{figure}[h!]
    \centering
    \includegraphics[width=0.65\linewidth]{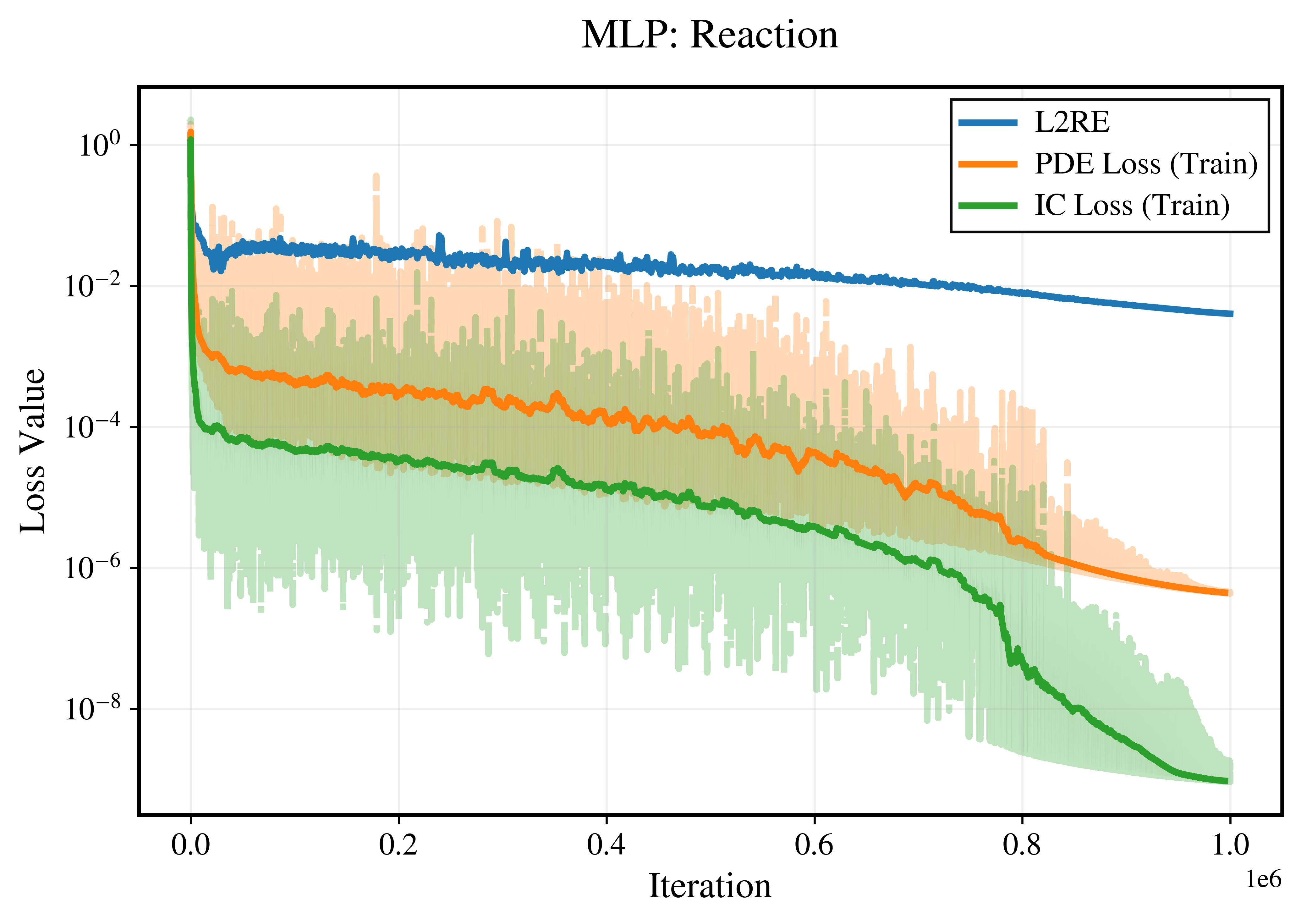}
    \includegraphics[width=0.65\linewidth]{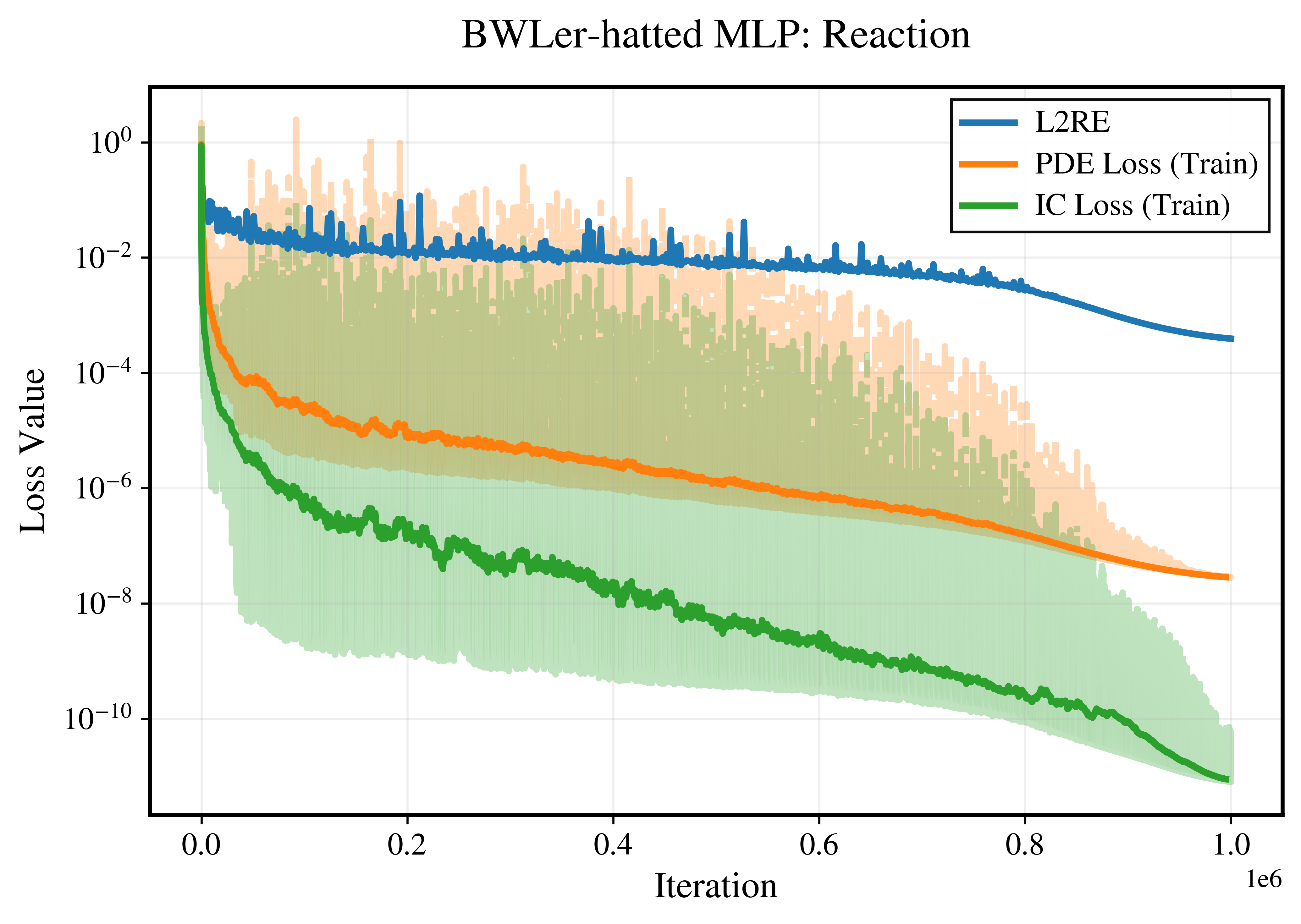}
    \includegraphics[width=0.65\linewidth]{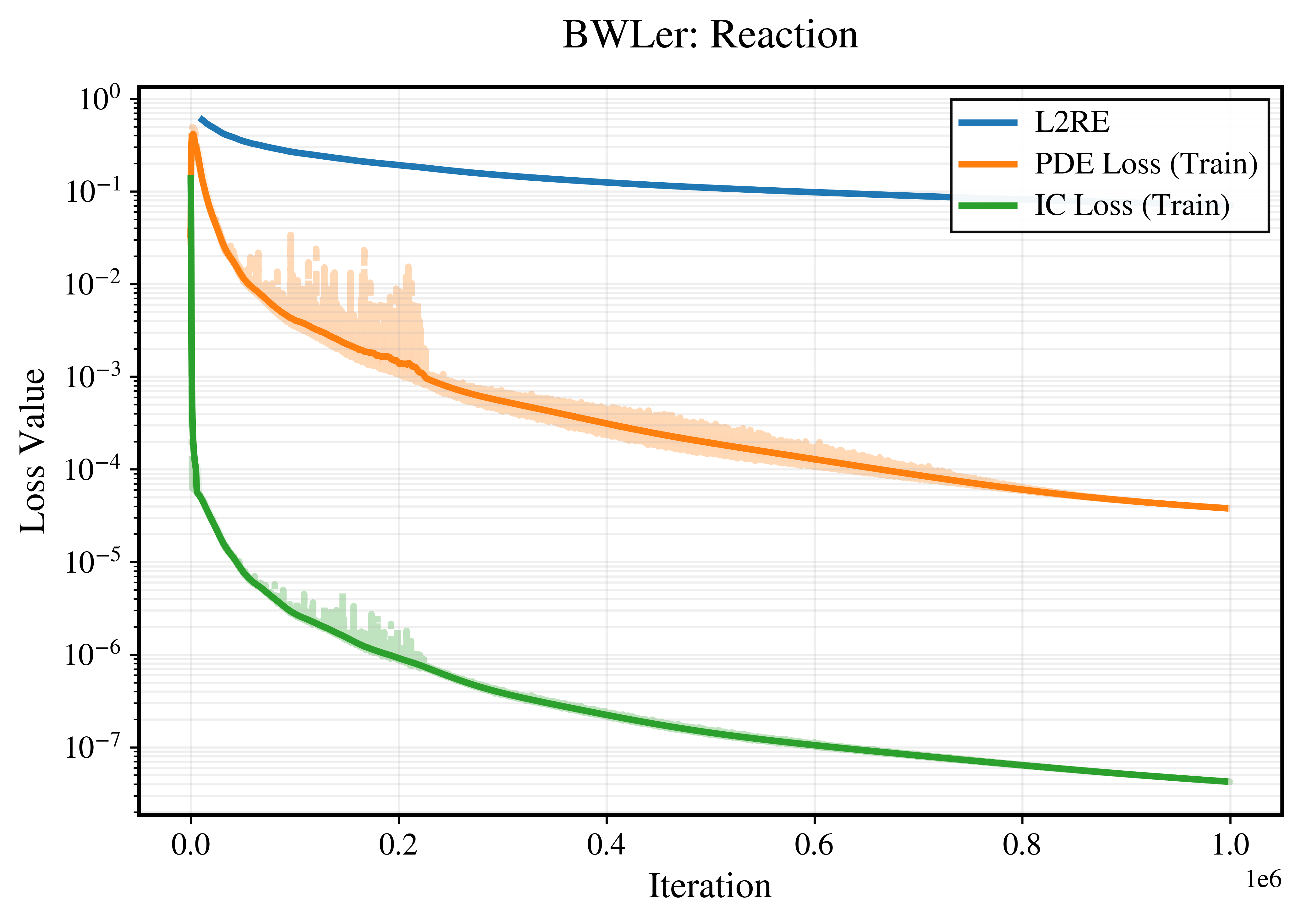}
    \caption{Loss curves for standard MLP, \methodname-hatted MLP, and explicit \methodname~trained with Adam on reaction equation (\Cref{tab:pdes_mlp_vs_mlpinterp_vs_interp}).}
    \label{fig:loss_adam_reaction}
\end{figure}

\newpage
\textbf{Wave Equation.}
\begin{figure}[h!]
    \centering
    \includegraphics[width=0.65\linewidth]{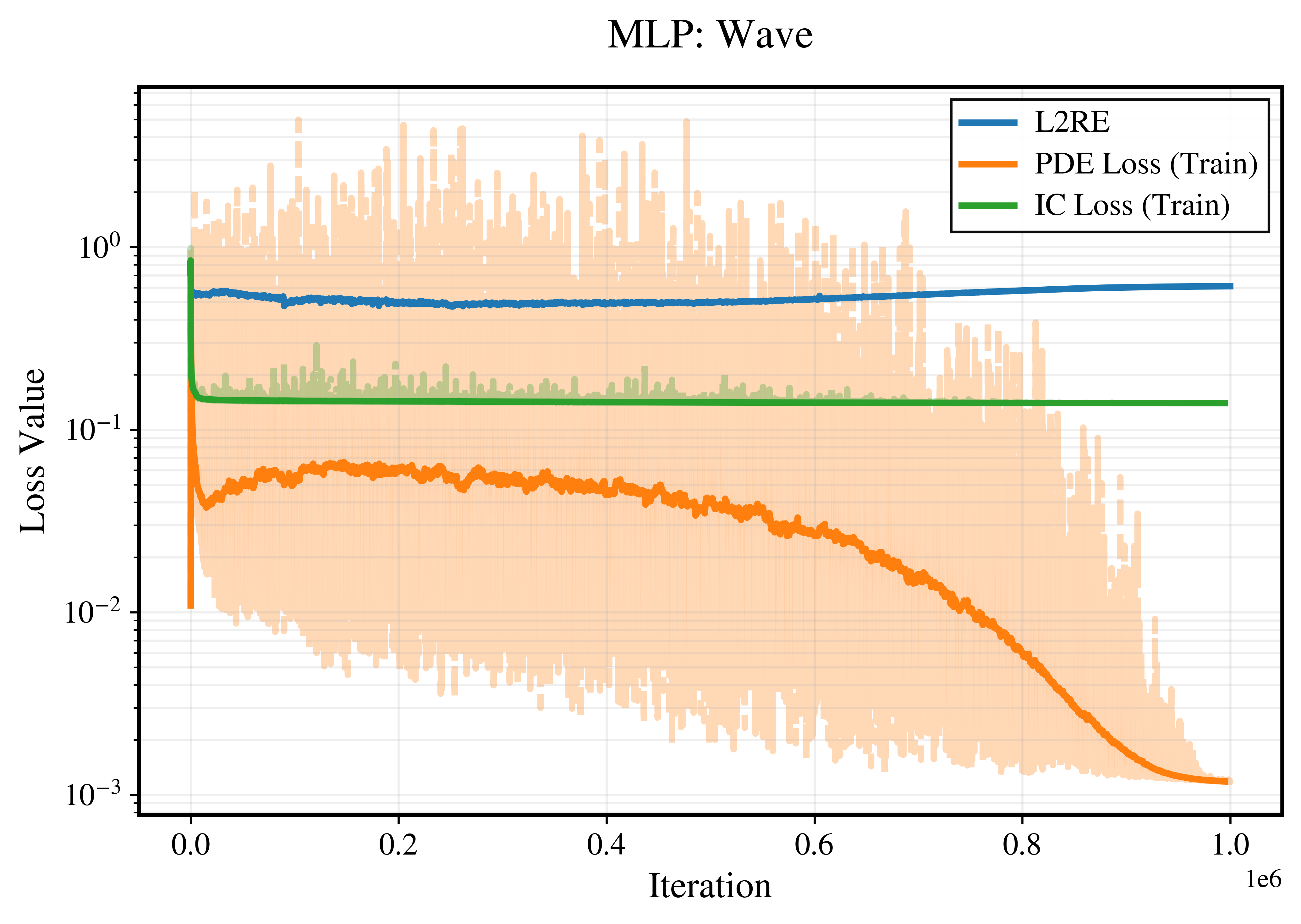}
    \includegraphics[width=0.65\linewidth]{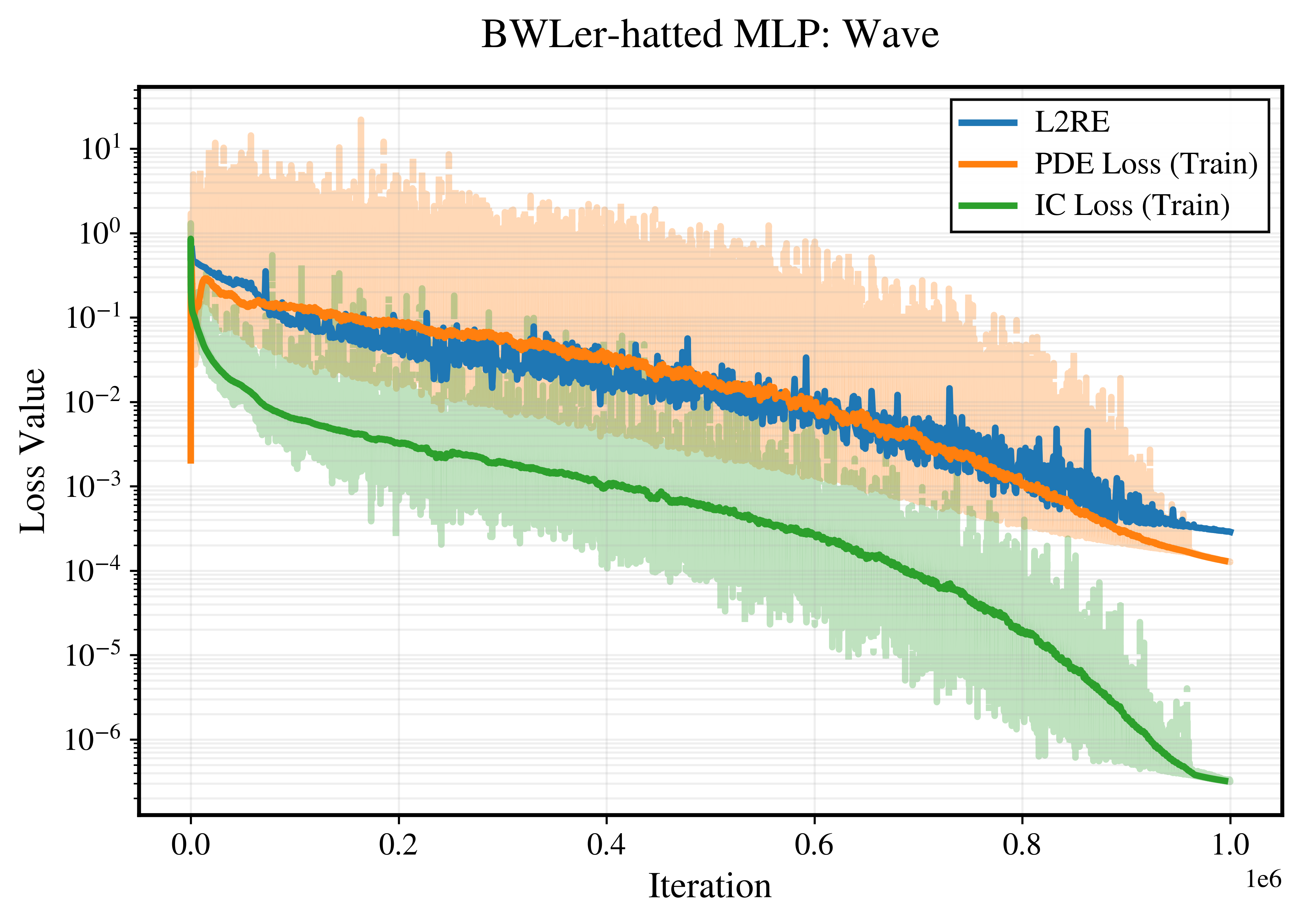}
    \includegraphics[width=0.65\linewidth]{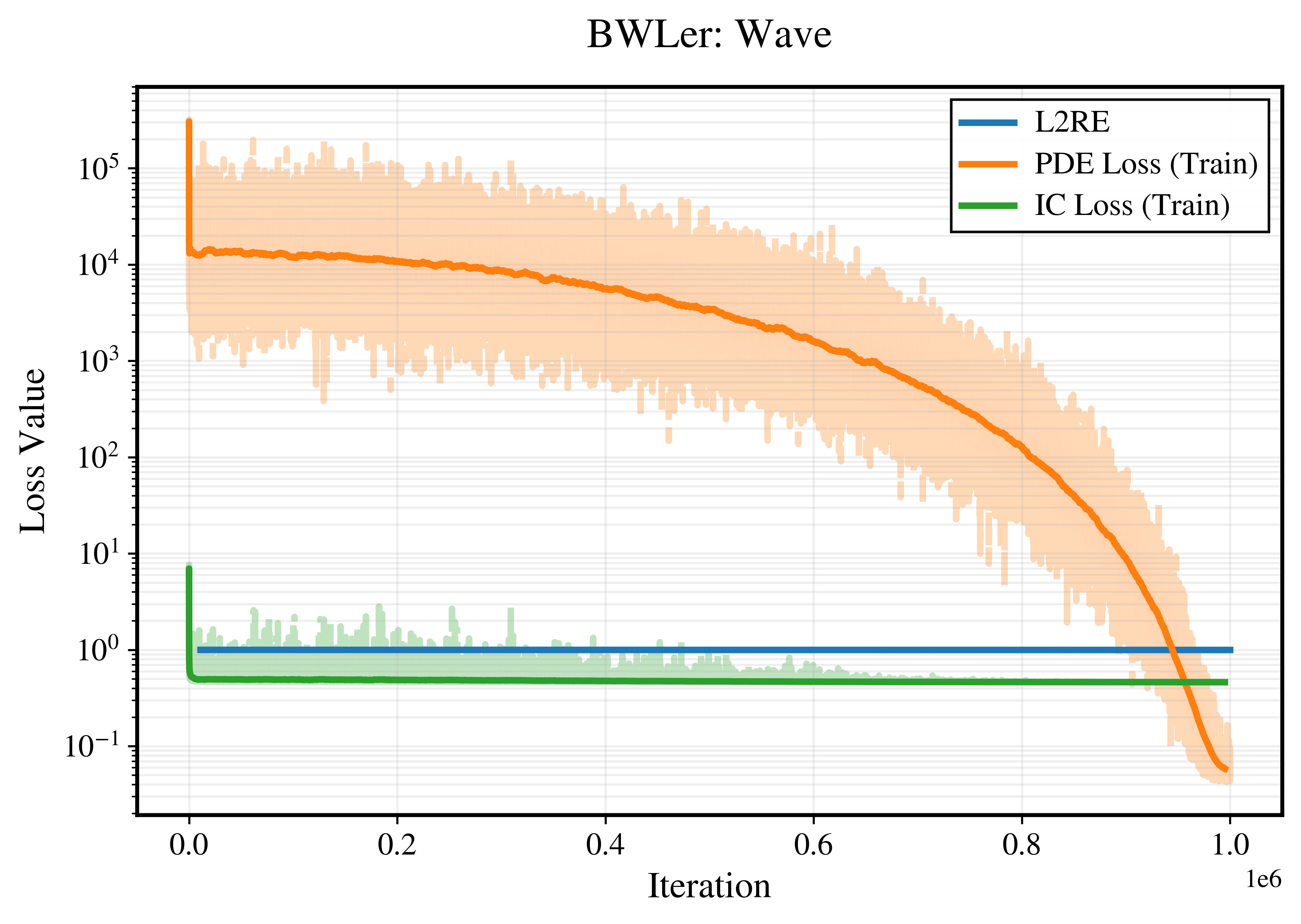}
    \caption{Loss curves for standard MLP, \methodname-hatted MLP, and explicit \methodname~trained with Adam on wave equation (\Cref{tab:pdes_mlp_vs_mlpinterp_vs_interp}).}
    \label{fig:loss_adam_wave}
\end{figure}

\newpage
\subsubsection{Results with explicit \methodname s}
\label{subsubapp:expt_pdes_bwler}

We describe the problem-specific \methodname~architecture hyperparameter settings used in Tables~\ref{tab:pdes_mlp_vs_mlpinterp_vs_interp},~\ref{tab:pdes_sota}, and the high-precision optimization settings for the five benchmark PDE problems in~\Cref{tab:pdes_sota}. For each problem, we provide the loss curves, final learned solutions, and error residuals of the explicit \methodname~experiments from~\Cref{tab:pdes_sota}.

\paragraph{Convection Equation, $c=40$.}

\begin{itemize}
    \item \textit{Architecture.}
    We use $N_t = 81$, $N_x = 80$, where we treat time with a Chebyshev basis and space with a Fourier basis.
    \item \textit{High-precision optimization.}
    We train with Nystr\"{o}m-Newton-CG for $350$ steps, with a preconditioner rank of $1000$ and $100$ CG iterations per step. On an A100, the total training takes about 5 minutes (about 1.2 iterations per second).
\end{itemize}

\begin{figure}[h!]
    \centering
    \includegraphics[width=0.7\linewidth]{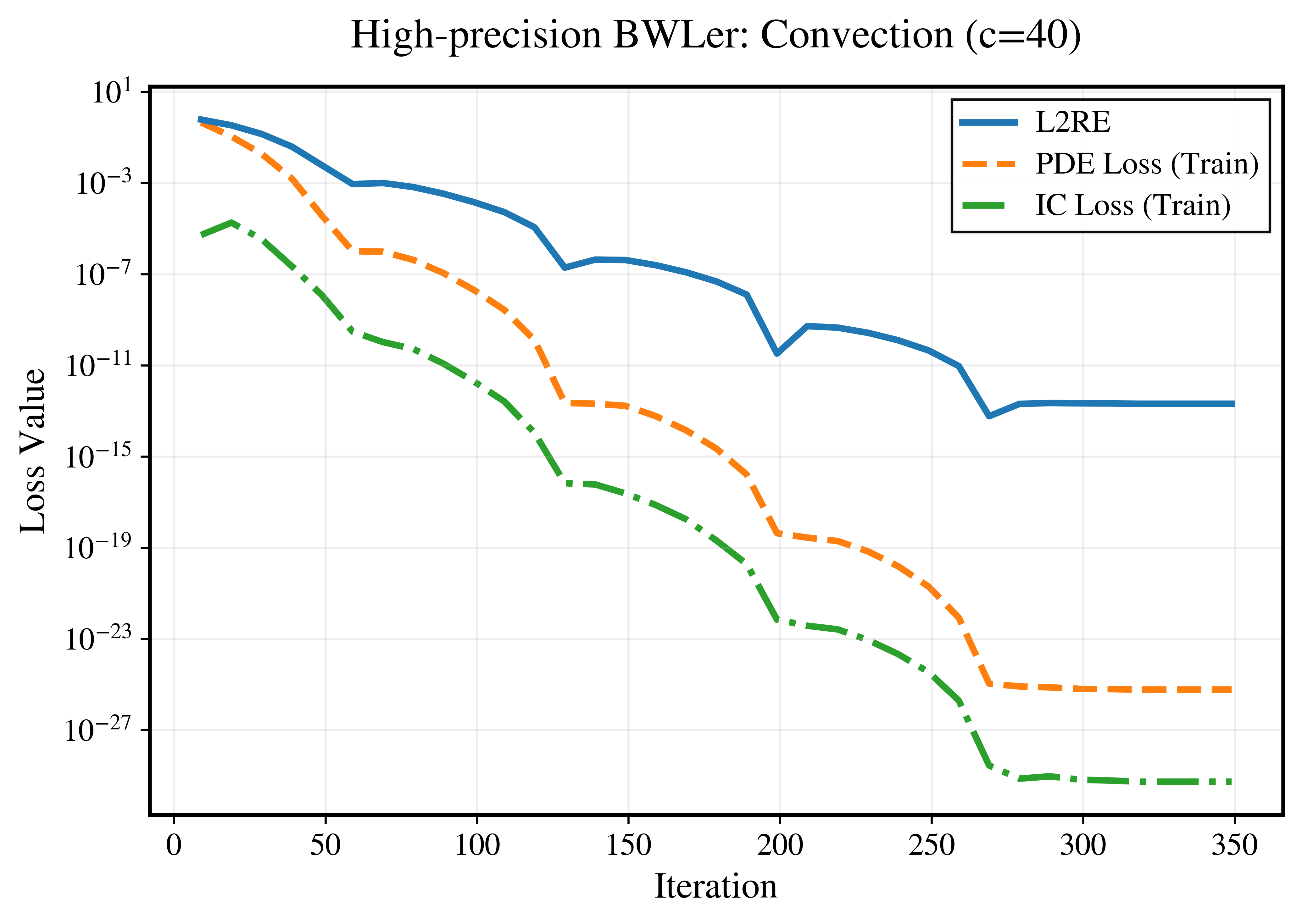}
    \caption{Loss curves for explicit \methodname~trained with NNCG on convection equation with $c=40$ (\Cref{tab:pdes_sota}).}
    \label{fig:loss_hp_convection_c=40}
\end{figure}
\begin{figure}[h!]
    \centering
    \includegraphics[width=\linewidth]{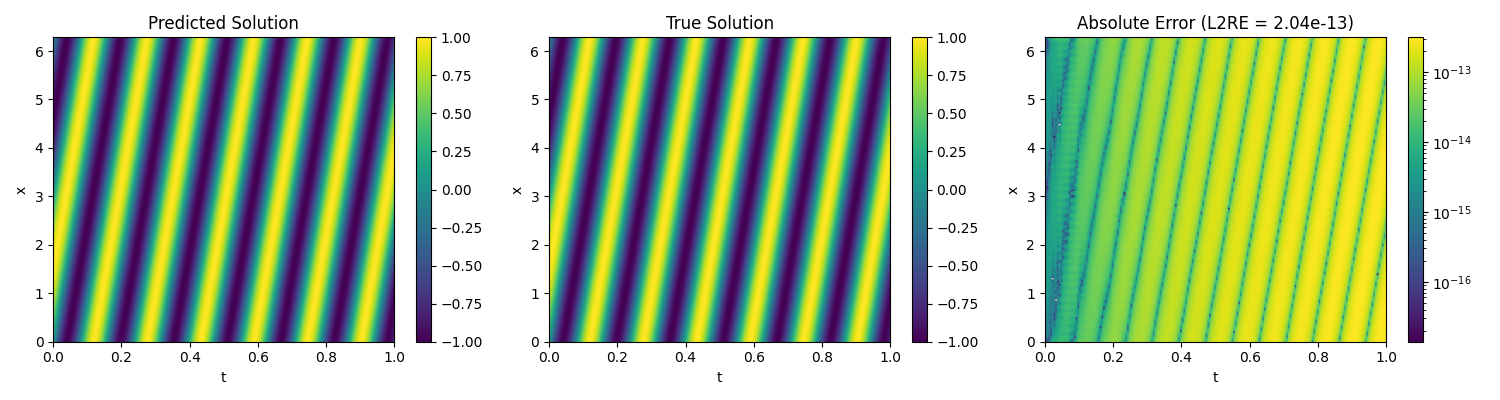}
    \caption{Explicit \methodname's learned solution and error residual on convection equation with $c=40$ (\Cref{tab:pdes_sota}).}
    \label{fig:residual_hp_convection_c=40}
\end{figure}

\newpage
\paragraph{Convection Equation, $c=80$.}

\begin{itemize}
    \item \textit{Architecture.}
    We use $N_t = 161$, $N_x = 160$, where we treat time with a Chebyshev basis and space with a Fourier basis.
    \item \textit{High-precision optimization.}
    We train with Nystr\"{o}m-Newton-CG for $2500$ steps, with a preconditioner rank of $1000$ and $100$ CG iterations per step. On an A100, the total training takes about 30 minutes (about 1.4 iterations per second).
\end{itemize}

\begin{figure}[h!]
    \centering
    \includegraphics[width=0.7\linewidth]{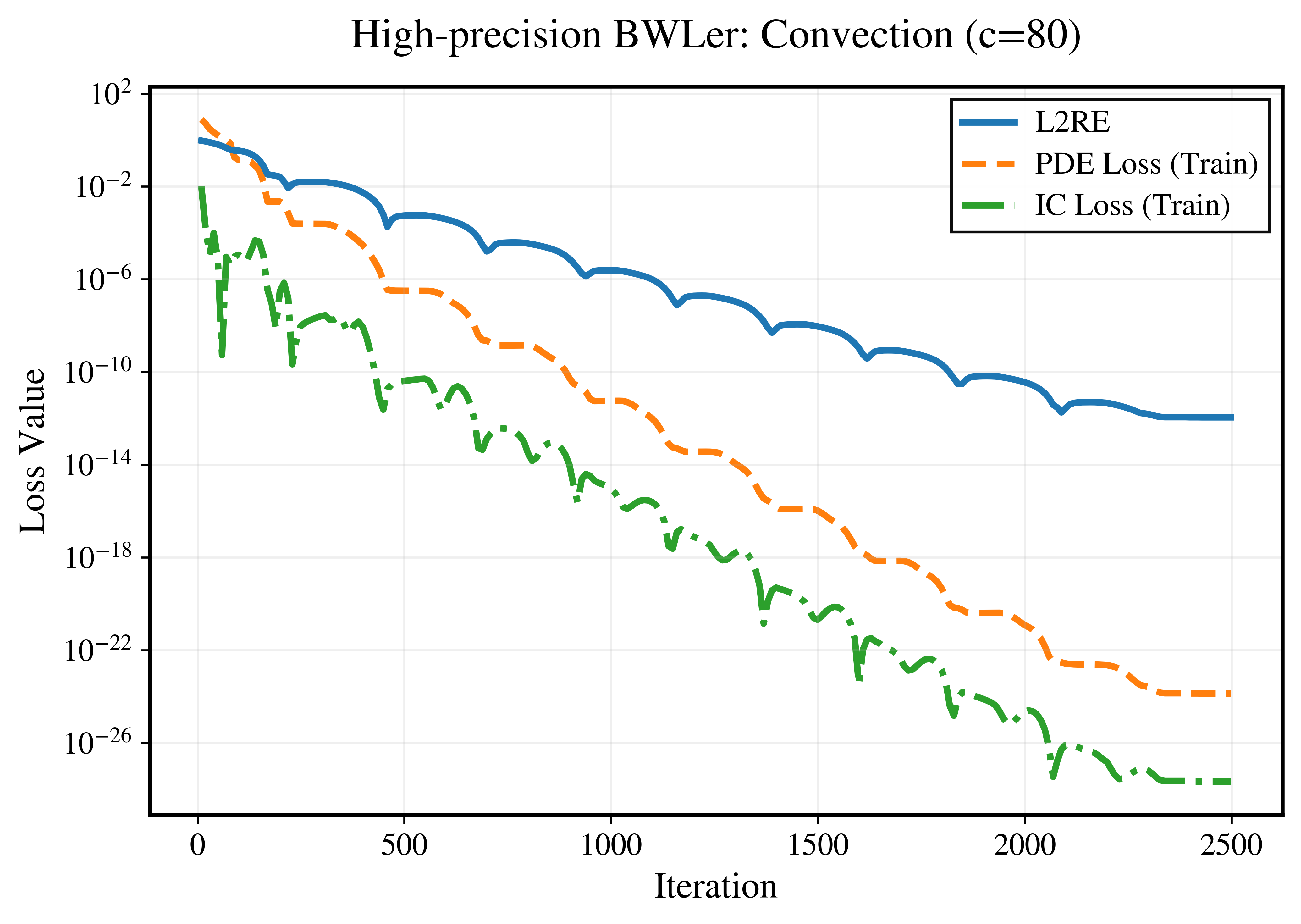}
    \caption{Loss curves for explicit \methodname~trained with NNCG on convection equation with $c=80$ (\Cref{tab:pdes_sota}).}
    \label{fig:loss_hp_convection_c=80}
\end{figure}
\begin{figure}[h!]
    \centering
    \includegraphics[width=\linewidth]{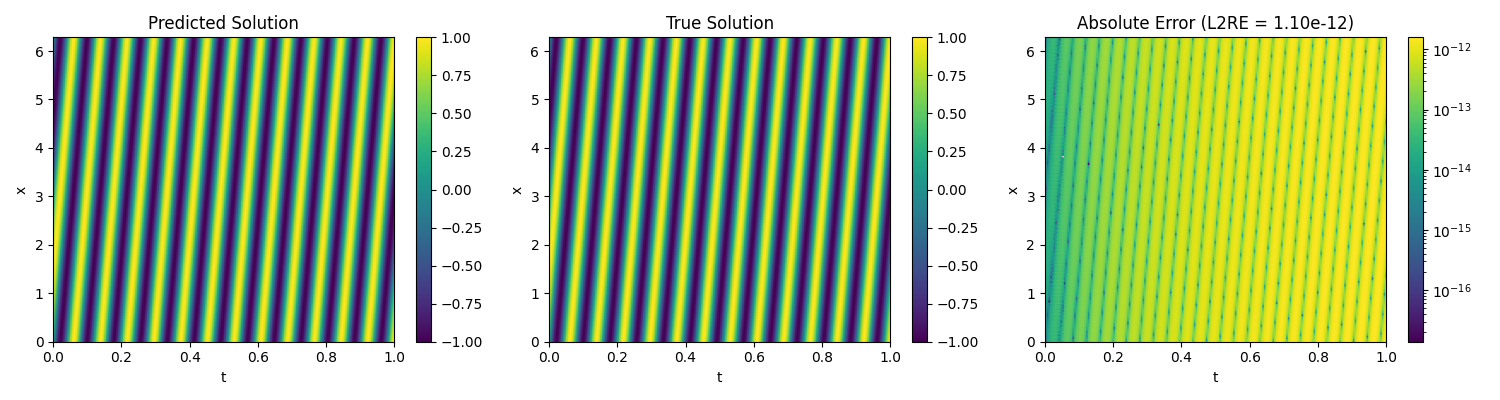}
    \caption{Explicit \methodname's learned solution and error residual on convection equation with $c=80$ (\Cref{tab:pdes_sota}).}
    \label{fig:residual_hp_convection_c=80}
\end{figure}

\newpage
\paragraph{Reaction Equation.}

\begin{itemize}
    \item \textit{Architecture.}
    We use $N_t = N_x = 81$, where we treat both time and space with a Chebyshev basis.
    \item \textit{High-precision optimization.}
    We train with Nystr\"{o}m-Newton-CG for $250,000$ steps, with a preconditioner rank of $16$ and $16$ CG iterations per step. On an A100, the total training takes about 8.5 hours (about 8.2 iterations per second).
\end{itemize}

\begin{figure}[h!]
    \centering
    \includegraphics[width=0.7\linewidth]{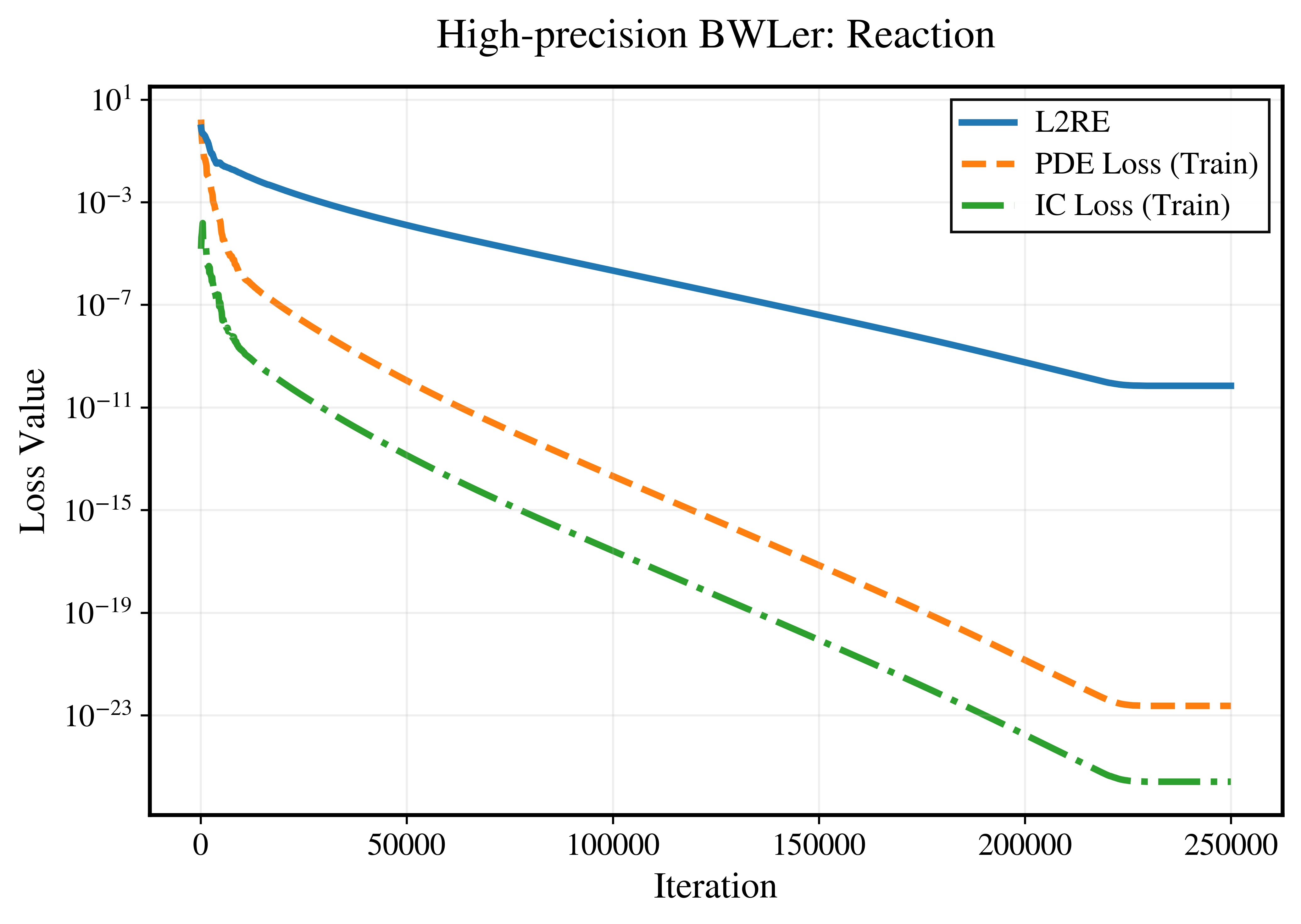}
    \caption{Loss curves for explicit \methodname~trained with NNCG on reaction equation (\Cref{tab:pdes_sota}).}
    \label{fig:loss_hp_reaction}
\end{figure}
\begin{figure}[h!]
    \centering
    \includegraphics[width=\linewidth]{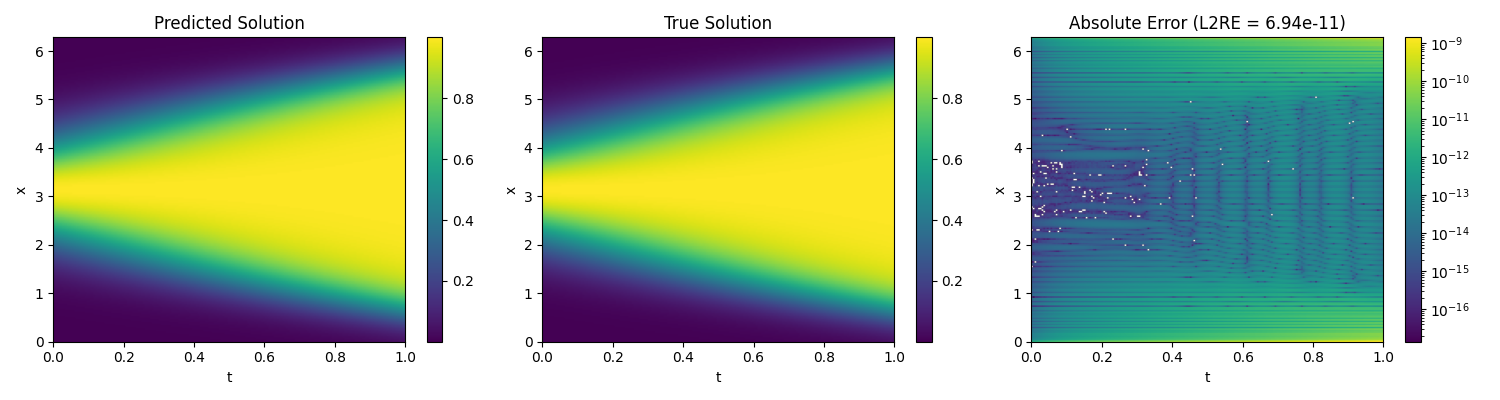}
    \caption{Explicit \methodname's learned solution and error residual on reaction equation (\Cref{tab:pdes_sota}).}
    \label{fig:residual_hp_reaction}
\end{figure}

\newpage\paragraph{Wave Equation.}

\begin{itemize}
    \item \textit{Architecture.}
    We use $N_t = N_x = 41$, where we treat both time and space with a Chebyshev basis.
    \item \textit{High-precision optimization.}
    We train with Nystr\"{o}m-Newton-CG for $200$ steps, with a preconditioner rank of $1000$ and $1000$ CG iterations per step. On an A100, the total training takes about 42 minutes (about 12.5 seconds per iteration).
\end{itemize}

\begin{figure}[h!]
    \centering
    \includegraphics[width=0.7\linewidth]{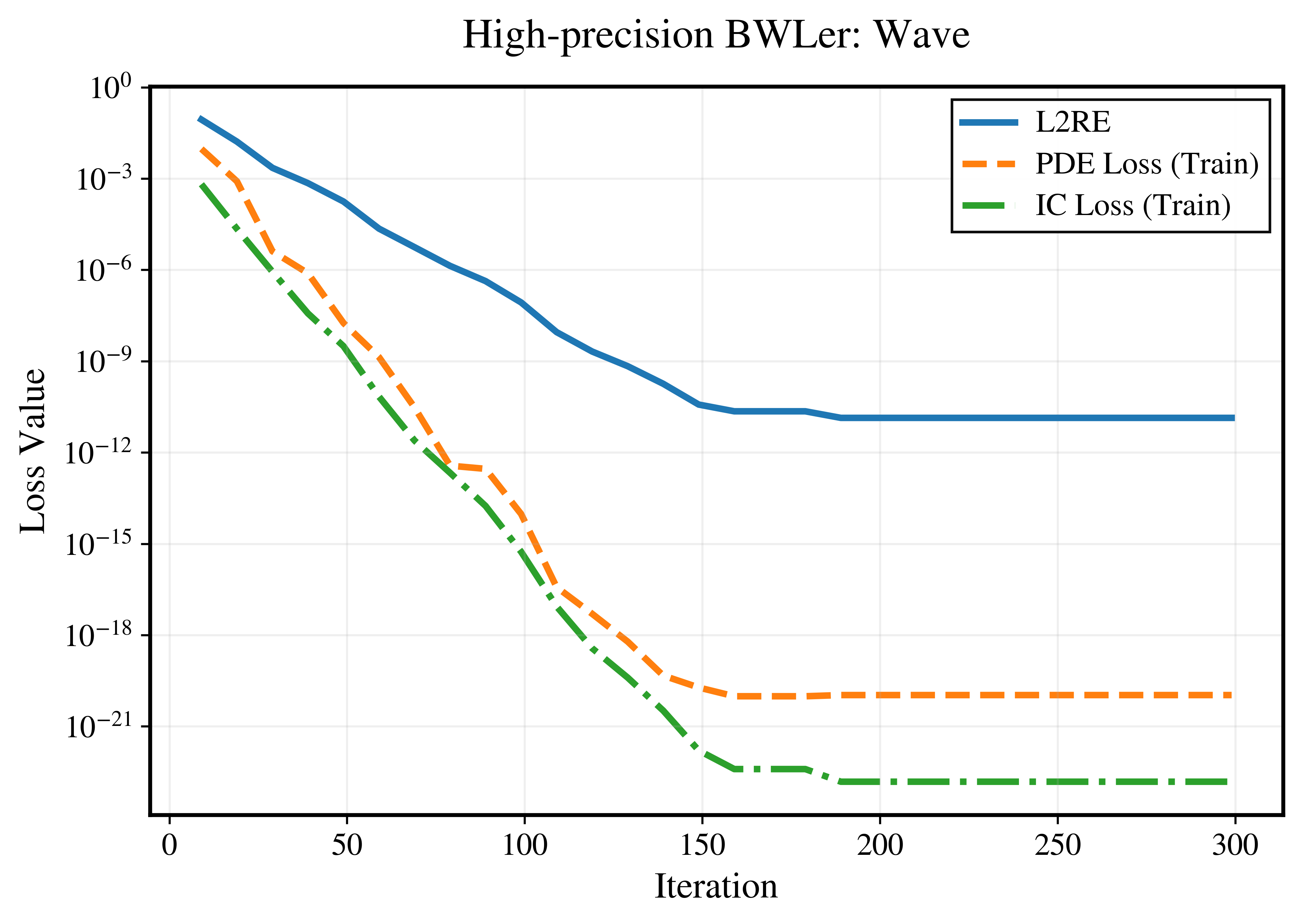}
    \caption{Loss curves for explicit \methodname~trained with NNCG on wave equation (\Cref{tab:pdes_sota}).}
    \label{fig:loss_hp_wave}
\end{figure}
\begin{figure}[h!]
    \centering
    \includegraphics[width=\linewidth]{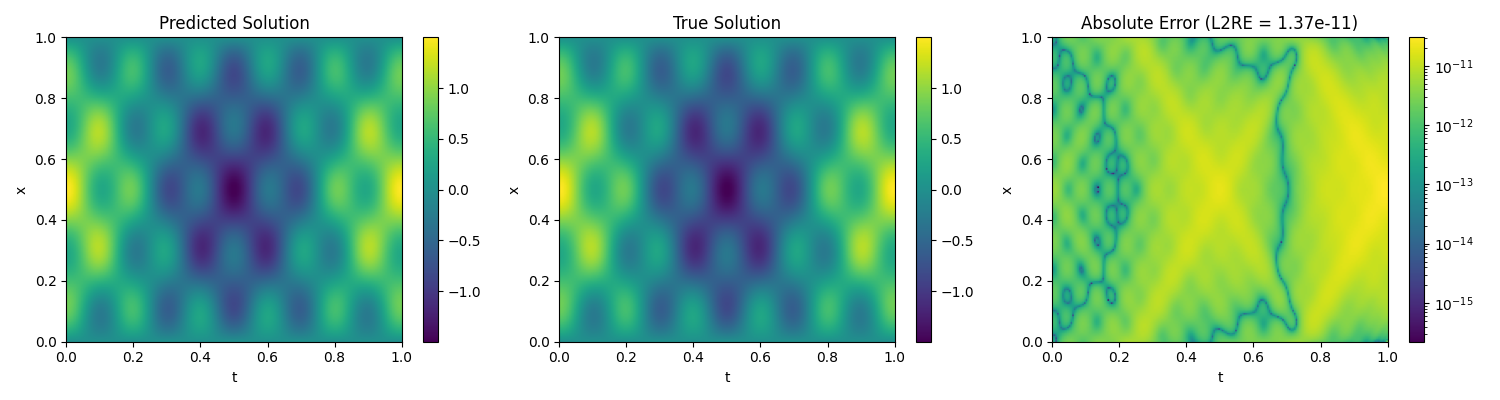}
    \caption{Explicit \methodname's learned solution and error residual on wave equation (\Cref{tab:pdes_sota}).}
    \label{fig:residual_hp_wave}
\end{figure}

\newpage
\paragraph{Burgers' Equation.}

\begin{itemize}
    \item \textit{Architecture.}
    We use $N_t = N_x = 321$, where we treat both time and space with a Chebyshev basis. For the experiments in~\Cref{tab:pdes_mlp_vs_mlpinterp_vs_interp}, we use spectral derivatives in both space and time. For the experiment in~\Cref{tab:pdes_sota}, we use spectral derivatives in space, and finite difference derivatives in time, using 1st-order, 3-point finite difference stencils.
    \item \textit{High-precision optimization.}
    We train with Nystr\"{o}m-Newton-CG for $850$ steps, with a preconditioner rank of $1000$ and $2000$ CG iterations per step. On an A100, the total training takes about 8.2 hours (about 35 seconds per iteration).
\end{itemize}

\begin{figure}[h!]
    \centering
    \includegraphics[width=0.7\linewidth]{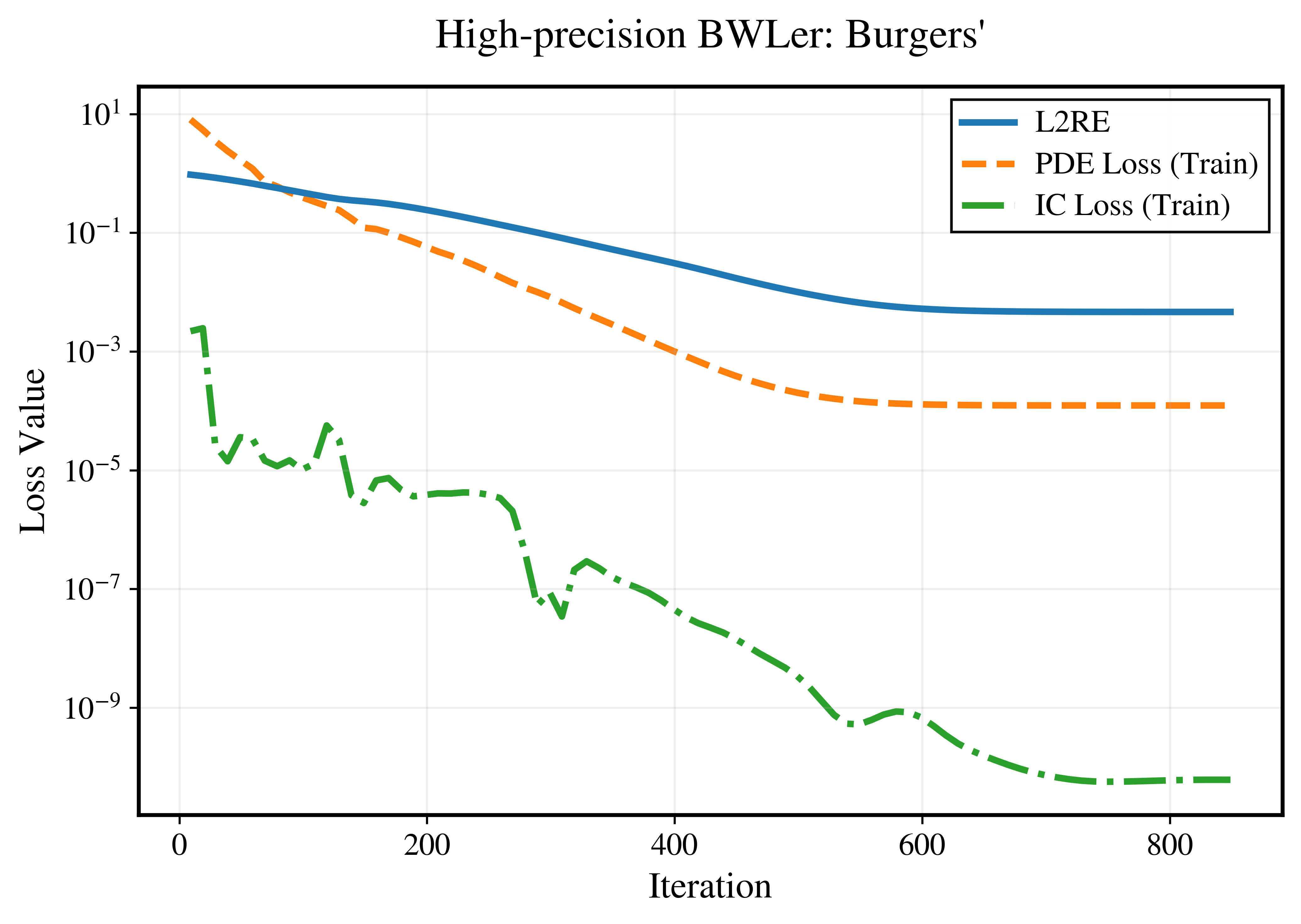}
    \caption{Loss curves for explicit \methodname~trained with NNCG on Burgers' equation (\Cref{tab:pdes_sota}).}
    \label{fig:loss_hp_burgers}
\end{figure}
\begin{figure}[h!]
    \centering
    \includegraphics[width=\linewidth]{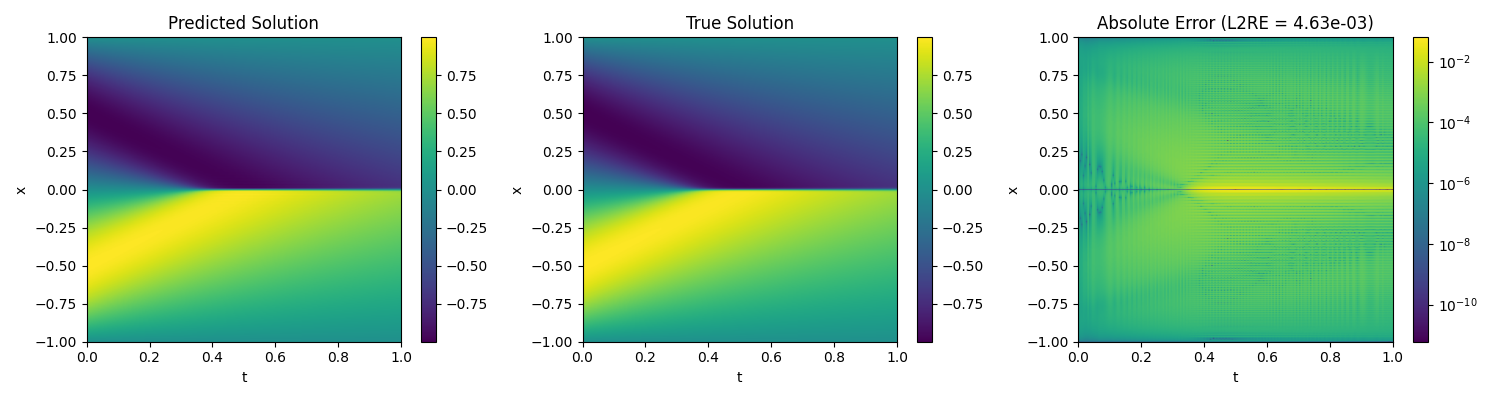}
    \caption{Explicit \methodname's learned solution and error residual on Burgers' equation (\Cref{tab:pdes_sota}).}
    \label{fig:residual_hp_burgers}
\end{figure}

\newpage
\paragraph{Poisson Equation.}

\begin{itemize}
    \item \textit{Architecture.}
    We use $N_x = N_y = 51$, where we treat both dimensions with a Chebyshev basis.
    \item \textit{High-precision optimization.}
    We train with Nystr\"{o}m-Newton-CG for 51,000 epochs with a preconditioner rank of 1000 and 64 CG iterations per step. On an A100, the total training time is about 8 hours (about 0.55 seconds per iteration).
\end{itemize}

\begin{figure}[h!]
    \centering
    \includegraphics[width=0.7\linewidth]{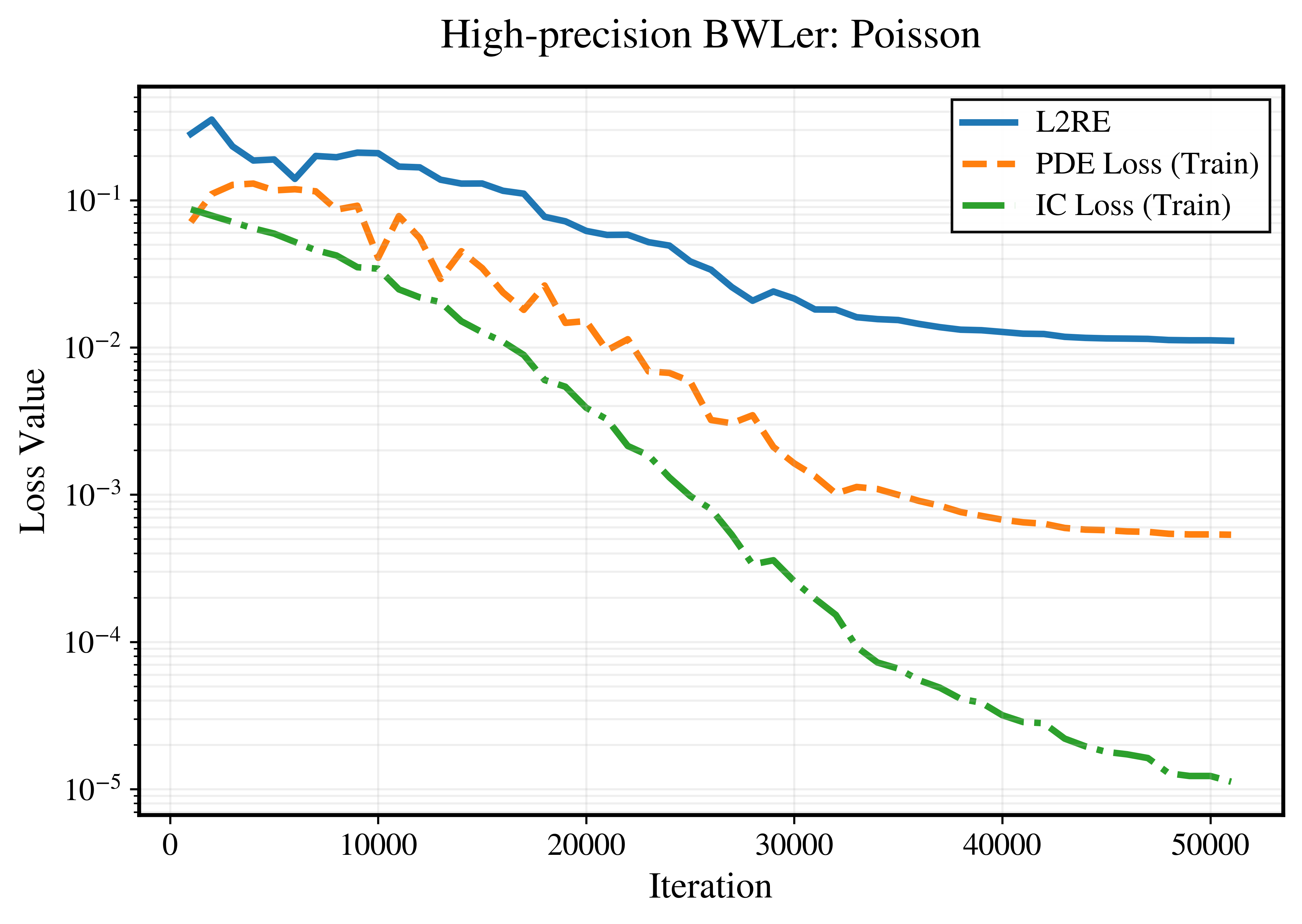}
    \caption{Loss curves for explicit \methodname~trained with NNCG on Poisson equation (\Cref{tab:pdes_sota}).}
    \label{fig:loss_hp_poisson}
\end{figure}
\begin{figure}[h!]
    \centering
    \includegraphics[width=\linewidth]{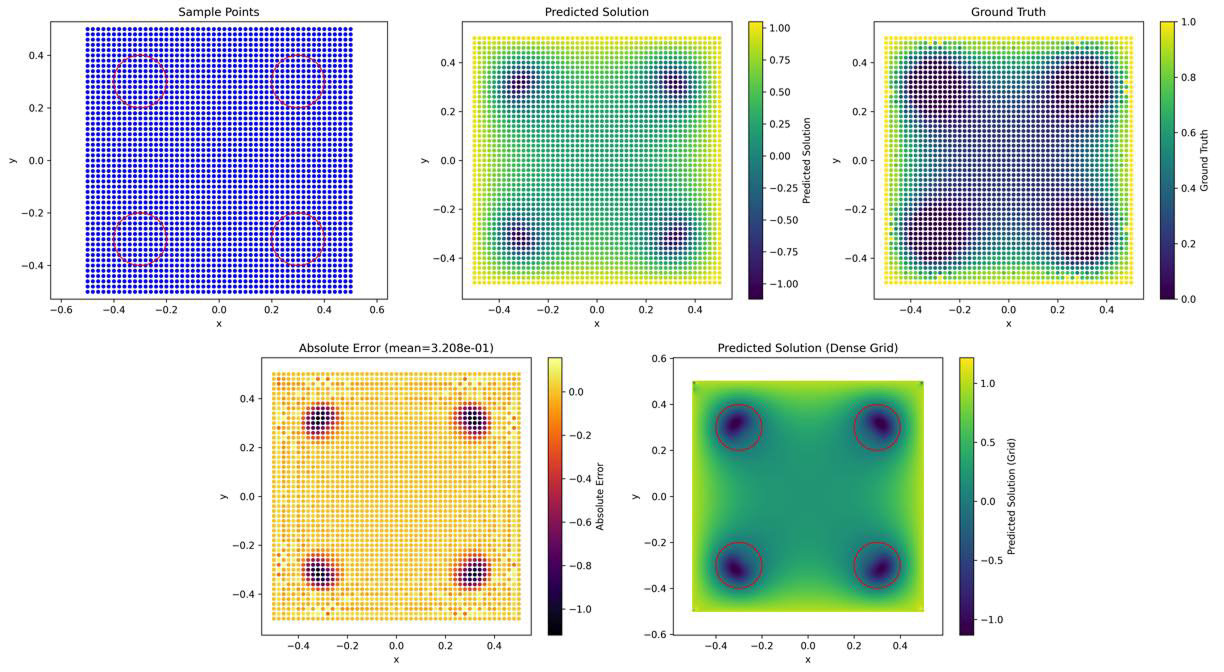}
    \caption{Explicit \methodname's learned solution and error residual on Poisson equation (\Cref{tab:pdes_sota}).}
    \label{fig:residual_hp_poisson}
\end{figure}

%% file: sections/appendix/theory.tex
\section{Theory}
\label{app:theory}

\input{sections/appendix/theory/interpolation}

\newpage
\input{sections/appendix/theory/pde}

%% file: sections/appendix/theory/interpolation.tex
\subsection{Formal statement and proof of Theorem~\ref{thm:bwler_interpolation_informal}}
\label{subapp:theory_interpolation}

%%%%%%%%%%%%%%%%%%%%%%%%%%%%%%%%%%%%%%%%%%%%%%%%%%%%%%%%%%%%%%%%%%%%%%%
%  Definition: interpolation matrix
%%%%%%%%%%%%%%%%%%%%%%%%%%%%%%%%%%%%%%%%%%%%%%%%%%%%%%%%%%%%%%%%%%%%%%%
\begin{definition}[Interpolation and empirical Gram matrices]
\label{def:interpolation_matrix}
Fix the Chebyshev–Gauss–Lobatto (CGL) nodes
\[
x_j=\cos\!\bigl(j\pi/N\bigr),\;j=0,\dots,N,
\]
and let \(\{\ell_j\}_{j=0}^N\) be the corresponding Lagrange basis polynomials~\citep{trefethen2013approximation}, defined by:
\[
\ell_j(x)
\;=\;
\prod_{\substack{0\le k\le N\\k\neq j}}
\frac{x - x_k}{x_j - x_k},
\qquad j=0,\dots,N.
\]
For any training set
\(\tilde X=\{\tilde x_i\}_{i=1}^M\subset[-1,1]\),
the \emph{interpolation matrix} \(L\in\R^{M\times(N+1)}\) is
\[
L_{i,j} \;=\; \ell_j(\tilde x_i).
\]
Given a vector $f = \{f_j\}_{j=0}^N$ of function values at the Chebyshev nodes,
the evaluations of the degree-\(N\) polynomial
\(f(x)=\sum_{j=0}^{N}f_j\ell_j(x)\)
at the training points satisfy
\(
\bigl[f(\tilde x_1),\dots,f(\tilde x_M)\bigr]^{\!\top}=L\,f.
\)

We also define the \emph{empirical value Gram matrix} as:
\[
G^M_{\mathrm{emp}}
\;=\;
\frac1M\,L^\top L.
\]
\end{definition}

%%%%%%%%%%%%%%%%%%%%%%%%%%%%%%%%%%%%%%%%%%%%%%%%%%%%%%%%%%%%%%%%%%%%%%%
%  Definition: population (continuous) Gram
%%%%%%%%%%%%%%%%%%%%%%%%%%%%%%%%%%%%%%%%%%%%%%%%%%%%%%%%%%%%%%%%%%%%%%%
\begin{definition}[Population (continuous) value Gram]
\label{def:population_gram}
The \emph{population Gram matrix} is
\[
G_{\mathrm{pop}}
\;=\;
\left[ \int_{-1}^1
\ell_j(x)\,\ell_k(x)
\;\frac{dx}{2} \right]_{j,k=0}^N.
\]
By exactness of Clenshaw–Curtis quadrature on the CGL nodes~\citep{trefethen2013approximation},
\(
G_{\mathrm{pop}}
=\diag\bigl(w_0^{\mathrm{CC}},\dots,w_N^{\mathrm{CC}}\bigr),
\)
where:
\begin{equation*}
    w_j^{CC} =
    \begin{cases}
        \frac{\pi}{(2N)} & j = 0, \, N\\
        \frac{\pi}{N} & j = 1, \dots, N-1,
    \end{cases}
\end{equation*}
so \(\kappa^2(G_{\mathrm{pop}})=2\).

Moreover, under uniform sampling,
\(\displaystyle G_{\mathrm{emp}}\to G_{\mathrm{pop}}\)
(as \(M\to\infty\)) in spectral norm almost surely~\citep{vershynin2018high}.
\end{definition}

%%%%%%%%%%%%%%%%%%%%%%%%%%%%%%%%%%%%%%%%%%%%%%%%%%%%%%%%%%%%%%%%%%%%%%%
%  Lemma: concentration of empirical Gram (simplified)
%%%%%%%%%%%%%%%%%%%%%%%%%%%%%%%%%%%%%%%%%%%%%%%%%%%%%%%%%%%%%%%%%%%%%%%
\begin{lemma}[Concentration of the empirical Gram]
\label{lem:empirical_gram_simplified}
With Definitions~\ref{def:interpolation_matrix}–\ref{def:population_gram},
fix \(0<\varepsilon<1\) and \(\delta\in(0,1)\).  There is a universal
constant \(C>0\) such that if
\[
M \;\ge\; C\,
\frac{(N+1)\,\log^{2}(N+1)\,\log\!\bigl((N+1)/\delta\bigr)}
     {\varepsilon^{2}},
\]
then with probability at least \(1-\delta\),
\[
(1-\varepsilon)\,G_{\mathrm{pop}}
\;\preceq\;
G^M_{\mathrm{emp}}
\;\preceq\;
(1+\varepsilon)\,G_{\mathrm{pop}},
\quad
\kappa^{2}(G^M_{\mathrm{emp}})
\;\le\;
2\,\frac{1+\varepsilon}{1-\varepsilon}.
\]
\end{lemma}

\begin{proof}
Each row \(u_i=(\ell_0(\tilde x_i),\dots,\ell_N(\tilde x_i))\) of \(L\)
satisfies
\[
\|u_i\|_2^2 \;\le\; (N+1)\,\Lambda_N^2,
\]
where $\Lambda_N$, the \emph{Lebesgue constant}, satisfies (\cite[Thm.~16.1]{trefethen2013approximation}):
\[
\Lambda_N \;=\;\sup_{x\in[-1,1]}\sum_{j=0}^N|\ell_j(x)|
\;=\;O(\log N).
\]
Hence
\(\|u_i\|_2^2\le C'(N+1)\log^2(N+1)\) for some constant $C'$.
By the matrix–Bernstein inequality~\citep{tropp2012user},  
\[
\|\,G_{\mathrm{emp}} - G_{\mathrm{pop}}\|_{\mathrm{op}}
\;\le\;\varepsilon
\]
with probability $\ge1-\delta$, provided 
\(
M\gtrsim(N+1)\,\log^2(N+1)\,\log((N+1)/\delta)/\varepsilon^2.
\)
This implies
\((1-\varepsilon)G_{\mathrm{pop}}\preceq G_{\mathrm{emp}}\preceq(1+\varepsilon)G_{\mathrm{pop}}\),
and hence 
\(\kappa^2(G_{\mathrm{emp}})\le2(1+\varepsilon)/(1-\varepsilon)\).
\end{proof}

We are now ready to state the full version of~\Cref{thm:bwler_interpolation_informal}:

%%%%%%%%%%%%%%%%%%%%%%%%%%%%%%%%%%%%%%%%%%%%%%%%%%%%%%%%%%%%%%%%%%%%%%%
%  Theorem: ANY data set, sharp ℓ∞ bound
%%%%%%%%%%%%%%%%%%%%%%%%%%%%%%%%%%%%%%%%%%%%%%%%%%%%%%%%%%%%%%%%%%%%%%%
\begin{theorem}[Expressivity--optimization decomposition for interpolation with \methodname]
\label{thm:bwler_interpolation_formal}

Let $f:[-1,1]\!\to\!\R$ extend analytically to the Bernstein ellipse
$E_\rho$ with $\rho>1$ and write $M_f=\max_{z\in E_\rho}|f(z)|$.
Fix training nodes
$\tilde X=\{\tilde x_i\}_{i=1}^M\subset[-1,1]$
and let $L\in\R^{M\times(N+1)}$ be the interpolation matrix
(\Cref{def:interpolation_matrix}).
Define the condition number
\(
\kappa^{2}(L)=\lambda_{\max}(L^\top L)/\lambda_{\min}(L^\top L) = \kappa(L)^2
\)
and the CGL Lebesgue constant
\(
\Lambda_N=\sup_{x\in[-1,1]}\sum_{j=0}^N|\ell_j(x)|.
\)

Initialize an (N+1)-parameter \methodname~with parameters $\theta^{(0)} = 0$. Run $t$ steps of gradient descent on the loss function
$\mathcal L(\theta)=M^{-1}\|L\theta-f_{\tilde X}\|_2^2$
with optimal step–size $\eta=1/\lambda_{\max}(L^{\top}L)$. Denote the parameters of the $t$-th iterate \methodname~polynomial by $\theta^{(t)}$ and the polynomial itself by
$p_N^{(t)} := p_N(x;\theta^{(t)})$.
Then for \emph{any} training set:
\begin{equation}
\label{bound:bwler_interpolation_formal}
\|f-p_N^{(t)}\|_\infty
\;\le\;
\underbrace{\frac{2M_f}{\rho^{N}-1}}_{\text{expressivity}}
\;+\;
\underbrace{\|\theta^{\star}\|_{2}\,\Lambda_N\,
           \exp\!\bigl(-t/\kappa^{2}(L)\bigr)}_{\text{optimization}}
\tag{$\dagger$}
\end{equation}
where
$\theta^{\star}=(f(x_0),\dots,f(x_N))^\top$
interpolates $f$ on the CGL nodes.
\end{theorem}

\begin{proof}
Gradient descent on the quadratic loss function yields
$\|\theta^{(t)}-\theta^{\star}\|_{2}\le e^{-t/\kappa^{2}(L)}\|\theta^{\star}\|_{2}$~\citep{boyd2004convex}.
For any $x\in[-1,1]$:
\begin{align*}
|p_N^{(t)} - p_N^{*}|
  &=\Bigl|\sum_{j=0}^{N}\!\bigl(\theta^{(t)}_j-\theta^{\star}_j\bigr)\ell_j(x)\Bigr| \\
   &\le\Bigl(\sum_{j=0}^{N}\!|\ell_j(x)|\Bigr)\|\Delta \theta\|_\infty          \\
  &\le\Lambda_N\,\|\Delta \theta\|_{2} \\
  & \le\Lambda_N\,\|\theta^{\star}\|_{2}\,
        e^{-t/\kappa^{2}(L)}.
\end{align*}
Taking the supremum in $x$
and adding the standard Bernstein-ellipse expressivity bound (\Cref{thm:spectral_convergence}) finishes
the proof.
\end{proof}

%%%%%%%%%%%%%%%%%%%%%%%%%%%%%%%%%%%%%%%%%%%%%%%%%%%%%%%%%%%%%%%%%%%%%%%
%  Corollary 1: uniform sampling
%%%%%%%%%%%%%%%%%%%%%%%%%%%%%%%%%%%%%%%%%%%%%%%%%%%%%%%%%%%%%%%%%%%%%%%
\begin{corollary}[Uniformly sampled nodes]
\label{cor:uniform}
Let $0<\varepsilon<\tfrac12,\;\delta\in(0,1)$ and draw
$\tilde X$ uniformly without replacement from $[-1,1]$.
If
\[
\textstyle
M\;\ge\; C\,(N{+}1)\,\log^{2}(N{+}1)\,
         \log\!\bigl((N{+}1)/\delta\bigr)/\varepsilon^{2},
\]
then by~\Cref{lem:empirical_gram_simplified}, with probability $\ge1-\delta$:
\[
\kappa^{2}(L)\le 2(1+\varepsilon)/(1-\varepsilon).
\]
Inserting this in~\Cref{bound:bwler_interpolation_formal} gives
\[
\|f-p_N^{(t)}\|_\infty
\le
\frac{2M_f}{\rho^{N}-1}
+\Lambda_N\,\|\theta^{\star}\|_{2}\,
  \exp\!\Bigl(-t\,\tfrac{1-\varepsilon}{2(1+\varepsilon)}\Bigr).
\]
\end{corollary}

%%%%%%%%%%%%%%%%%%%%%%%%%%%%%%%%%%%%%%%%%%%%%%%%%%%%%%%%%%%%%%%%%%%%%%%
%  Corollary 2: Chebyshev-density sampling
%%%%%%%%%%%%%%%%%%%%%%%%%%%%%%%%%%%%%%%%%%%%%%%%%%%%%%%%%%%%%%%%%%%%%%%
% \begin{corollary}[Chebyshev-density sampling]
% \label{cor:cheb}
% If the training nodes are i.i.d.\ from the Chebyshev density
% $\rho_{\mathrm{Cheb}}(x)=1/(\pi\sqrt{1-x^{2}})$
% and $M$ satisfies the same bound as above, then with probability
% $1-\delta$
% \(
% \kappa^{2}(L)\le (1+\varepsilon)/(1-\varepsilon)
% \),
% so the exponent doubles:
% \[
% \|f-p_N^{(t)}\|_\infty
% \le
% \frac{2M_f}{\rho^{N}-1}
% +\Lambda_N\,\|w^{\star}\|_{2}\,
%   \exp\!\Bigl(-t\,\tfrac{1-\varepsilon}{1+\varepsilon}\Bigr).
% \]
% \end{corollary}

%%%%%%%%%%%%%%%%%%%%%%%%%%%%%%%%%%%%%%%%%%%%%%%%%%%%%%%%%%%%%%%%%%%%%%%
%  Precision–conditioning intuition
%%%%%%%%%%%%%%%%%%%%%%%%%%%%%%%%%%%%%%%%%%%%%%%%%%%%%%%%%%%%%%%%%%%%%%%

Intuitively,~\Cref{thm:bwler_interpolation_formal} and~\Cref{cor:uniform} capture a precision-conditioning tradeoff involving $N$ and $M$:
\begin{itemize}
\item \textbf{\(N\) too small (high bias).}
      The expressivity term dominates, and error convergence is exponential (dependent on target function smoothness) as $N$ increases.
\item \textbf{\(N\ll M\) but large.}
      The empirical Gram is well conditioned ($\kappa^{2}(L)=O(1)$) as
      soon as $M\gtrsim N\log^{2}N$, so the training gap decreases exponentially.
\item \textbf{\(N+1=M\) random sampling (poor conditioning).}
      When $M=N+1$ and the points $\{\tilde x_i\}$ are drawn arbitrarily, $L$ is square and generically invertible but has a condition number that grows rapidly with $N$.
      As a result, gradient descent converges only at rate $\exp\!\bigl(-\,t/\kappa^2(L)\bigr)$ with 
      $\kappa^2(L)\gg1$,
      so achieving small training error requires a long training time, $t\gg\kappa^2(L)$.
\end{itemize}

%% file: sections/appendix/theory/pde.tex
%%%%%%%%%%%%%%%%%%%%%%%%%%%%%%%%%%%%%%%%%%%%%%%%%%%%%%%%%%%%%%%%%%%%%%%%
%  Appendix D.2 — PDE version (collocation = CGL nodes)
%%%%%%%%%%%%%%%%%%%%%%%%%%%%%%%%%%%%%%%%%%%%%%%%%%%%%%%%%%%%%%%%%%%%%%%%
\subsection{Formal statement and proof of~\Cref{thm:bwler_pde_informal}}
\label{subapp:theory_pde}

%%%%%%%%%%%%%%%%%%%%%%%%%%%%%%%%%%%%%%%%%%%%%%%%%%%%%%%%%%%%%%%%%%%%%%%
%  Setup
%%%%%%%%%%%%%%%%%%%%%%%%%%%%%%%%%%%%%%%%%%%%%%%%%%%%%%%%%%%%%%%%%%%%%%%
We make the simplifying assumption that our collocation points for the PINN loss are chosen to be the same Chebyshev–Gauss–Lobatto (CGL) nodes 
\[x_j=\cos(j\pi/N),\;j=0,\dots,N\]
as used in the \methodname~parameterization. (This is the ``fixed nodal collocation'' scheme we describe in~\Cref{app:method}.)
Define the Lagrange basis of the CGL nodes
\(\{\ell_j\}_{j=0}^{N}\), and let  
\[
\Lambda_N=\sup_{x\in[-1,1]}\sum_{j=0}^{N}|\ell_j(x)|
          =O(\log N)
\]
be the Lebesgue constant~\citep{trefethen2013approximation}.

\begin{definition}[Collocation matrix for a PDE]\label{def:pde_collocation}
Given a linear differential operator
\begin{equation}
    L=\sum_{\alpha=0}^{d}a_\alpha(x)\,\partial_x^{\alpha}
\end{equation}
and its numerical surrogate
\(\widetilde L\),
define the square collocation matrix
\[
\widetilde A_{i,j}=(\widetilde L\ell_j)(x_i),\qquad
\widetilde b_i=g(x_i)\;(=Lu(x_i)),
\]
where the collocation points are the
\emph{same} CGL nodes \(x_i\).
Also let \(\kappa^{2}(\widetilde A)=
      \lambda_{\max}(\widetilde A^{\top}\widetilde A)/
      \lambda_{\min}(\widetilde A^{\top}\widetilde A).\)
\end{definition}

\begin{definition}[Operator mis-specification]\label{def:op_bias}
For polynomials $v$ of degree \(\le N\), define:
\[
\varepsilon_{\mathrm{op}}(N):=
\sup_{\deg(v)\le N,\;\|v\|_\infty\le1}\|(L-\widetilde L)v\|_\infty.
\]
Intuitively, $\epsilon_{op}(N)$ represents the worst-case bias introduced by replacing the true differential operator $L$ with its numerical surrogate $\widetilde L$.
\end{definition}

We are now ready to state the full version of~\Cref{thm:bwler_pde_informal}:

%%%%%%%%%%%%%%%%%%%%%%%%%%%%%%%%%%%%%%%%%%%%%%%%%%%%%%%%%%%%%%%%%%%%%%%
%  Main theorem (all three gaps)
%%%%%%%%%%%%%%%%%%%%%%%%%%%%%%%%%%%%%%%%%%%%%%%%%%%%%%%%%%%%%%%%%%%%%%%
\begin{theorem}[Expressivity–bias–optimization decomposition for PDE learning with \methodname]
\label{thm:bwler_pde_formal}
Let \(u:[-1,1]\to\R\) solve \(Lu=g\) with analytic data and extend
analytically to the Bernstein ellipse \(E_\rho\) (\(\rho>1\)); set  
\(M_u=\max_{z\in E_\rho}|u(z)|\).
Form the collocation system \((\widetilde A,\widetilde b)\)
from Definition~\ref{def:pde_collocation}.

Initialize an \((N{+}1)\)-parameter \methodname\ with parameters
\(\theta^{(0)}=0\) and run \(t\) steps of gradient descent on the quadratic
loss  
\(\mathcal L(\theta)=\tfrac1{N+1}\|\widetilde A\,\theta-\widetilde b\|_2^{2}\)
using the optimal step size  
\(\eta=1/\lambda_{\max}(\widetilde A^{\top}\widetilde A)\).
Denote the parameters of the \(t\)-th iterate by \(\theta^{(t)}\) and the
resulting polynomial by
\[
u_N^{(t)}(x)
=\sum_{j=0}^{N}\theta^{(t)}_j\,\ell_j(x).
\]
Moreover define
\[
u^*(x)
:=\sum_{j=0}^N u(x_j)\,\ell_j(x),
\qquad
\widetilde u(x)
:=\sum_{j=0}^N \theta^*_j\,\ell_j(x),
\]
where
\(\theta^*=\arg\min_{\theta}\|\widetilde A\,\theta-\widetilde b\|_2\).
Then, when the collocation points coincide with the CGL grid:
\begin{equation}
\label{bound:bwler_pde_formal}
\|u - u_N^{(t)}\|_\infty
\;\le\;
\underbrace{\frac{2M_u}{\rho^{N}-1}}_{\text{expressivity}}
\;+\;
\underbrace{M_u\,\Lambda_N\,\varepsilon_{\mathrm{op}}(N)}_{\text{bias/misspecification}}
\;+\;
\underbrace{\Lambda_N\,\|\theta^*\|_{2}\,
            \exp\!\bigl(-t/\kappa^{2}(\widetilde A)\bigr)}
            _{\text{optimization}}.
\end{equation}
\end{theorem}

\begin{proof}
\textbf{1. Expressivity term.}
This term accounts for the gap between the true solution to the true PDE, $u$, and the best polynomial approximation to it, $u^*$. Let \(u^*\) be the degree-\(N\) interpolant of the true solution on CGL. Then~\Cref{thm:spectral_convergence} yields:
\(\|u - u^*\|_\infty\le2M_u/(\rho^N-1)\).

\smallskip\noindent
\textbf{2. Bias/misspecification term.}
This term accounts for the gap between the best polynomial approximation to the PDE solution, $u^*$, and the true solution to the numerical surrogate, $\widetilde{u}$. At each node,
\[
r_i=(\widetilde L - L)\,u^*(x_i),
\qquad
|r_i|\le M_u\,\varepsilon_{\mathrm{op}}(N).
\]
Hence
\(\|\widetilde A\,\theta^*-\widetilde b\|_\infty
=\max_i|r_i|\), and interpolating these residuals off the grid gives
\[
\|\widetilde u - u^*\|_\infty
\;\le\;
\Lambda_N\,\max_i|r_i|
\;\le\;
\Lambda_N\,M_u\,\varepsilon_{\mathrm{op}}(N).
\]

\smallskip\noindent
\textbf{3. Optimization term.}
This term accounts for the gap between the $t$-th iterate, $u_N^{(t)}$, and the true solution to the numerical surrogate PDE, $\widetilde{u}$. Gradient descent on the quadratic loss function yields
\[
\|\theta^{(t)}-\theta^*\|_2
\;\le\;
\exp\!\bigl(-t/\kappa^{2}(\widetilde A)\bigr)\,
\|\theta^*\|_2.
\]
For any \(x\),
\[
|u_N^{(t)}-\widetilde u(x)|
=\Bigl|\sum_{j=0}^N(\theta^{(t)}_j-\theta^*_j)\,\ell_j(x)\Bigr|
\le\Lambda_N\,\|\theta^{(t)}-\theta^*\|_2,
\]
so
\(\|\widetilde u - u_N^{(t)}\|_\infty
\le\Lambda_N\,\|\theta^*\|_2\,e^{-t/\kappa^2(\widetilde A)}\).

\smallskip
Combining the three bounds yields~\eqref{bound:bwler_pde_formal}.
\end{proof}

%%%%%%%%%%%%%%%%%%%%%%%%%%%%%%%%%%%%%%%%%%%%%%%%%%%%%%%%%%%%%%%%%%%%%%%
%  Corollary A – exact operator on CGL grid
%%%%%%%%%%%%%%%%%%%%%%%%%%%%%%%%%%%%%%%%%%%%%%%%%%%%%%%%%%%%%%%%%%%%%%%
% \begin{corollary}[Exact spectral collocation ($\widetilde L=L$)]
% \label{cor:pde_exact}
% If the surrogate equals the true operator (\(\varepsilon_{\mathrm{op}}(N)=0\)),
% bound \((\star)\) simplifies to
% \[
% \|u-u_N^{(t)}\|_\infty
% \;\le\;
% \frac{2M_u}{\rho^{N}-1}
% +\Lambda_N\,\|w^\star\|_{2}\,
%   \exp\!\bigl(-t/\kappa^{2}(A)\bigr).
% \]
% For many classical differential operators the CGL collocation matrix
% satisfies  
% \(\kappa(A)=O(N^{d})\) (see
% \cite[Chap.~9]{trefethen2000spectral}), whence the GD exponent behaves
% like \(\exp\!\bigl(-t/N^{2d}\bigr)\).
% \end{corollary}

%%%%%%%%%%%%%%%%%%%%%%%%%%%%%%%%%%%%%%%%%%%%%%%%%%%%%%%%%%%%%%%%%%%%%%%
%  Corollary B – $k$th–order finite-difference surrogate
%%%%%%%%%%%%%%%%%%%%%%%%%%%%%%%%%%%%%%%%%%%%%%%%%%%%%%%%%%%%%%%%%%%%%%%
\begin{corollary}[Finite–difference surrogate of order \(k\)]
\label{cor:fd_pde}
Let \(\widetilde L\) replace each \(d\)-th derivative in \(L\) by a
\(k\)-th-order finite–difference stencil on the CGL grid
\citep{fornberg1988generation}.
Then
\[
\varepsilon_{\mathrm{op}}(N)=O\!\bigl(N^{-(k+1-d)}\bigr),
\]
so~\Cref{thm:bwler_pde_formal} yields
\[
\|u-u_N^{(t)}\|_\infty
\;\le\;
\frac{2M_u}{\rho^{N}-1}
\;+\;
\widetilde{O} \left( N^{-(k+1-d)} \right)
\;+\;
\Lambda_N\,\|\theta^\star\|_{2}\,
          \exp\!\bigl(-t/\kappa^{2}(\widetilde A)\bigr).
\]
\end{corollary}

%%%%%%%%%%%%%%%%%%%%%%%%%%%%%%%%%%%%%%%%%%%%%%%%%%%%%%%%%%%%%%%%%%%%%%%
%  Precision–conditioning intuition
%%%%%%%%%%%%%%%%%%%%%%%%%%%%%%%%%%%%%%%%%%%%%%%%%%%%%%%%%%%%%%%%%%%%%%%
Intuitively,~\Cref{thm:bwler_pde_formal} and~\Cref{cor:fd_pde} capture a precision–conditioning tradeoff involving the accuracy of the derivative approximation:
\begin{itemize}
  \item \textbf{\(N\) too small (low precision ceiling).}  
        The expressivity term 
        \(\displaystyle\frac{2M_u}{\rho^N-1}\)
        dominates. Even though error decays exponentially in \(N\), we use too few polynomial basis elements to resolve the solution’s high-frequency features.
  \item \textbf{Low–order finite differences (low precision ceiling, faster convergence).}  
        If \(\widetilde L\) uses a \(k\)th–order stencil with \(k+1-d\) small, then the bias term
        \[
          \text{bias} 
          =M_u\,\Lambda_N\,\varepsilon_{\rm op}(N)
          =O\!\bigl(N^{-(k+1-d)}\log N\bigr),
        \]
        decays only algebraically and dominates.
  \item \textbf{Spectral collocation (high precision ceiling, slower convergence).}  
        With \(\widetilde L\approx L\) and large \(N\), the optimization term
        \(\Lambda_N\|\theta^*\|_2\,e^{-t/\kappa^2(A)}\)
        dominates. For a \(d\)-th order operator the CGL collocation matrix has
        \(\kappa^2(A) = O(N^{2d})\) \citep{trefethen2000spectral}, so GD converges at rate
        \(\exp\!\bigl(-t/O(N^{2d})\bigr)\), requiring \(t\gg N^{2d}\) iterations.
\end{itemize}